\PassOptionsToPackage{table}{xcolor}
\PassOptionsToPackage{numbers,sort}{natbib}
\documentclass{article}
\ifdefined\pdfminorversion
\fi
\usepackage[main,final]{neurips_2026}

\usepackage[T1]{fontenc}
\usepackage{microtype}
\usepackage{graphicx}
\usepackage{subcaption}
\usepackage{booktabs}
\usepackage{tabularx}
\usepackage{caption}
\usepackage{wrapfig}
\usepackage{placeins}
\usepackage{xspace}

\usepackage{amsmath}
\usepackage{amsfonts}
\usepackage{hyperref}

\usepackage[capitalize,noabbrev]{cleveref}
\usepackage{titletoc}

\usepackage{xcolor}
\usepackage{tcolorbox}
\usepackage{enumitem}
\usepackage{listings}
\tcbuselibrary{listings,breakable,skins}

\definecolor{darkblue}{rgb}{0,0,0.5}
\hypersetup{colorlinks=true,citecolor=darkblue,linkcolor=darkblue,urlcolor=darkblue}
\hypersetup{pdftitle={Contrastive Discovery: Open-Ended Scientific Discovery over Competing Explanations},pdfauthor={Ziang Liu, James J. Kim, Yijia Dai, Jennifer J. Sun}}

\newcommand{\gemini}{\texttt{gemini-3-flash}\xspace}
\newcommand{\gpt}{\texttt{gpt-4o}\xspace}
\newcommand{\sonnet}{\texttt{claude-sonnet-4-6}\xspace}
\newcommand{\point}{\textit{pointwise}\xspace}
\newcommand{\Point}{\textit{Pointwise}\xspace}
\newcommand{\cont}{\textsc{Contrastive}\xspace}

\lstdefinestyle{promptstyle}{
    basicstyle=\ttfamily\scriptsize,
    breaklines=true,
    breakatwhitespace=false,
    columns=fullflexible,
    keepspaces=true,
    showstringspaces=false,
    upquote=true,
    language={}
}

\newtcblisting{prompt}[1][]{
    enhanced,
    breakable,
    colback=white,
    colframe=black,
    colbacktitle=black,
    coltitle=white,
    fonttitle=\bfseries\small,
    arc=1mm,
    boxrule=0.5pt,
    left=1mm,
    right=1mm,
    top=1mm,
    bottom=1mm,
    listing only,
    listing engine=listings,
    listing options={style=promptstyle},
    #1
}

\title{\cont Discovery: Open-Ended Scientific Discovery over Competing Explanations}

\author{Ziang Liu\textsuperscript{*}\quad James J. Kim\textsuperscript{*}\quad Yijia Dai\quad Jennifer J. Sun \\[0.65em]
Department of Computer Science \\
Cornell University \\
\href{https://codiscovery.ai/contrastive-discovery}{codiscovery.ai/contrastive-discovery}}

\begin{document}

\maketitle
\begingroup
\renewcommand{\thefootnote}{\fnsymbol{footnote}}
\footnotetext[1]{Equal contribution. Correspondence to \texttt{\{zl873,jjk297\}@cornell.edu}.}
\endgroup

\begin{abstract}

LLM agents are increasingly used for open-ended, data-driven scientific discovery, yet most of them evaluate hypotheses \point, testing an empirical claim $h$ only against its complement $\neg h$. When evidence rejects $h$, the agent learns that one explanation is unlikely but not which alternative the evidence supports. To recover that alternative, we introduce \textbf{\mbox{\cont}~Discovery}, which replaces the single hypothesis with a \emph{contrast set}, a scientific question paired with mutually exclusive and collectively exhaustive competing explanations. Each experiment is scored by how its evidence redistributes belief across the set, and the question is resolved when belief concentrates on one explanation. Because contrast sets change only the unit an agent proposes and scores, an existing agent can use them to re-evaluate evidence it has already collected or to guide its search. Across six domains, 89\% of the \point candidates with the largest belief shifts are rejections. On the same evidence, contrast sets resolve 73\% of these candidates to a specific alternative. When contrast sets guide search, the agent resolves 67\% more questions than \point surprisal under the same experiment budget. In a blinded study, 38 human experts chose the same explanation on 78\% of resolved questions and on 82\% of questions where the \point agent rejected its hypothesis and stopped. Together, these results support explicit competing explanations as a way for discovery agents to resolve more scientific questions from the same evidence and validation budget.
\end{abstract}

\section{Introduction}
\begin{figure*}[t]
\centering
\includegraphics[
    width=1.0\textwidth,
    trim={0in 2in 0in 0in},
    clip
]{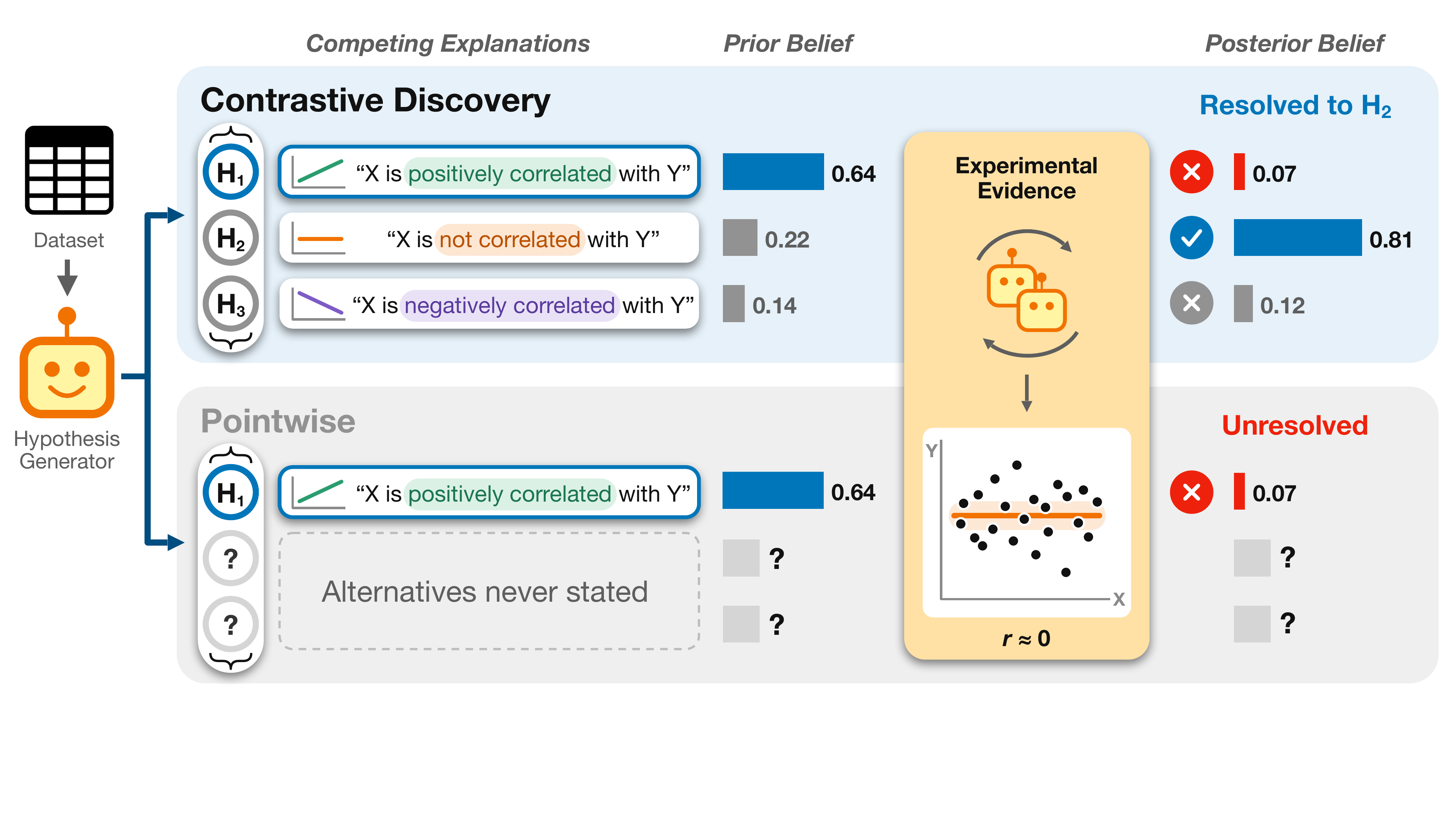}
\caption{
\textbf{A contrast set turns a rejection into a resolution.} A hypothesis generator proposes a contrast set of three competing explanations ($H_1$--$H_3$) for how X relates to Y, and coder, analyst, reviewer, and reviser agents run an experiment that produces evidence. Bars show elicited belief in each explanation before and after the same evidence. \cont Discovery updates belief over all three explanations and moves the lead from $H_1$ (0.64) to $H_2$ (0.81), so the question resolves to $H_2$ and $H_1$ is ruled out. \Point discovery updates belief only in $H_1$ (0.64 to 0.07), so it records a rejection but leaves the question unresolved.
See \S\ref{Sec:cont-discovery} and \S\ref{sec:exp-setup} for details.
}
\vspace{-0.6em}
\label{fig:main-figure}
\end{figure*}

The shift toward autonomous scientific discovery has been accelerated by agents that generate hypotheses, design experiments, collect evidence, and iterate on candidate findings~\citep{lu2024aiscientistfullyautomated, yamada2025aiscientistv2workshoplevelautomated, jansen2025codescientist, schmidgall2025agentlaboratoryusingllm}. Recent systems scale this loop to evaluate hundreds or thousands of discoveries across scientific domains~\citep{agarwal2026autodiscovery, Gottweis2025TowardsAA, popperhuang2025}. At this scale, measuring progress becomes a problem. In closed tasks, progress is measured against fixed objectives, answer keys, or benchmarks~\citep{novikov2025alphaevolve, yuksekgonul2026TTT}, but open-ended discovery has no oracle specifying which discovery should come next. Without an oracle, progress depends not only on how each candidate is scored, but on what \textit{unit} of  knowledge the agent treats as learned.

In most current discovery systems, that unit is a single hypothesis. The agent proposes an empirical claim $h$, tests it, scores the result, and then keeps, revises, or discards $h$~\citep{anonymous2025experigen, majumder2024ddd, Gottweis2025TowardsAA}. The score may be an LLM judgment of novelty or feasibility~\citep{Baek2024ResearchAgentIR, garikaparthi-etal-2025-iris, li2024learning, qi2024large}, or a belief-shift objective such as Bayesian surprise~\citep{agarwal2026autodiscovery, bayesiansurprisal}, which treats discovery as evidence-induced belief change rather than a subjective judgment of interestingness. Yet the representational unit of progress stays the same: evidence updates belief in one focal hypothesis $h$ against its implicit complement $\neg h$. We call this \point discovery. This formulation is shared across agentic discovery systems that otherwise differ in how they propose hypotheses, gather evidence, and score discoveries.

The \point representation is epistemically limiting because scientific evidence often discriminates among plausible alternatives rather than supporting a single hypothesis in isolation. This limitation is clearest when evidence rejects a proposed hypothesis $h$: the agent learns that the focal hypothesis is unlikely, but not which alternative is better supported. For example, evidence against ``self-employed applicants with poor credit face higher mortgage denial rates'' does not say whether their denial rates are lower or no different. Under a fixed evidence-generation budget, this is information-inefficient: the agent may need additional hypothesis proposals and evaluations to resolve a scientific question that the original evidence already helps answer.

We argue that open-ended discovery requires a larger unit of progress: \textbf{a scientific question together with the plausible explanations that could answer it}. This follows contrastive accounts of scientific explanation, in which evidence is informative because it discriminates among alternatives~\citep{vanfraassen1980scientific, lipton2004inference}. We introduce \textbf{\cont Discovery}, which represents each discovery step as a \emph{contrast set} (Eq.~\ref{eq:contrast-set}, \autoref{fig:main-figure}): a set of mutually exclusive and collectively exhaustive (MECE) explanations answering one question. The agent elicits prior and posterior beliefs over the set and scores a candidate discovery by how much its evidence redistributes belief. A question is \emph{resolved} when posterior belief concentrates on one explanation, so a single experiment can both rule out alternatives and identify the best-supported one. Otherwise, the evidence remains informative but the question is \emph{unresolved}.

Contrast sets change the unit an agent proposes and scores, not its architecture, so they can be layered onto an existing agent in two ways. As an \emph{evaluation lens}, they re-evaluate evidence the agent has already collected, without modifying the agent. To \emph{guide search}, they replace the agent's hypothesis and score, so the search is rewarded for evidence that discriminates among competing explanations. We evaluate \cont Discovery across six domains spanning materials science, astronomy, mortgage discrimination, online discourse, software engineering practices, and social behavior. Under a shared scaffold and evidence-generation budget, \point baselines are guided by surprisal, novelty, or feasibility, while \cont Discovery is guided by belief shift over contrast sets. We find:
\begin{itemize}[leftmargin=*, itemsep=0.1em, topsep=0.1em]
    \item \textbf{\Point rejections conceal resolvable alternatives.} Across three elicitor models, a majority of \point proposals receive a ``maybe true'' prior, and 89\% of the highest-surprisal \point candidates are rejections. Used as an evaluation lens on the same evidence, contrast sets resolve 73\% of these candidates to a specific alternative without running new experiments (\S\ref{sec:exp1}).
    \item \textbf{\cont Discovery turns rejections into resolved questions.} Guiding search with contrast sets resolves 67\% more questions than \point surprisal under the same experiment budget. Nearly all the gain comes from corrective resolutions, where evidence shifts support away from the explanation favored beforehand; \point discovery records these cases only as rejections (\S\ref{sec:exp2}).
    \item \textbf{The gain comes from explicit alternatives and holds up under validation.} Ablations trace the gain primarily to evaluating evidence against the full contrast set (\S\ref{results:exp3}). In a blinded study, 38 human experts chose the same explanation on 78\% of resolved questions and on 82\% where the \point agent rejected its hypothesis and stopped (\S\ref{sec:exp4}). The gain holds when each alternative is scored independently with the \point scorer and at matched token and dollar cost (\S\ref{sec:exp5}).
\end{itemize}

\section{Open-Ended, Data-Driven Discovery}
\noindent\textbf{Discovery loop.} We study open-ended, data-driven scientific discovery, where an agent uses a dataset $D$ to propose hypotheses $h \in \mathcal{H}$, evaluates them using evidence computed from the data, and scores them, repeating under a budget of $N$ steps. The setting is open-ended because no predefined reward function determines the value of a discovery, which depends instead on the scientific question, the available evidence, and prior knowledge.
This formulation captures recent LLM-based discovery systems~\citep{agarwal2026autodiscovery}.

\noindent\textbf{\Point discovery.}  We call a discovery method \point when, at step $t$, the agent proposes a single hypothesis $h_t$ and independently assesses it by scoring only that proposed candidate and its evidence. Many recent LLM-based discovery approaches are \point, assessing each hypothesis through novelty and feasibility~\citep{Baek2024ResearchAgentIR, garikaparthi-etal-2025-iris, li2024learning}, evidential support~\citep{Gottweis2025TowardsAA, popperhuang2025}, or Bayesian surprise~\citep{agarwal2026autodiscovery}. In our notation (\S\ref{sec:point-is-binary}), \point discovery is the binary case of \cont Discovery, where the only alternative to $h_t$ is its negation.

\noindent\textbf{The rejection case.} In \point discovery, confirmation and rejection have different epistemic consequences. Confirmation identifies a supported explanation, but rejection only rules against the proposed one. The evidence may bear on alternatives, yet the \point score exposes only its effect on $h_t$.
Resolving among competing explanations then takes a separate step for each alternative, each drawn from the finite budget of $N$ steps, even when one experiment's evidence bears on all of them.

\section{\cont Discovery}
\label{Sec:cont-discovery}

We reframe a discovery step from validating a single candidate hypothesis to resolving a scientific question.
We keep the evidence-induced belief-shift principle of surprisal-based discovery~\citep{agarwal2026autodiscovery, bayesiansurprisal} but change the unit over which belief is represented, from one hypothesis to a structured set of plausible explanations for the same question. The alternatives thereby become part of the unit, so they determine what the evidence is compared against, how belief is updated, and what outcome the step can produce.

\subsection{Constructing contrast sets}
At each discovery step $t$, the agent poses a local scientific question and constructs a contrast set
\begin{equation}
\mathcal{C}_t = \{h_{t,1}, \ldots, h_{t,M}\},
\label{eq:contrast-set}
\end{equation}
where each $h_{t,m}$ is a plausible explanation for the same question. We require these explanations to be locally mutually exclusive and collectively exhaustive (MECE): within the scope of the question, no two of them can hold at once, and together they cover the main outcomes the evidence is meant to distinguish.
For example, for the scientific question ``\emph{Given a poor credit score, how is self-employment associated with mortgage denial rates?}'', the three explanations in the contrast set are that self-employed applicants face higher, no different, or lower denial rates than other applicants (further examples in Appendix~\ref{app:contrast-set-examples}).
\subsection{Eliciting beliefs over contrast sets}
\label{sec:belief-elicitation-contrast}

Given a contrast set $\mathcal{C}_t$, we next define the belief state used to score evidence-induced change. Before observing evidence, we elicit
$p_t^{\text{prior}} \in \Delta^{M-1}$
where $\Delta^{M-1}$ is the probability simplex over the $M$ explanations and $p_{t,m}^{\text{prior}}$ is the subjective belief assigned to explanation $h_{t,m}$. After the agent runs its experiment and observes evidence $e_t$, we elicit
$p_t^{\text{post}} \in \Delta^{M-1}$.
We treat these elicited distributions as the prior and posterior beliefs over the contrast set.

Computing the posterior by Bayes' rule would require a hand-specified likelihood model $p(e_t \mid h_{t,m})$ for every analysis. We therefore obtain each distribution directly through a constant-sum allocation task: a belief elicitor, an LLM separate from the agent (\S\ref{sec:exp-setup}), assigns 100 allocation units across the explanations in $\mathcal{C}_t$, first without and then with access to $e_t$. To reduce stochastic variance, we sample five allocations and average the normalized valid responses (see Appendices~\ref{app:contrastive-surprisal} and~\ref{app:elicitation-stability}). We also randomize the order of explanations during elicitation to mitigate positional bias in LLM-based evaluation (see \autoref{app:hypothesis-order}).

\subsection{Scoring contrastive belief shifts}
\label{sec:scoring}
Because the unit of discovery is a belief distribution over competing explanations, the score should measure how evidence changes that distribution. Given prior and posterior beliefs $p_t^{\text{prior}}$ and $p_t^{\text{post}}$, we compute the Kullback--Leibler divergence ($D_{\mathrm{KL}}$) of $p_t^{\text{post}}$ with respect to $p_t^{\text{prior}}$.
This direction is the Bayesian surprise of \citet{bayesiansurprisal}, the realized information gain from observing $e_t$. The divergence tends to grow with the size of the contrast set, reaching $\log_2 M$ bits when a uniform prior collapses onto one explanation, so we divide by $\log_2 M$ to compare sets of different sizes. We call the result the \emph{contrastive belief shift}, an entropy-scaled Bayesian surprise:
\begin{equation*}
\kappa_t
=
\frac{D_{\mathrm{KL}}(p_t^{\text{post}} \,\|\, p_t^{\text{prior}})}{\log_2 M}.
\end{equation*}
We use $\kappa_t$ as the \cont Discovery score, and larger values mean that the evidence moved more belief across the competing explanations. Scaling does not bound $\kappa_t$ by one, because evidence can move belief onto an explanation the prior considered unlikely. When we use $\kappa_t$ as the tree-search reward (\S\ref{sec:exp-setup}), we therefore clip it at one, $r_t=\min(\kappa_t,1)$, and all reported information-gain analyses use the unclipped $\kappa_t$ (see Appendix~\ref{app:contrastive-surprisal}).

\subsection{Discovery outcomes}
\label{sec:outcomes}
The contrastive belief shift $\kappa_t$ measures the magnitude of belief shift, but not the type of discovery produced by that shift. Since the prior and posterior are defined over the same competing explanations, the update can also be characterized by whether evidence confirms the prior-leading explanation, shifts support to another explanation, rules out alternatives, or leaves the question unresolved. Let
\begin{equation*}
{m}_t^{\star,\mathrm{prior}} = \arg\max_{m} p_{t,m}^{\mathrm{prior}},
\quad
{m}_t^{\star,\mathrm{post}} = \arg\max_{m} p_{t,m}^{\mathrm{post}}
\end{equation*}
denote the indices of the prior- and posterior-leading explanations. We say the question is \emph{resolved} when the posterior assigns sufficient support to one explanation,
\begin{equation*}
\max_{m} p_{t,m}^{\text{post}} \ge \tau_{\mathrm{support}},
\end{equation*}
where $\tau_{\mathrm{support}}$ is the minimum posterior support required to identify a leading explanation (we use $\tau_{\mathrm{support}}=0.80$ and vary it in \mbox{\S\ref{sec:exp5}}). A resolved update is \emph{confirmative} if
${m}_t^{\star,\mathrm{post}} = {m}_t^{\star,\mathrm{prior}}$
and \emph{corrective} if
${m}_t^{\star,\mathrm{post}} \ne {m}_t^{\star,\mathrm{prior}}$, that is, when evidence moves support from the prior-leading explanation to another one. If no explanation reaches $\tau_{\mathrm{support}}$, the outcome is \emph{unresolved}. A generated contrast set may omit the explanation the evidence supports. As a safeguard, we optionally add an \emph{abstention} alternative, ``none of the listed alternatives matches the evidence.'' If its belief reaches $\tau_{\mathrm{support}}$, we flag a possible omission and report no resolution. The main experiments do not include this safeguard, and \mbox{\S\ref{sec:exp4}} tests it.

We separately track explanations that are ruled out: an explanation $h_{t,m}$ is \emph{ruled out} when
$p_{t,m}^{\text{post}} < \tau_{\mathrm{reject}}$,
where $\tau_{\mathrm{reject}}$ denotes a low posterior-support threshold (we use $\tau_{\mathrm{reject}}=0.10$). Ruling out is tracked apart from the three outcomes above, because a single contrastive update can identify a leading explanation and also rule out competing explanations. For the \point setting, where a single hypothesis has scalar prior and posterior beliefs, we classify the update analogously: a hypothesis is \emph{confirmed} if its posterior reaches $\gamma_{\mathrm{support}}$, \emph{rejected} if it is at most $\gamma_{\mathrm{reject}}$, and \emph{ambiguous} otherwise (we use $\gamma_{\mathrm{support}}=0.625$ and $\gamma_{\mathrm{reject}}=0.375$, with calibration details in~\autoref{app:threshold-calibration}). The taxonomy thus separates \emph{how much} support for a hypothesis rises or falls, which \point discovery captures, from \emph{which} explanation the evidence favors, which requires explicit alternatives.

\subsection{\Point discovery as the binary case}
\label{sec:point-is-binary}

\Point belief-shift methods score evidence by how much it changes belief in a single hypothesis $h_t$. In our notation, this is the binary contrast set
$\mathcal{C}_t^{\mathrm{point}} = \{h_t, \neg h_t\}$
with belief state
$p_t = \big(p_{t,1},p_{t,2}\big) \in \Delta^1$,
where $p_{t,1}$ is belief in $h_t$ and $p_{t,2}=1-p_{t,1}$ is belief in its complement.
In this representation, \point discovery is the $M=2$ instance of \cont Discovery, where the only alternative $\neg h_t$ contains all ways the focal hypothesis could be false. Thus, when evidence weakens $h_t$, \point discovery can record negative evidence about the claim but cannot identify which alternative explanation receives that support. \cont Discovery replaces this complement with an explicit local partition (Eq.~\ref{eq:contrast-set}), so the same belief shift is distributed over named competing explanations.

For \point belief elicitation we follow an established categorical procedure~\citep{agarwal2026autodiscovery}: the belief elicitor rates $h_t$ on a five-level scale (definitely false, maybe false, uncertain, maybe true, definitely true), and a ``cannot comment'' response counts as missing. We draw $10$ ratings each for the prior and posterior, map them to numeric scores, fit a Beta distribution to the prior ratings, and update it with the posterior ratings. \Point surprisal is the normalized absolute difference in the two Beta means, and we use it as the \point discovery score (details in Appendix~\ref{app:belief-score}; design comparison in \autoref{app:elicitation-design}).

\section{Experimental Setup}
\label{sec:exp-setup}
We evaluate \cont Discovery in an open-ended, data-driven setting: given a tabular dataset $D$ and metadata, we give the agent a fixed budget of $N=200$ experimental validations. Each validation consists of exploring a scientific question by generating a contrast set, eliciting prior belief over the set, designing and executing an experiment to evaluate it, checking the experiment with a reviewer agent, and eliciting posterior belief.
To choose which question to explore next, we use Monte Carlo tree search (MCTS)~\citep{mcts}, following prior work on agentic scientific discovery and hypothesis search~\citep{agarwal2026autodiscovery, yamada2025aiscientistv2workshoplevelautomated, garikaparthi-etal-2025-iris, sprueill2024chemreasonerheuristicsearchlarge} and on tree search for LLM reasoning and machine learning engineering~\citep{ha-etal-2025-dsg, jiang2025aideaidrivenexplorationspace, toledo2026ai}.
Each node is one experimental validation, and its children are follow-up contrast sets that the agent proposes from the node's result. Search is guided by the upper-confidence bound on trees (UCT)~\citep{uct2006}, with the clipped contrastive belief shift
$r_t=\min(\kappa_t,1)$
as the node reward (reward in Appendix~\ref{app:contrastive-surprisal}; tree structure in Appendix~\ref{app:agent-framework}).

Our primary evaluation metric is the number of resolved scientific questions among the $K$ candidates that a method's own score ranks highest, and the \emph{resolution rate} divides this number by $K$. Candidates are ranked within each run, and counts are averaged over seeds within each dataset and then over the six datasets. A question is resolved when the experiment concentrates posterior belief strongly enough to identify a leading explanation (\mbox{\S\ref{sec:outcomes}}), and for \point baselines a confirmed hypothesis counts as resolved. Among resolved questions, we further distinguish \emph{confirmative} and \emph{corrective} resolutions, corresponding to belief confirmation and belief revision, respectively. Appendix~\ref{app:dedup} repeats the resolved-question count after semantic deduplication.

\noindent\textbf{Discovery agent scaffold.}
Our base agent is the AutoDiscovery scaffold~\citep{agarwal2026autodiscovery}, using \gpt for the discovery agent backbone and a separate \gpt-based belief elicitor (agent details in Appendix~\ref{app:agent-framework}; prompts in \autoref{app:prompts-agent-scaffold}).
In \cont Discovery, the elicitor assigns a belief distribution over all explanations in a contrast set (\mbox{\S\ref{sec:belief-elicitation-contrast}}); in the \point setting, it assigns scalar beliefs for a single hypothesis at a time to compute \point surprisal (see Appendix~\ref{app:belief-score}).

\noindent\textbf{Baselines.}
\label{sec:baselines}
We compare against \point baselines under the same $N=200$ evidence-generation budget. In this setting, each experimental validation proposes a single hypothesis and guides MCTS using \point surprisal. With the number of experimental validations fixed, the comparison asks whether \cont Discovery resolves more questions per validation by discriminating among alternatives. To provide additional \point comparisons, we guide MCTS using LLM-judge scores for hypothesis novelty and feasibility, following prior work on generating and evaluating research ideas~\citep{Baek2024ResearchAgentIR, garikaparthi-etal-2025-iris, hu-etal-2025-nova, li2024learning} and scientific hypotheses~\citep{Gottweis2025TowardsAA, qi2024large, yang2026moosechem} (judge prompts in Appendix~\ref{app:prompt-novelty-llm-judge}).

\noindent\textbf{Datasets.}
\label{sec:datasets}
We evaluate on six datasets spanning scientific and social domains: Materials Project~\citep{Horton2025}, SDSS Galaxy~\citep{sdsscollaboration2025nineteenthdatareleasesloan}, BLADE Mortgage~\citep{gu-etal-2024-blade}, BLADE Conversation~\citep{gu-etal-2024-blade}, DiscoveryBench Requirements Engineering~\citep{majumder2025discoverybench}, and DiscoveryBench NLS Raw~\citep{majumder2025discoverybench}. These datasets vary in feature spaces, sample sizes, and domain structure. Dataset statistics, preprocessing, and filtering criteria are reported in~\autoref{app:dataset-details}. We report results on three independent runs for the first four datasets and one run on the two highest-cost datasets.

\section{Results}
\label{sec:results}

We evaluate whether \cont Discovery increases the epistemic yield of discovery under a fixed evidence-generation budget by tracking confirmed and rejected \point hypotheses, resolved and unresolved contrast sets, and ruled-out alternatives.
We first show that \point rejections conceal alternatives that the same evidence can already resolve (\mbox{\S\ref{sec:exp1}}). We then show that guiding search with contrast sets turns these rejections into resolved questions (\mbox{\S\ref{sec:exp2}}). Finally, we validate the gain: ablations trace it to evaluating evidence against explicit alternatives (\mbox{\S\ref{results:exp3}}), resolutions agree with blinded experts and other independent references (\mbox{\S\ref{sec:exp4}}), and the gain holds across scoring rules, matched token and dollar cost, and resolution thresholds (\mbox{\S\ref{sec:exp5}}).

\subsection{\Point rejections conceal resolvable alternatives}
\label{sec:exp1}

\begin{figure*}[t]
\centering

\includegraphics[
    width=1.0\textwidth,
    trim={0.1in 0.1in 0.1in 0.1in},
]{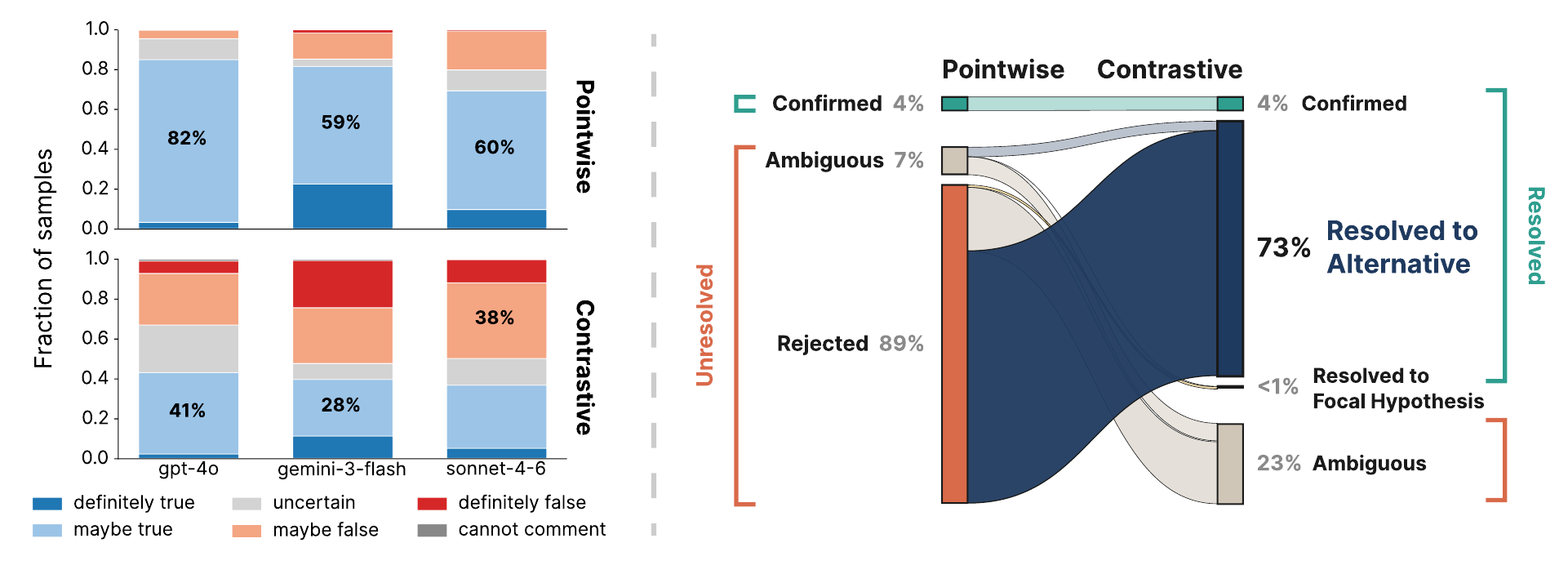}
\caption{
\textbf{\Point discovery masks resolvable alternatives behind rejection.} \textbf{\textit{Left:}} \point proposals show prior collapse, with most mass in the ``maybe true'' bin under each of three belief elicitors (columns), while explanations in contrast sets span a broader range of prior beliefs. \textbf{\textit{Right:}} among the top-50 hypotheses ranked by \point surprisal, 89\% are rejected under \point evaluation, but re-evaluating them with a post-hoc \cont companion on the same evidence resolves 73\% of all 50 to a specific alternative. On the \cont side, \emph{confirmed} marks resolutions to the explanation \point confirmed, and \emph{ambiguous} marks unresolved sets. Per-dataset breakdowns over each run's 200 experimental validations are in \autoref{fig:exp-1-extras-1}.
}
\label{fig:exp-1}
\end{figure*}

We first ask what high-scoring \point surprisal candidates actually represent. \autoref{fig:exp-1}~(left) shows that \point hypothesis proposals concentrate heavily in the ``maybe true'' prior bin across three belief elicitors (at least 59\% for each). This concentration is strongest when the same model, \gpt, both proposes and elicits (82\%), but remains pronounced under cross-model elicitation. In contrast, explanations drawn from contrast sets produce a much broader prior distribution, with the largest bin containing only 41\% of mass (\gpt). This suggests that \point proposals cluster around hypotheses already considered plausible but uncertain, while contrast sets cover a wider range of competing explanations for the same scientific question.

We call this concentration \emph{prior collapse}, and it affects which candidates \point surprisal ranks highest. Because most \point hypotheses begin near the ``maybe true'' bin, there is limited headroom for confirmatory shifts and much more for disconfirming ones (see Appendix~\ref{app:belief-score} for the scoring rule and its normalization). Consistent with this asymmetry, high-surprisal \point candidates in our runs are disproportionately negative. Among the top-50 candidates from each run's 200 experimental validations, only 4\% are confirmed, 7\% are ambiguous, and 89\% are rejected. \Point surprisal thus primarily surfaces evidence against a focal claim, and because the complement is unstructured, that evidence does not identify which alternative explanation is better supported.

To test whether those rejected hypotheses contain recoverable information, we construct a post-hoc contrastive companion for each \point hypothesis: we keep the original hypothesis as one explanation, generate alternatives using the same contrast-set proposer, and reuse the same evidence (prompt in \hyperref[app:companion-generator-prompt]{Appendix}~\ref*{app:companion-generator-prompt}). This uses contrast sets as an evaluation lens: no new experiments are run. The companion generator sees the hypothesis, the dataset's variables, and the same prior context as a live proposer, but not the analysis, its output, or the verdict, and it is applied whether the hypothesis was confirmed or rejected.
On the same evidence, the companion resolves 73\% of the 50 candidates to an alternative (\autoref{fig:exp-1}, right). Counting only confirmations as resolutions, \point discovery leaves 96\% of them unresolved, against 23\% for \cont. Because both readings use the same evidence, the difference comes from how the hypothesis space is represented: the same evidence \point discovery spends to reject a claim can instead identify what is more plausible.

\subsection{\cont Discovery turns rejections into resolved questions}
\label{sec:exp2}

\begin{wrapfigure}[23]{r}{0.48\linewidth}
\centering
\includegraphics[
    width=\linewidth,
    trim={0 0 406pt 0},
    clip
]{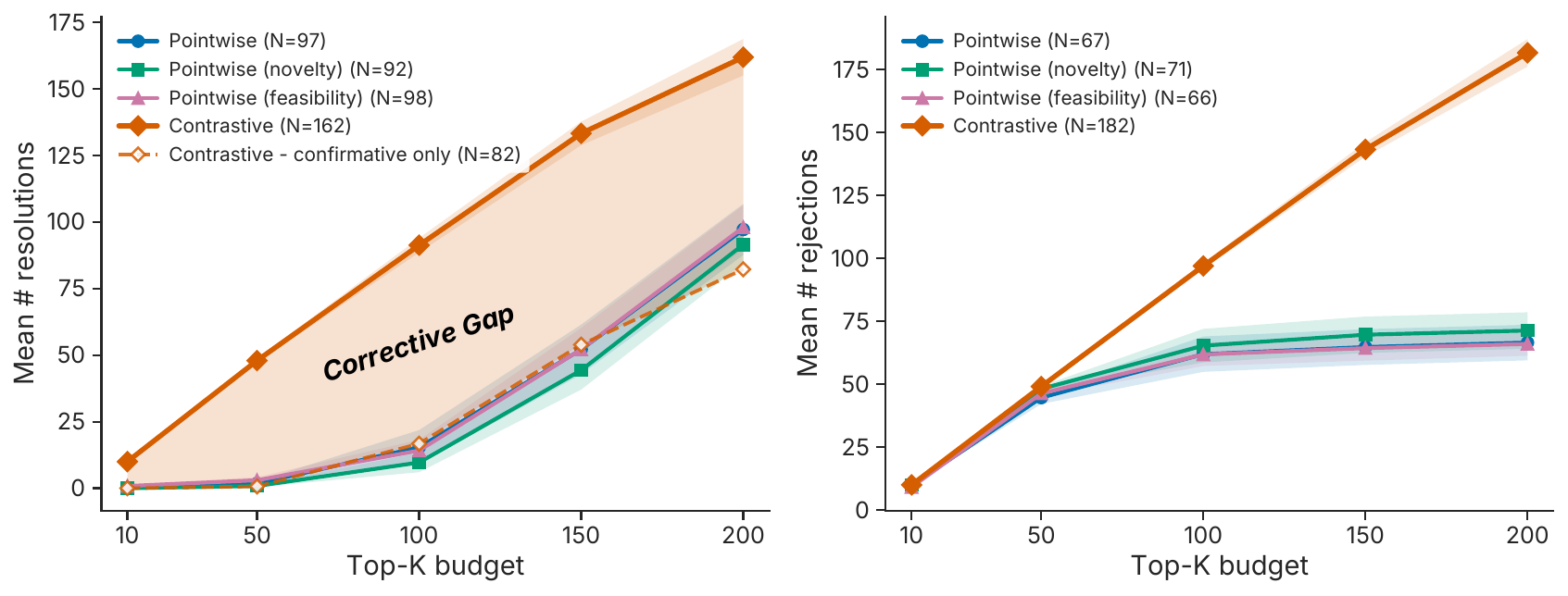}
\caption{
\textbf{\cont Discovery resolves more scientific questions across top-$K$ budgets.} Shaded gap marks corrective resolutions, which \point baselines record only as rejections. The dashed curve counts confirmative resolutions only, and legend values are resolved counts at $K{=}200$. Curves are means over datasets and bands are $\pm1$ standard error across the six datasets. The number of ruled-out alternatives is shown in Appendix~\ref{app:rejection-yield}.
}
\label{fig:exp-2}
\end{wrapfigure}

Having shown that \point rejection often contains information that resolves an alternative, we next test whether using contrast sets to guide search directly captures this latent information to resolve more scientific questions.
For both methods, we rank candidates by their respective discovery scores (i.e., $\kappa_t$ or surprisal) and evaluate the top-$K$ questions retained for inspection.
\Point baselines cluster together (\autoref{fig:exp-2}): ranking candidates by surprisal, novelty, or feasibility yields similar numbers of resolved questions. This suggests that the limitation is in the \point representation rather than its scoring function since an isolated hypothesis can only be confirmed or rejected. In contrast, \cont Discovery produces substantially more resolved questions across budgets: at $K{=}200$, it resolves 161.9 questions on average versus 97.2 for \point surprisal (+67\%).

The confirmative-only \cont curve shows that the gain is not driven by simply confirming more prior-leading explanations: when restricted to confirmative resolutions, \cont Discovery sits at or below the \point surprisal baseline (82.2 vs.\ 97.2 at $K{=}200$), plausibly because $\kappa_t$, as the search reward, values confirming an already-leading explanation less, so the search pursues such questions less often. The additional yield comes almost entirely from corrective resolutions, where evidence shifts support from the prior-leading explanation to another member of the contrast set. This corrective gap is the information lost in \point discovery. There, the same evidence can reject the focal hypothesis, but because the alternative space is collapsed into an unstructured complement, it cannot identify which competing explanation becomes better supported.

Resolving more questions does not come at the expense of ruling out alternatives: placing $0.80$ belief on one explanation leaves at most $0.20$ for the rest, so at the typical size of three, a resolved set almost always rules out at least one alternative (see Appendix~\ref{app:rejection-yield}). To count a verdict in either direction for both methods, we report the \emph{decisive-unit rate}: the fraction of top-$K$ units reaching any verdict (confirmed or rejected for \point, resolved or ruled out for \cont). At $K{=}200$, it is $0.91$ for \cont versus $0.82$ for \point and is higher across all six datasets (Appendix~\ref{app:exp3-extension}).

\subsection{The gain comes from evaluating evidence against explicit alternatives}
\label{results:exp3}

\begin{figure*}[!t]
\centering
\includegraphics[
    width=1.0\textwidth,
    trim={0.0in 0in 0in 0in},
    clip
]{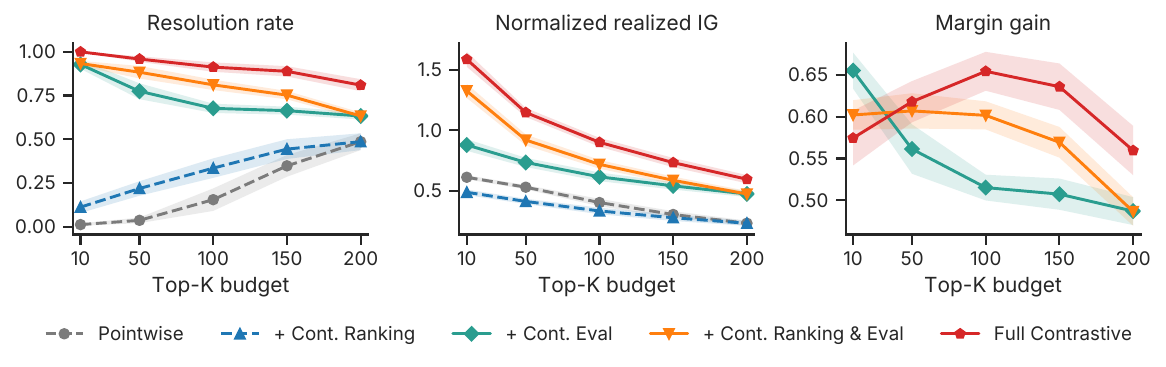}
\caption{
\textbf{\cont ranking, evaluation, and discovery-time construction all contribute to resolved-question yield.} Across resolution rate \textbf{(left)}, realized information gain (IG; the contrastive belief shift $\kappa_t$, not bounded by $1$) \textbf{(middle)}, and margin gain \textbf{(right)}, Full \cont performs best at nearly every budget. \cont (``Cont.'' in the legend) ranking improves resolution even under \point evaluation, \cont evaluation produces a larger step change by allowing evidence to resolve over alternatives, and combining both gives stronger updates than either alone. Since \point evaluation has no explicit runner-up, margin gain is defined only for contrastively evaluated variants. Curves are means and bands are $\pm1$ standard error across the six datasets.
}
\vspace{-0.8em}
\label{fig:exp-3-main}
\end{figure*}

To evaluate which change produces the gain, we isolate three factors in the discovery pipeline: \emph{ranking} (\point surprisal vs.\ \cont $\kappa_t$), \emph{evaluation} (focal hypothesis only vs.\ full contrast set), and \emph{discovery-time construction} (contrast set built post-hoc vs.\ during the discovery run). We compare five variants: \point only, +\cont ranking, +\cont evaluation, +\cont ranking \& evaluation, and Full \cont. Post-hoc variants reuse evidence collected for a \point hypothesis with a companion contrast set; Full \cont constructs the contrast set during discovery time. We evaluate on three metrics: resolution rate measures whether the update answers the question, realized information gain (IG; the contrastive belief shift $\kappa_t$, with $M=2$ under \point evaluation) measures the magnitude of belief shift, and margin gain measures how much the belief gap between the leading explanation and runner-up changes after observing the evidence~(\autoref{fig:exp-3-main}).

The largest separation comes from changing how evidence is evaluated. Variants with \point evaluation remain low in resolution rate (at most 0.22 at $K{=}50$), whereas variants with \cont evaluation improve substantially (ranging from 0.77 to 0.96 at $K{=}50$). Once alternatives are explicitly represented, the same evidence can then adjudicate among explanations. \cont ranking raises resolution under either evaluator, but under \point evaluation it yields lower information gain than \point ranking (\autoref{fig:exp-3-main}, middle), revealing a ranker--evaluator mismatch: the ranker prioritizes evidence informative over alternatives, while the evaluator collapses that evidence back onto a single focal claim. When \cont ranking is paired with \cont evaluation, both resolution and realized information gain increase.

Finally, post-hoc companion sets improve over \point evaluation, but Full \cont leads on all three metrics at almost every top-$K$ budget, consistent with alternatives built during discovery shaping the question, analysis plan, and evidence collected. Resolved-question yield is driven primarily by contrastive evaluation, strengthened by contrastive ranking, and highest when alternatives are built during discovery.

\noindent\textbf{Compatibility with \point surprisal.} The contrastive belief shift stays compatible with the \point surprisal signal used by prior work~\citep{agarwal2026autodiscovery}. To compare the two scores on shared candidates, we take the highest-prior hypothesis of each contrast set and score it with \point surprisal. The two scores are aligned (Spearman $\rho=0.675$); where they disagree, candidates ranked higher by \cont are often corrective and those ranked higher by \point confirmative. Under this single-hypothesis scoring, \cont Discovery also matches the \point surprisal, novelty, and feasibility baselines in its yield of surprising hypotheses, those whose \point surprisal reaches 0.5 (see~\autoref{app:surprisal-compat}).

\subsection{Resolutions hold up against experts and independent references}
\label{sec:exp4}

A resolution only records where one model's belief settled, so we further validate it against independent references that do not depend on that model.

\noindent\textbf{Blinded human expert evaluation.}
Thirty-eight evaluators with MS or PhD degrees judged 60 questions across all datasets using only the question, candidate explanations, and evidence; a median of 8 rated each item, and we use the majority vote. As \autoref{tab:human} shows, experts agreed with \cont Discovery's explanation on 78.4\% of resolved questions. Of the 17 disagreements across all 60 items, they favored a competing explanation in only three cases; the rest were inconclusive or split. For the \point comparison, we used the same 60 questions, evidence, and expert judgments, evaluating only the explanation with the highest prior belief in the contrast set. When \point confirmed its hypothesis, experts endorsed 23 of the 32 findings (71.9\%) for each method; when \point rejected and stopped, they selected the alternative identified by \cont Discovery on 81.8\%. Evaluators also judged contrast sets well formed on 80\% of ratings, flagged a plausible alternative as missing on 15\%, and rated the result a substantive finding on 51\%. As quality controls, evaluators reliably flagged planted defective sets, and pairing questions with unrelated evidence reduced appropriate-analysis ratings from 73\% to 7\% (\autoref{app:human-study}).

\begin{wraptable}[14]{r}{0.46\linewidth}
\caption{\textbf{Experts chose the explanation \cont Discovery identified on most resolved questions.} The same 60 items are grouped by the \cont outcome and the \point verdict.}
\label{tab:human}
\centering\footnotesize
\setlength{\tabcolsep}{4pt}
\begin{tabular}{@{}lcc@{}}
\toprule
Questions & $n$ & Experts agree \\
\midrule
\cont resolved & 51 & 40 (78.4\%) \\
\cont unresolved & 9 & 3 (33.3\%) \\
\Point confirmed & 32 & 23 (71.9\%) \\
\Point rejected & 22 & 18 (81.8\%) \\
\Point ambiguous & 6 & 2 (33.3\%) \\
All items & 60 & 43 (71.7\%) \\
\bottomrule
\end{tabular}
\end{wraptable}
\noindent\textbf{Resolutions follow the evidence.}
To verify that resolutions follow the evidence rather than the wording of the alternatives, we gave the belief elicitor only the numerical results for 14 claims adapted from DiscoveryBench~\citep{majumder2025discoverybench} with documented findings, with no analyst text, the alternatives worded identically across conditions, and their order randomized. Given the coefficient and confidence interval, it ranked the documented answer first in all 11 cases where an interval could be recovered. Holding alternative wording fixed, reversing the evidence's sign reversed resolutions on all 14 claims, while removing evidence prevented resolution. Separately, recomputing 331 analyses directly from raw data reproduced the agent's reported statistics on 96.1\% (\autoref{app:fidelity-audits}).

\noindent\textbf{Agreement across model families.} When we re-elicited beliefs on 180 contrast sets with \gpt, \gemini, and \sonnet, all three models chose the same leading explanation on 55\% of sets before seeing the evidence and on 88.9\% after, and agreement rose in every dataset (see Appendix~\ref{sec:prior-elicitation-audit}).

\noindent\textbf{Incomplete contrast sets.} A contrast set that omits the explanation the evidence supports can produce a confident wrong resolution. For 163 questions whose supported explanation was independently identified from recomputed statistics and which the system had originally resolved correctly, we removed that explanation from its contrast set and re-elicited beliefs. The system declined to resolve on 46.6\% and resolved to a wrong explanation on 53.4\%, mostly to ``no detectable effect.'' Adding the abstention alternative (\mbox{\S\ref{sec:outcomes}}) reduced false resolutions to 0.6\%, typically flagging the omission rather than withholding an answer. On re-elicitation of the same intact sets, abstention was never selected, and 162/163 resolved to the supported explanation versus 159/163 without abstention. Our main results use the generated alternatives without abstention: a strict majority of experts identified a missing alternative in only 6 of 60 rated sets (\autoref{app:incomplete-sets}).

\subsection{The gain holds across scoring rules, costs, and thresholds}
\label{sec:exp5}

To test the robustness of the gain in \mbox{\S\ref{sec:exp2}}, we vary key evaluation choices (details in~\autoref{app:baselines-robustness}).

\noindent\textbf{Joint versus independent scoring.} \cont Discovery both writes out the alternatives and scores them together in one belief distribution; the gain could come from either. Keeping the contrast sets but scoring each alternative separately with the \point elicitor and its threshold $\gamma_{\mathrm{support}}$ resolves 157.8 questions against 161.9 for joint scoring, and both select the same leading alternative on 89.3\% of sets. Joint scoring therefore accounts for little of the gain, and contrast sets can be layered onto an agent that keeps its existing \point scorer.

\noindent\textbf{Budget.} Both methods are given the same number of experimental validations, at similar cost per candidate (61.1k tokens for \point against 52.8k for \cont). We also recompute the comparison with both methods capped at the same total tokens or the same dollar cost. \cont Discovery still resolves more under each: 148.3 against 87.7 at equal tokens and 145.8 against 92.2 at equal dollars.

\noindent\textbf{Thresholds.} If our resolution threshold were easier to clear than the \point confirmation threshold, we would resolve more questions for that reason alone. Because $\tau_{\mathrm{support}}$ bounds a share of a distribution over $M$ alternatives and $\gamma_{\mathrm{support}}$ bounds a scalar belief, we match the two by how often they accept the same evidence. On 1,543 cases where the \point elicitor scored the same evidence for the prior-leading explanation and it stayed the leader, we pair each $\tau_{\mathrm{support}}$ from 0.60 to 0.90 with the $\gamma_{\mathrm{support}}$ that accepts the closest fraction of cases. \cont Discovery resolves more at every matched pair and in each of the six datasets. At our $\tau_{\mathrm{support}}=0.80$, the matched $\gamma_{\mathrm{support}}$ is 0.695, stricter than the 0.625 we use, so the reported gain is conservative (\autoref{app:threshold-calibration}).

\noindent\textbf{Implicit contrast sets.} \Point discovery might independently propose the alternatives in our contrast sets, allowing post-hoc recovery. Using a separate LLM judge, we compare every \point hypothesis against a stratified sample of 300 contrast sets. Among the questions that at least one \point hypothesis also addresses, the \point hypotheses contain the complete alternative set for only 2.6\% within a single run at matched budget, and for only 15.4\% after pooling all \point runs on the dataset ($4$--$6\times$ that budget). At matched budget, the posterior-leading explanation appears among the \point hypotheses 52.0\% of the time when the evidence confirms the prior-leading explanation, but only 25.4\% when it corrects it. So where \point evaluation would reject its hypothesis, it has usually not proposed the explanation the evidence favors (\autoref{app:baselines-robustness}).

\section{Related Work}

Recent surveys trace the shift from task-specific automation toward autonomous LLM-driven discovery systems~\citep{zheng-etal-2025-automation, alkan2025surveyhypothesisgenerationscientific}. Related work develops hypothesis-generation and refinement methods~\citep{duan2025bayesentropycollaborativedrivenagents, ke2025biodiscomultiagenthypothesisgeneration}, alongside pairwise LLM evaluation~\citep{liusie2024llmcomparativeassessmentzeroshot} and scientific argument alignment~\citep{gao-etal-2024-evaluating}. In contrast to such settings that generate or evaluate hypotheses individually or in pairs, \cont Discovery focuses on how evidence distinguishes a set of mutually exclusive and collectively exhaustive competing explanations for the same question.

This reframing builds on a longer tradition of information-theoretic experimental design, which values experiments by their expected effect on belief~\citep{lindley1956, mackay_1992, chaloner1995}. Our objective measures realized posterior--prior belief change through Bayesian surprise~\citep{baldibayesian}, following prior evidence-informed LLM discovery work~\citep{agarwal2026evidenceinformedllmbeliefscontinual, agarwal2026autodiscovery}. ClaimSpect~\citep{kargupta-etal-2025-beyond} also moves beyond binary claim assessment, but organizes existing evidence into non-exclusive aspects rather than constructing mutually exclusive alternatives to be resolved by new analyses. A related line of work also generates research hypotheses from scientific literature, from early text-mining systems~\citep{spangler2014} to LLM agents~\citep{Baek2024ResearchAgentIR, Wang_2024, si2024llmsgeneratenovelresearch}. \cont Discovery is compatible with this setting because contrastive resolution requires only a scientific question and plausible alternatives. In this work, we focus on open-ended, data-driven discovery~\citep{majumder2024ddd, liu2026hypobench} over tabular datasets~\citep{majumder2025discoverybench, gu-etal-2024-blade}, where experiments can be run directly against rich scientific data.

\section{Conclusion and Limitations}
\label{sec:conclusion}
We introduced \cont Discovery, which replaces the single hypothesis that a scientific discovery agent proposes and scores with a contrast set of competing explanations. Because only this unit of discovery changes, an existing agent can use contrast sets without changing its architecture, either to re-evaluate evidence it has already collected or to guide its search. Across six domains, contrast sets turn \point rejections into resolved questions, almost entirely through corrective resolutions. Ablations trace this gain primarily to evaluating evidence against the full contrast set. In a blinded study, human experts broadly agree with the explanations the system selects on resolved questions. Together, these results support explicit competing explanations as a way for discovery agents to resolve more scientific questions from the same evidence and validation budget.

Our experiments use tabular, observational datasets and a single base agent with a \gpt backbone. On such data, a resolution is a statistical rather than a causal finding. Because contrast sets need only an agent's hypotheses and evidence, we expect the gains to carry over to other discovery agents, which we leave to future work. Contrast sets may also be malformed or omit plausible explanations. Structural audits and expert ratings find most sets well formed (Appendix~\ref{app:mece-audit} and \mbox{\S\ref{sec:exp4}}), and adding an abstention alternative flags omitted explanations (\mbox{\S\ref{sec:exp4}}). Finally, resolution is measured with LLM-elicited beliefs, so resolved-question counts depend on the elicitor, even though most resolutions agree with recomputed statistics and experts (\mbox{\S\ref{sec:exp4}}). Future work should test whether these gains extend to laboratory experiments and longer-horizon interventions.

\newpage

\begin{ack}
This material is based on work supported by the AI Research Institutes program supported by the NSF and Intel Corporation under NSF award DMR-2433348.
This research is supported in part by the National Science Foundation (IIS-2505098). Research supported by the NVIDIA Academic Grant Program using 4 × RTX PRO 6000 Blackwell GPUs.

\end{ack}

\bibliography{references}
\bibliographystyle{unsrtnat}

\newpage
\appendix
\section*{Technical Appendices and Supplementary Material}
\startcontents[appendices]
\subsection*{Contents}
\printcontents[appendices]{}{1}{\setcounter{tocdepth}{1}}
\clearpage
\section{\texorpdfstring{Cross-Elicitor Agreement and Prior Collapse}{Cross-Elicitor Agreement and Prior Collapse}}
\label{sec:prior-elicitation-audit}

A useful and effective hypothesis generator should produce claims that span the prior-belief axis, because scientific discovery is bidirectional: evidence may either support an initially unlikely claim or refute an initially plausible one. We audit LLM-elicited priors for hypotheses generated by free-form prompting (\point) and contrastive partitioning (\cont), and identify prior collapse, a failure of free-form generation that the contrastive construction structurally mitigates.

\paragraph{Setup.}
We fix the hypothesis generator to \gpt and re-elicit beliefs using three independent elicitors: \gpt, \gemini, and \sonnet. For each of six datasets, we sample 90 \point hypotheses and 30 \cont sets, stratified by tree level. A \point node contains a single hypothesis, whereas a \cont node contains a partition of $M \ge 2$ mutually exclusive cells (typically $M=3$), each treated as a standalone hypothesis for the per-cell label analysis. This gives us a sample of 540 \point hypotheses and 180 \cont sets (573 \cont cells).

Following our generation process exactly, for each hypothesis (or contrast cell) and elicitor, we draw 10 categorical belief samples over five labels: \texttt{definitely\_true}, \texttt{maybe\_true}, \texttt{uncertain}, \texttt{maybe\_false}, and \texttt{definitely\_false}. We map these labels to scores $1.0$, $0.75$, $0.5$, $0.25$, and $0.0$, respectively, and aggregate them into a $\mathrm{Beta}(\alpha,\beta)$ posterior under a Jeffreys prior. We use the Beta mean as the per-hypothesis belief score. To measure cross-elicitor agreement on \cont at the unit of analysis the runtime metric actually uses, we additionally elicit \emph{set-level} partition allocations: each elicitor distributes 100 probability points jointly over the $M$ cells of a set (5 samples for each set, elicitor, and prior or posterior call, averaged into one $M$-vector).

\paragraph{Prior collapse is a generator-side failure, not an elicitor-side one.}
The prior collapse on \point priors does not imply that elicitors are simply noisy. \autoref{tab:per-dataset-spearman} shows cross-elicitor agreement on per-hypothesis ranks (\point, Spearman $\rho$) and on per-set partition allocations (\cont, mean per-set Pearson $r$, with $1{-}\mathrm{TVD}$ as a total-variation similarity robustness check at the partition level the runtime metric actually uses). On \emph{both} methods, evidence sharply raises agreement: pooled $\rho$ rises $0.660 \to 0.887$ on \point, pooled $r$ rises $0.664 \to 0.923$ on \cont (paired Wilcoxon $p < 10^{-4}$ for both, all per-dataset cells significant under both Bonferroni and Benjamini--Hochberg corrections over 24 tests). The symmetry of this convergence argues against elicitor noise as the source of the unreliability we observe at the prior step: when given evidence, the elicitors are more cross-elicitor consistent on both methods. What is \emph{not} symmetric is the prior collapse, which is a property of the free-form generator that the elicitor cannot fix. Constructing hypotheses as a mutually exclusive partition rather than independent free-form claims reduces the collapse at the source. This matters because discovery is bidirectional: maximal information gain either comes from evidence refuting an initially plausible claim or from evidence supporting an initially unlikely one, and each requires the generator to propose hypotheses on the corresponding side of the prior axis. Under \point prior collapse, the refutational direction dominates. Because the generator rarely produces claims with low elicited prior, discoveries in which evidence \emph{confirms} an initially unlikely claim are rare. \cont's partition tends to include lower-prior cells by forcing explicit alternatives, restoring access to the full prior axis on which discovery operates, while leaving the elicitor's evidence-driven posterior convergence intact.

\paragraph{Do different model families pick the same explanation?}
Agreement on ranks and allocations does not show that the families agree on \emph{which} explanation the evidence supports, so we also compared their leading cells. On the same 180 contrast sets, all three families lead with the same explanation on 55.0\% of sets before the evidence (99 of 180; chance, given each family's observed choice frequencies, is 21.8\%) and on 88.9\% after it (160 of 180; McNemar exact $p=1.2\times10^{-13}$). The rise holds in every dataset (Conversation 63$\to$90\%, Mortgage 63$\to$90\%, Materials 37$\to$93\%, NLS 73$\to$87\%, Requirements 47$\to$90\%, SDSS 47$\to$83\%). How often each family crosses $\tau_{\mathrm{support}}=0.80$ differs (\gpt 77.2\%, \gemini 86.1\%, \sonnet 67.8\%), so the resolution rate depends on the elicitor while the leading explanation does not. Three families that start apart and converge after reading the evidence are hard to explain as one model agreeing with itself. One caveat remains: all three read the same \gpt analyst summary, so part of the convergence could come from shared framing rather than the evidence alone.

\begin{table}[ht]
\centering
\small
\renewcommand{\arraystretch}{1.15}
\setlength{\tabcolsep}{6pt}
\caption{Per-dataset cross-elicitor agreement, mean across the three elicitor pairs. \textbf{\Point columns} ($n{=}90$ hypotheses per dataset, $n{=}540$ pooled): mean pairwise Spearman $\rho$ on per-hypothesis mean-belief scores. \textbf{\cont columns} ($n{\approx}30$ sets per dataset, $n{=}177$ pooled after dropping $M{=}2$ sets from the Pearson statistic): mean pairwise per-set Pearson $r$ on the $M$-cell partition allocation, with mean $1{-}\mathrm{TVD}$ total-variation similarity bracketed as a robustness check (interpretable as fraction of probability mass agreeing across raters; computed on all aligned sets including $M{=}2$). $M{=}2$ sets ($\sim$1.7\% of \cont sets) are degenerate for per-set Pearson and dropped from that statistic only. Bonferroni and Benjamini--Hochberg corrections applied over 24 tests (two methods $\times$ prior and posterior $\times$ 6 datasets, treating each per-dataset cell of this table as the unit of significance claim); all cells significant at $p < 0.05$ under both corrections.}
\label{tab:per-dataset-spearman}
\begin{tabular}{llcc}
\toprule
\textbf{Dataset} & \textbf{Setting} & \textbf{\Point} & \textbf{\cont\ $r$ [$1{-}\mathrm{TVD}$]} \\
\midrule

Conversation & Prior     & 0.730 & 0.764 [0.822] \\
             & Posterior & 0.873 & 0.917 [0.893] \\
\addlinespace[0.3em]

Mortgage     & Prior     & 0.602 & 0.730 [0.786] \\
             & Posterior & 0.854 & 0.934 [0.894] \\
\addlinespace[0.3em]

Materials Project & Prior     & 0.672 & 0.506 [0.780] \\
                  & Posterior & 0.906 & 0.933 [0.897] \\
\addlinespace[0.3em]

NLS Raw      & Prior     & 0.632 & 0.779 [0.776] \\
             & Posterior & 0.875 & 0.936 [0.905] \\
\addlinespace[0.3em]

Requirements Eng. & Prior     & 0.506 & 0.647 [0.837] \\
                  & Posterior & 0.890 & 0.922 [0.878] \\
\addlinespace[0.3em]

SDSS Galaxy  & Prior     & 0.642 & 0.559 [0.737] \\
             & Posterior & 0.862 & 0.898 [0.888] \\
\midrule

\textbf{Pooled} & Prior     & \textbf{0.660} & \textbf{0.664 [0.790]} \\
                & Posterior & \textbf{0.887} & \textbf{0.923 [0.893]} \\
\bottomrule
\end{tabular}
\end{table}

\FloatBarrier
\section{\texorpdfstring{Stability of Contrastive Belief Elicitation}{Stability of Contrastive Belief Elicitation}}
\label{app:elicitation-stability}

Our contrastive score relies on elicited prior and posterior distributions over each contrast set. To reduce stochastic variation, the production pipeline draws $n=5$ point-allocation samples for each prior and posterior elicitation, normalizes each valid allocation to the simplex, and averages the resulting distributions before computing contrastive belief shift. This appendix evaluates whether $n=5$ is a sufficient cost--stability tradeoff for the quantities used by the pipeline.

We analyze $3{,}213$ contrast sets with complete $n=5$ prior and posterior elicitations. For contrast set $i$, let $p_i^{(1)},\ldots,p_i^{(5)}$ denote prior samples and $q_i^{(1)},\ldots,q_i^{(5)}$ denote posterior samples. The production estimates are
\begin{equation}
\bar p_i = \frac{1}{5}\sum_{r=1}^5 p_i^{(r)},
\qquad
\bar q_i = \frac{1}{5}\sum_{r=1}^5 q_i^{(r)}.
\end{equation}
We focus on the two derived quantities used downstream: the realized contrastive information gain
\begin{equation}
L_i = D_{\mathrm{KL}}(\bar q_i \,\|\, \bar p_i),
\end{equation}
and posterior concentration
\begin{equation}
\bar q^{\max}_i = \max_{m} \bar q_{i,m}.
\end{equation}

\begin{figure}[t]
    \centering
    \includegraphics[width=0.62\linewidth]{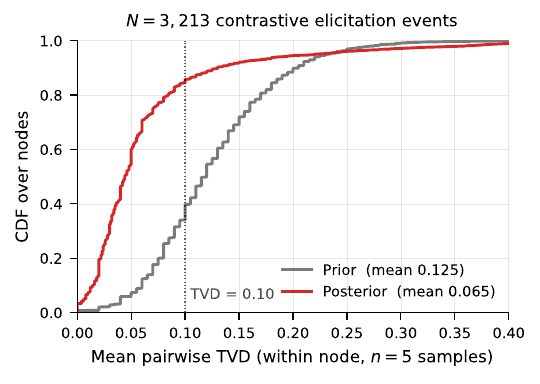}
    \caption{Within-contrast-set dispersion of the $n=5$ belief samples. Prior elicitations are more dispersed, while posterior elicitations are consistently sharper after conditioning on evidence.}
    \label{fig:elicitation-stability-tvd}
\end{figure}

\paragraph{Prior samples are more dispersed, but the averaged score is stable.}
We first measure within-contrast-set dispersion using the mean pairwise total variation distance among the five elicited distributions. Prior samples are more dispersed than posterior samples: the pooled mean pairwise TVD is $0.125$ for priors and $0.065$ for posteriors. This gap is expected, since prior elicitation is performed before observing evidence, while posterior elicitation is conditioned on the experiment result. The higher prior dispersion is not itself a failure mode; it reflects uncertainty before evidence is observed. What matters downstream is whether the averaged prior and posterior produce stable belief-shift scores. Figure~\ref{fig:elicitation-stability-tvd} shows the full CDF, with $85.5\%$ of posterior contrast sets below pairwise TVD $0.10$.

\paragraph{Five samples are not dominated by any single elicitation.}
Table~\ref{tab:elicitation-stability} reports the diagnostics most relevant to the choice of $n=5$. The posterior mean has an average per-hypothesis standard error of $0.0165$, corresponding to about $1.7$ points on the $100$-point allocation scale. The posterior concentration $\bar q^{\max}_i$ is stable under resampling: its median bootstrap $95\%$ CI half-width is $0.029$, and dropping any one of the five samples changes it by only $0.005$ in median. The KL score $L_i$ depends on both the prior and posterior estimates, and is therefore the relevant joint stability check for the discovery score. Although $L_i$ is noisier than $\bar q^{\max}_i$, as expected for a nonlinear function of two finite-sample distributions, its median drop-one drift is only $0.053$ bits. These sensitivities indicate that the production score is not typically driven by a single elicitation sample.

Increasing $n$ would further reduce the standard error, but only at the usual $1/\sqrt{n}$ rate while increasing elicitation cost linearly. We therefore use $n=5$ as a practical tradeoff: it reduces elicitation variance enough for ranking and aggregate analysis while keeping belief elicitation cost manageable.

\begin{table}[t]
\centering
\small
\setlength{\tabcolsep}{5pt}
\caption{Stability diagnostics for the $n=5$ contrastive belief elicitor. Prior allocations are more dispersed than posterior allocations, as expected before evidence is observed. The downstream quantities used by the pipeline remain stable under bootstrap and drop-one sensitivity.}
\label{tab:elicitation-stability}
\begin{tabular}{ll}
\toprule
\textbf{Diagnostic} & \textbf{Value} \\
\midrule
Contrast sets with complete $n=5$ prior/posterior elicitations & $3{,}213$ \\
Mean pairwise TVD, prior samples & $0.125$ \\
Mean pairwise TVD, posterior samples & $0.065$ \\
Posterior contrast sets with pairwise TVD below $0.10$ & $85.5\%$ \\
Posterior per-hypothesis SE of sample mean & $0.0165$ \\
Posterior maximum-probability sample SD & $0.043$ \\
Median bootstrap CI half-width for $\bar q^{\max}_i=\max_{m} \bar q_{i,m}$ & $0.029$ \\
Median drop-one drift in $\bar q^{\max}_i$ & $0.005$ \\
Median drop-one drift in $L_i=D_{\mathrm{KL}}(\bar q_i\|\bar p_i)$ & $0.053$ bits \\
\bottomrule
\end{tabular}
\end{table}

\FloatBarrier
\section{Audit of MECE Structure in Generated Contrast Sets}
\label{app:mece-audit}

\cont Discovery assumes that each contrast set defines a well-formed partition of possible analysis outcomes. In particular, the explanations in a contrast set should be mutually exclusive and collectively exhaustive (MECE): any plausible result of the planned analysis should support at most one explanation, and the set should not omit an obvious residual outcome. The generator prompt (\autoref{app:prompts}) asks for one of a fixed set of partition templates, a residual outcome, and a quantitative criterion that maps the planned analysis to each alternative; this audit checks the sets it actually produced.

\paragraph{Audit protocol.}
We extracted $3{,}260$ parseable contrast sets from the full \cont runs across $6$ datasets and $14$ runs. From this pool, we drew a frozen stratified sample of $384$ contrast sets, stratifying by the requested partitioning dimension and by the number of hypotheses in the set. This intentionally oversamples rare partition types, making failure modes easier to inspect.

Each sampled contrast set was judged by \texttt{gpt-4o} using two independent passes, with $5$ randomized-order votes per pass. The first pass identified pairs of hypotheses that were not logically distinct, including paraphrases, redundant hypotheses, and subsumption relationships. The second pass assigned structural failure tags, including missing residual outcome, redundant hypothesis, subsumption, insufficient number of hypotheses, and inherently non-MECE structure. We classify a contrast set as structurally MECE if no overlap pair, missing-residual, redundancy, subsumption, or inherently non-MECE flag receives at least $4/5$ votes. Insufficient granularity is evaluated separately and does not count as a structural failure. We use this supermajority rule because manual inspection showed that unanimity misses recurring redundancy failures, while simple majority admits more judge noise.

\paragraph{Results.}
Under the primary $4/5$ rule, $372/384$ sampled contrast sets are structurally MECE (Table~\ref{tab:mece-audit-summary}), corresponding to a raw audit pass rate of $96.9\%$. Reweighting the stratified sample back to the full $3{,}260$-item pool gives an estimated population pass rate of $98.4\%$ with a stratified bootstrap 95\% confidence interval of $[97.3\%,99.3\%]$. The conclusion is stable across reasonable vote thresholds: the raw pass rate is $93.8\%$ under a $3/5$ rule and $99.0\%$ under a $5/5$ rule.

\begin{table}[t]
\centering
\small
\caption{Structural MECE audit summary under the primary $4/5$ rule.}
\label{tab:mece-audit-summary}
\begin{tabular}{lcc}
\toprule
Quantity & Count / estimate & Rate \\
\midrule
Structurally MECE sampled contrast sets & $372/384$ & $96.9\%$ \\
Design-weighted population estimate & -- & $98.4\%$ \\
95\% stratified bootstrap CI & -- & $[97.3\%,99.3\%]$ \\
Overlap or redundancy failures & $9/384$ & $2.3\%$ \\
Inherently non-MECE mechanism failures & $3/384$ & $0.8\%$ \\
\bottomrule
\end{tabular}
\end{table}

\paragraph{Failure modes.}
The non-MECE cases are concentrated in two recurring patterns. First, $9$ of the $12$ failures contain overlapping or redundant hypotheses. These usually occur when the proposer includes two versions of a null or residual outcome. For example, one contrast set contains both ``association disappears after controlling for PhD prestige'' and ``no effect of male contributions on total citations.'' These are not exclusive outcomes; both describe the absence of a residual association. Another contrast set contains both ``distribution does not follow a specific crystal system'' and ``no pattern across crystal systems,'' which again duplicate the same residual hypothesis.

Second, $3$ failures arise from mechanism-style contrast sets that are not naturally MECE. These contrast sets ask the hypotheses to distinguish causal pathways, such as whether residence type affects later wealth through social capital or through initial economic advantage. Without stronger identification assumptions, these mechanisms are not mutually exclusive and may not be empirically separable from the proposed analysis. These cases suggest that mechanism-discriminating templates should either be restricted to observationally distinguishable outcomes or excluded from automatic MECE generation.

\paragraph{Cross-provider re-audit.}
The audit above uses a judge from the same provider as the proposer, so we re-ran it on the same 384 sets with Claude Sonnet, keeping the prompts, the two passes, and the $4/5$ rule (\autoref{tab:mece-cross-provider}). The cross-provider judge passes 87.0\% of sets (91.4\% design-weighted) against 96.9\% (98.4\%) for \texttt{gpt-4o}, flagging 50 sets where \texttt{gpt-4o} flagged 12. The same-provider judge is the more lenient of the two, so we report both figures and carry the stricter one. The disagreement runs almost entirely one way (on the overlap check, 24 sets flagged only by Sonnet against 3 flagged only by \texttt{gpt-4o}), and the extra flags are mostly alternatives that overlap or entail one another. They concentrate by question type: half of the mechanism-discriminating sets and 40\% of the ordering sets fail, against none of the directional sets. A structural audit cannot tell whether a plausible explanation is missing altogether; for that, the human raters of \autoref{app:human-study}, who judged the set well formed on 80\% of ratings and flagged a missing alternative on 15\%, are the external estimate.

\begin{table}[htbp]
\centering\small
\caption{Structural MECE pass rate on the same 384-set sample under the same rule, by judge.}
\label{tab:mece-cross-provider}
\begin{tabular}{llccc}
\toprule
Judge & Provider & Raw pass & Design-weighted pass & Sets flagged \\
\midrule
\texttt{gpt-4o} (same provider as proposer) & OpenAI & 96.9\% & 98.4\% & 12 \\
Claude Sonnet & Anthropic & 87.0\% & 91.4\% & 50 \\
\bottomrule
\end{tabular}
\end{table}

\paragraph{Does the defect change reported answers?}
Of the 50 flagged sets, 48 link to a recorded run. On 19 of them the system never reached the resolution threshold, so nothing was reported. Of the 29 resolved questions, 7 selected a member of a judge-named overlapping pair, making those resolutions ambiguous; the other 22 were not identified as implicated by this check. Those 7 are 2.3\% of the 302 resolutions in the audited sample, or 1.4\% reweighted to the corpus (the audit oversamples risky question types). Two further checks point the same way. Mechanically merging each overlapping pair into one alternative, without re-eliciting beliefs, changes the leading explanation on none of the 379 sets that link to a recorded run, and would newly resolve up to 8 questions left open. Flagged sets also reach the threshold less often than clean ones (60.5\% against 85.0\% design-weighted, a 24-percentage-point gap with 95\% CI 10 to 40), as expected when belief splits between two alternatives that mean the same thing. The defect therefore costs findings rather than inventing them; these are patterns in the logged runs, not a re-elicitation. The residual is small: seven resolutions are ambiguous between two overlapping alternatives, and 19 of the 48 flagged sets could not be tested because the judge named no pair. Whether an explanation is missing altogether is a separate failure, tested in \autoref{app:incomplete-sets}.

\paragraph{Interpretation.}
This audit supports the structural premise of \cont: the generated contrast sets usually define valid partitions of the analysis outcome space. The remaining failures are not arbitrary incoherence. They are concentrated in a small number of recognizable template errors, primarily duplicated residual hypotheses and mechanism-style hypotheses that cannot be cleanly partitioned without additional assumptions. We therefore treat MECE structure as satisfied for most sets, while using these failure modes to identify targeted prompt and template improvements. Because a same-provider judge is the more lenient, we carry the cross-provider figure of 87.0\% as the conservative estimate.

\FloatBarrier
\section{Hypothesis-Order Randomization in Belief Elicitation}
\label{app:hypothesis-order}

\cont Discovery elicits two probability distributions over the explanations in a contrast set: a prior distribution before observing the analysis result and a posterior distribution after observing it. Since language models can be sensitive to list position, these probabilities could depend on the order in which hypotheses are shown. We therefore randomize the hypothesis order independently for each elicitation call, then invert the permutation before computing priors, posteriors, KL, or any downstream statistic.

\paragraph{Validation protocol.}
We validate this choice with two checks, summarized in Table~\ref{tab:hypothesis-order}. First, we run a controlled permutation study on four contrast sets from two datasets, with each set evaluated under five different hypothesis orderings. We compare a canonical protocol, which preserves the displayed ordering, against the randomized protocol used in our experiments. After each call, all probabilities are mapped back to the same canonical hypothesis identities. We then measure pairwise consistency across orderings using total variation distance (TVD), where lower TVD means the elicited distribution is less affected by presentation order. Second, we run two parallel \cont runs on BLADE Mortgage that differ only in whether hypothesis order is randomized. This checks whether randomization changes the aggregate scale of the elicited KL updates or posterior confidence.

\begin{table}[t]
\centering
\small
\caption{Effect of hypothesis-order randomization on belief elicitation. The top block measures pairwise consistency across five displayed hypothesis orderings; lower TVD means greater order robustness. The bottom block compares aggregate belief statistics in two parallel \cont runs that differ only in whether hypothesis order is randomized. KL is measured in bits, and $\mathrm{KL}/\log_2 M$ normalizes by the number of explanations $M$ in the contrast set. $\Delta$ is the randomized minus the canonical mean, with a bootstrap 95\% CI.}
\label{tab:hypothesis-order}
\begin{tabular}{lccc}
\toprule
\textbf{Pairwise consistency} & canonical & randomized & ratio \\
\midrule
Prior pairwise TVD & $0.077$ & $0.042$ & $1.83\times$ \\
Posterior pairwise TVD & $0.031$ & $0.025$ & $1.24\times$ \\
\midrule
\textbf{Aggregate belief statistics} & canonical & randomized & $\Delta$ [95\% CI] \\
\midrule
$\mathbb{E}[\mathrm{KL}]$ & $0.987$ & $0.975$ & $-0.012$ $[-0.332,+0.284]$ \\
$\mathbb{E}[\mathrm{KL}/\log_2 M]$ & $0.554$ & $0.568$ & $+0.014$ $[-0.167,+0.181]$ \\
$\mathbb{E}[\max p^{\text{post}}]$ & $0.874$ & $0.873$ & $-0.001$ $[-0.046,+0.045]$ \\
\bottomrule
\end{tabular}
\end{table}

\paragraph{Findings.}
Table~\ref{tab:hypothesis-order} shows that randomization improves pairwise consistency across displayed hypothesis orderings. The effect is strongest for the prior distribution: prior pairwise TVD falls from $0.077$ to $0.042$, a $1.83\times$ reduction in order-induced dispersion. Posterior elicitation is already more stable, but still improves, with posterior pairwise TVD falling from $0.031$ to $0.025$.

The aggregate check shows no evidence that randomization changes the scale of the elicited belief updates. The confidence intervals for average KL, average normalized KL, and average posterior confidence all include zero. Thus, randomization reduces presentation-order sensitivity without a detectable change in the contrastive belief shift.

\paragraph{Interpretation.}
These results justify hypothesis-order randomization as a control for presentation bias. It improves consistency across equivalent prompt orderings while preserving the aggregate behavior of \cont. We therefore randomize hypothesis order during belief elicitation and always map probabilities back to canonical hypothesis identities before computing the metric.

\FloatBarrier
\section{Threshold Calibration}
\label{app:threshold-calibration}

This appendix justifies the numerical thresholds used by the pointwise and contrastive metrics. The main paper defines the contrastive update score $\kappa_t$, the contrastive support/rule-out gates, and the corresponding pointwise gates. Here we explain why the chosen numerical values are principled and empirically aligned.

\paragraph{Contrastive thresholds.}
We use
\[
    \tau_{\mathrm{support}} = 0.80,
    \qquad
    \tau_{\mathrm{reject}} = 0.10 .
\]
The support threshold requires strong posterior concentration on one hypothesis, not merely a plurality winner. At the modal contrast-set size $M=3$, posterior mass $0.80$ leaves at most $0.10$ for each remaining hypothesis under uniform residual splitting, giving an $8\times$ leader-to-average-rival ratio (at least $4\times$ leader-to-runner-up in the worst case). The rule-out threshold $0.10$ is the corresponding residual mass implied by this $M=3$ anchor:
\[
    \frac{1-0.80}{3-1}=0.10 .
\]
We keep this value fixed across $M$ so that ruling out has the same numerical meaning across contrast-set sizes.

For contrastive surprise we threshold the score at $\kappa_t \geq 0.5$. Because $\kappa_t = \mathrm{KL}/\log_2 M$ is divided by $\log_2 M$ but not bounded by one, this is a threshold on the scaled score and should not be read as ``half of the available information gain''; it is placed at the same numerical value as the pointwise belief-change threshold of $0.5$ and avoids the $M$-dependence of a raw KL threshold.

\paragraph{Pointwise thresholds.}
We use
\[
    \gamma_{\mathrm{support}} = 0.625,
    \qquad
    \gamma_{\mathrm{reject}} = 0.375 .
\]
These values are anchored to the elicitor's five-level belief scale
\[
    \{0,\;0.25,\;0.5,\;0.75,\;1.0\}.
\]
The support threshold $0.625$ is the midpoint between ``uncertain'' $(0.5)$ and ``maybe true'' $(0.75)$; the rejection threshold $0.375$ is the midpoint between ``maybe false'' $(0.25)$ and ``uncertain'' $(0.5)$. Thus, the pointwise thresholds correspond to crossing out of the uncertainty region in either direction, rather than being tuned post hoc to maximize agreement with the contrastive thresholds.

\paragraph{Paired calibration check.}
We verify that these independently motivated thresholds align on paired production data. The calibration set contains $3{,}215$ paired elicitations across $6$ datasets and $14$ runs. Each item contains both the contrastive posterior over the full contrast set and the pointwise posterior mean for the focal hypothesis, elicited from the same evidence.

For support, we compare the pointwise support gate against the contrastive support gate on cases where the focal hypothesis is the contrastive posterior winner. For rejection, we compare the pointwise rejection gate against the contrastive rule-out gate on cases where the focal hypothesis is not the contrastive posterior winner. We then evaluate the full two-sided gate on all paired items.

\begin{table}[t]
\centering
\small
\caption{Paired calibration between pointwise and contrastive thresholds. Precision is the fraction of pointwise positives that also satisfy the corresponding contrastive gate; recall is the fraction of contrastive positives recovered by the pointwise gate. Confidence intervals are bootstrap 95\% intervals.}
\label{tab:threshold-calibration}
\begin{tabular}{lccccc}
\toprule
Comparison & $n$ & $F_1$ & 95\% CI & Precision & Recall \\
\midrule
Support gate & $1{,}543$ & $0.937$ & $[0.928,0.947]$ & $0.915$ & $0.961$ \\
Rejection gate & $1{,}672$ & $0.896$ & $[0.884,0.907]$ & $0.981$ & $0.824$ \\
Joint two-sided gate & $3{,}215$ & $0.934$ & $[0.927,0.940]$ & $0.973$ & $0.898$ \\
\bottomrule
\end{tabular}
\end{table}

Table~\ref{tab:threshold-calibration} shows that the pointwise and contrastive gates are well aligned when applied to the same evidence. The support gate reaches $F_1=0.937$, indicating that pointwise support closely tracks contrastive support. The rejection gate reaches $F_1=0.896$ with very high precision, meaning that pointwise rejection almost always agrees with the contrastive rule-out gate. The full two-sided gate reaches $F_1=0.934$ across all $3{,}215$ paired items.

\begin{table}[htbp]\centering\small
\caption{Threshold robustness of the reported gain, recomputed through the exact top-$K{=}200$ estimator of the headline result. Each contrastive threshold $\tau$ is paired with the \point threshold $\gamma$ of matched firing rate; the middle columns show how closely each pair is matched on the $1{,}543$ paired elicitations where both gates score the same evidence. Gain is the difference as a fraction of the budget. $^\dagger$At $\tau{=}0.95$ the positive-set overlap falls to $0.567$, so that row is exploratory.}
\label{tab:threshold-robustness}
\begin{tabular}{cc cc c cc c}\toprule
& & \multicolumn{3}{c}{\emph{evidence that the gates are matched}} & \multicolumn{3}{c}{\emph{resolved units at top-200}} \\
\cmidrule(lr){3-5}\cmidrule(lr){6-8}
$\tau$ & $\gamma$ & contr.\ fires & point.\ fires & F1 & contrastive & pointwise & gain \\ \midrule
0.600 & 0.535 & 95.3\% & 95.3\% & 0.966 & 185.5 & 104.8 & +0.403 \\
0.667 & 0.585 & 93.3\% & 93.5\% & 0.963 & 178.7 & 100.8 & +0.390 \\
0.700 & 0.605 & 92.1\% & 92.0\% & 0.959 & 174.8 & 99.3 & +0.378 \\
0.750 & 0.635 & 90.5\% & 90.6\% & 0.952 & 169.7 & 96.7 & +0.365 \\
\textbf{0.800} & \textbf{0.695} & 86.8\% & 86.3\% & 0.927 & \textbf{161.9} & 88.3 & \textbf{+0.368} \\
0.850 & 0.730 & 80.5\% & 81.1\% & 0.897 & 145.9 & 80.1 & +0.329 \\
0.889 & 0.760 & 71.8\% & 71.4\% & 0.867 & 125.8 & 50.1 & +0.378 \\
0.900 & 0.800 & 65.1\% & 65.1\% & 0.835 & 108.4 & 43.7 & +0.324 \\
$0.950^\dagger$ & 0.895 & 33.8\% & 35.5\% & 0.567 & 50.7 & 20.9 & +0.149 \\
\bottomrule\end{tabular}\end{table}

\paragraph{Threshold robustness.}
The contrastive advantage persists across the tested matched thresholds. We recomputed the headline top-$K{=}200$ comparison of \autoref{sec:exp2} for $\tau_{\mathrm{support}}$ from $0.60$ to $0.95$. Because the two gates sit on different scales, we moved both together: each $\tau$ is paired with the pointwise threshold $\gamma$ that fires on the closest fraction of the $1{,}543$ paired elicitations where both gates score the same evidence, and \autoref{tab:threshold-robustness} reports those firing rates and the overlap of the sets they admit so the pairing can be checked. The contrastive advantage is positive at every threshold, and in every dataset at every threshold (54 of 54 cells). At the reported $\tau=0.80$ the matched $\gamma$ is $0.695$; the paper uses $0.625$, which fires 4.9 percentage points more often, so the published gain is the more conservative one. Raising $\tau$ alone with $\gamma$ fixed at $0.625$ stays positive through $0.90$ and reverses at $0.95$, where the contrastive gate demands 95 of 100 points on one alternative while the pointwise bar stays below ``maybe true''; even at the matched $\gamma=0.895$, the two gates admit substantially different sets there ($F_1=0.567$), so we treat that row as exploratory and rest the claim on $\tau\le0.90$. We present the matched sweep as a robustness check rather than a preregistered test, since it was designed after the fixed-$\gamma$ sensitivity.

\paragraph{Are the allocations calibrated?}
We do not treat the 100-point allocations as calibrated probabilities. $\kappa_t$ is a divergence between two distributions over the same alternatives, and ``resolved'' says that belief has concentrated, not that the leading explanation is 80\% likely to be true. What the metric needs is that allocations be stable, insensitive to presentation, and responsive to evidence, which \autoref{app:elicitation-stability}, \autoref{app:hypothesis-order}, and \autoref{app:fidelity-audits} check. We nonetheless compared stated confidence with a reference outside the model. For the 170 directional contrast sets (125 distinct analyses) whose statistic can be recomputed from the raw data, we read off the cell the recomputed statistic implies. On the 166 the method resolves, the leading explanation agrees with that cell on 163 (98.2\%) at a mean stated confidence of 0.945, so confidence sits 0.037 below agreement (95\% CI $-0.055$ to $-0.015$, clustered on the analysis); the gap remains negative after deduplication and leave-one-dataset-out. The top-label expected calibration error is 0.040, and the multiclass Brier score falls from 0.601 for the prior to 0.047 for the posterior. There is no sign of overconfidence against this reference. This check covers the mechanically decidable directional subset, which is easier than the corpus (97.6\% of these nodes resolve, against about 83\% corpus-wide), and its reference is a significance decision on the same sample; calibration against scientific truth is a separate question.

\paragraph{Interpretation.}
The thresholds are not arbitrary numerical knobs. The contrastive thresholds are anchored to posterior concentration and normalized information gain. The pointwise thresholds are anchored to the elicitor's categorical belief scale. The paired calibration then verifies that these two independently motivated threshold systems produce closely aligned decisions on the same evidence, supporting a fair comparison between pointwise and contrastive metrics rather than one driven by mismatched gate strictness.

\FloatBarrier
\section{Deduplicating Replicated Hypotheses}
\label{app:dedup}

Following prior work~\cite{agarwal2026autodiscovery}, we apply a post-run deduplication pass to test whether a method receives extra credit for repeatedly generating the same semantic hypothesis under different surface forms.

\paragraph{Shared deduplication procedure.}
Both pointwise and contrastive runs use the same semantic-equivalence machinery. We first collapse exact-string duplicates, embed the remaining hypothesis texts with the same embedding model, build a Ward agglomerative clustering tree, and use an LLM judge to decide whether proposed cluster merges are semantically equivalent. Each merge decision is judged with $10$ votes and accepted only when at least $70\%$ of votes mark the two clusters as the same hypothesis. Thus, both methods use the same embedding model, clustering procedure, judge prompt, judge model, vote count, and merge threshold.

\paragraph{Pointwise deduplication.}
In the pointwise setting, each search node proposes one focal hypothesis. Deduplication is therefore performed at the node level: two nodes are merged if their hypothesis texts express the same claim. This matches the representation used by the pointwise metric, where each candidate is a single hypothesis.

\paragraph{Contrastive deduplication.}
In the contrastive setting, each search node proposes a contrast set containing $M$ mutually exclusive hypotheses. Deduplicating only at the node level would be inappropriate: it would treat the whole contrast set as one textual object and would not remove repeated hypotheses across different contrast sets. We therefore flatten contrastive outputs and deduplicate at the hypothesis level. Each hypothesis inside a contrast set is treated as a candidate for cross-node deduplication.

We add one structural constraint: hypotheses from the same contrast set are never merged with each other. They are MECE alternatives to the same analysis question, not replications. For example, ``positive effect,'' ``negative effect,'' and ``no effect'' may be textually related, but they are distinct alternatives by construction. This within-set hard gate prevents deduplication from collapsing valid contrastive alternatives.

\paragraph{Resolved questions after deduplication.}
We recount the resolved questions of \autoref{sec:exp2} so that each resolved claim counts once, grouping confirmed \point hypotheses and the leading explanations of resolved contrast sets by the clusters above. At $K{=}200$, the average number of resolved questions falls from 97.2 to 87.4 for \point surprisal and from 161.9 to 141.9 for \cont Discovery, so duplicates make up 10.1\% and 12.4\% of their resolutions, respectively. \cont Discovery still resolves 62\% more questions than \point surprisal, against 67\% before deduplication, and leads on all six datasets.

\FloatBarrier
\section{Dataset Details}
\label{app:dataset-details}

We evaluate our framework on six datasets spanning scientific and social domains, including materials science, astrophysics, finance and policy, software engineering, sociology and labor economics, and online discourse. Table~\ref{tab:dataset-summary} summarizes the datasets used in our experiments, grouping datasets from the same source benchmark where applicable. We then describe dataset-specific caveats and the filtering logic used for the two largest scientific datasets.

\begin{table}[ht]
\centering
\small
\renewcommand{\arraystretch}{1.15}
\setlength{\tabcolsep}{6pt}
\caption{
Summary of experimental datasets. The benchmark portfolio spans heterogeneous scientific and social domains and covers a broad range of sample sizes, enabling evaluation across different levels of statistical power, feature spaces, and domain structure.
}
\label{tab:dataset-summary}
\begin{tabularx}{\linewidth}{
    >{\raggedright\arraybackslash}p{3.2cm}
    >{\raggedright\arraybackslash}X
    r
    r
    >{\raggedright\arraybackslash}p{3.1cm}
}
\toprule
\textbf{Dataset} & \textbf{Domain} & \textbf{Rows} & \textbf{Cols.} & \textbf{Row unit} \\
\midrule

Materials Project & Materials science & 68{,}805 & 28 & Inorganic crystal \\
SDSS Galaxy (DR19) & Astronomy & 50{,}000 & 28 & Galaxy \\

\addlinespace[0.4em]
\rowcolor{gray!10}
\multicolumn{5}{l}{\textit{BLADE datasets}} \\
\addlinespace[2pt]

Mortgage & Finance / policy & 2{,}380 & 13 & Mortgage application \\
Conversation & Online discourse & 7{,}975 & 70 & Debate contribution \\

\addlinespace[0.4em]
\rowcolor{gray!10}
\multicolumn{5}{l}{\textit{DiscoveryBench datasets}} \\
\addlinespace[2pt]

Requirements Engineering & Software engineering & 276 & 202 & Survey respondent \\
NLS Raw & Sociology / economics & 12{,}686 & 61 & Survey respondent \\

\bottomrule
\end{tabularx}
\end{table}

\paragraph{Materials Project.}
The Materials Project dataset contains 68{,}805 rows and 28 columns, with each row corresponding to a unique inorganic crystalline material identified by \texttt{material\_id}. For our experiments, we restrict the queried subset to materials satisfying \texttt{nelements} $\in \{2,3,4\}$, excluding pure elements and materials with five or more constituent elements. We additionally require \texttt{energy\_above\_hull} $\leq 0.05$ eV/atom to focus on near-stable materials. A key caveat is that elastic moduli are missing for approximately 86\% of rows; only 9.6K materials have computed elasticity values. Consequently, hypotheses involving elastic properties are evaluated on a self-selected subsample rather than on the full dataset.

\paragraph{SDSS Galaxy (DR19).}
The SDSS Galaxy dataset contains 50{,}000 rows and 28 columns, with each row corresponding to one galaxy identified by a unique \texttt{specObjID}. The dataset represents the astronomy and extragalactic domain. Our query applies the following filters: \texttt{class = 'GALAXY'}, \texttt{zWarning = 0} to retain clean redshift fits, redshift in the range $0.02 \leq z \leq 0.3$, and apparent magnitude in the range $14 \leq \texttt{modelMag\_r} \leq 22$. The \texttt{zWarning} filter is applied at query time, although the \texttt{zWarning} column itself is not retained in the saved CSV. The query returns the top 50{,}000 entries under the server-default ordering, which should be interpreted as a minor caveat because the subset is not randomly sampled. Several photometric columns, including \texttt{modelMag\_u}, \texttt{modelMag\_g}, \texttt{modelMag\_z}, \texttt{petroR50\_r}, and \texttt{petroR90\_r}, use \texttt{-9999} as a missing-data sentinel. The pipeline does not filter these sentinels. Where we inspected SDSS analyses during the computation audit (\autoref{app:fidelity-audits}), the agent's own code dropped or replaced them before computing statistics. In addition, \texttt{subClass} is missing for approximately 62\% of rows.

\subsection{\texorpdfstring{BLADE datasets}{BLADE datasets}}

\paragraph{Mortgage.}
The BLADE Mortgage dataset contains 2{,}380 rows and 13 columns, where each row corresponds to one mortgage application. The dataset is situated in the finance and policy domain, with particular relevance to lending and discrimination. Its columns are:

\texttt{female}, \texttt{black}, \texttt{housing\_expense\_ratio}, \texttt{self\_employed}, \texttt{married}, \texttt{mortgage\_credit}, \texttt{consumer\_credit}, \texttt{bad\_history}, \texttt{PI\_ratio}, \texttt{deny}, \texttt{loan\_to\_value}, \texttt{denied\_PMI}, \texttt{accept}

Missingness is limited but nonzero: \texttt{female} contains 18 missing values, \texttt{married} contains 2 missing values, and \texttt{PI\_ratio} contains 5 missing values.

\paragraph{Conversation.}
The BLADE Conversation dataset contains 7{,}975 rows and 70 columns, with each row corresponding to one contribution, or comment, in an Edge.org debate thread. The dataset belongs to the online discourse and gender participation domain. It contains repeated observations along two axes: 522 unique \texttt{ThreadId} values, with an average of approximately 15 rows per thread and a maximum of 436 rows in a single thread; and 728 unique \texttt{Id\_num} contributors, with an average of approximately 11 rows per contributor and a maximum of 504 rows for a single contributor. To avoid contributor leakage, we use an \texttt{Id\_num}-level split: 70\% of contributors are assigned to discovery and 30\% to validation. This preserves row-level granularity while preventing contributor-specific attributes from leaking across splits. We prioritize contributor-level splitting because the contributor axis contains more invariant attributes, including gender, PhD institution, and h-index.

\subsection{\texorpdfstring{DiscoveryBench datasets}{DiscoveryBench datasets}}

\paragraph{Requirements Engineering for ML-Enabled Systems.}
The DiscoveryBench Requirements Engineering dataset contains 276 rows and 202 columns, with each row corresponding to a unique survey respondent identified by \texttt{ID}. The dataset belongs to the software engineering domain and captures practices in requirements engineering for ML-enabled systems. Because the dataset has only 276 respondents, it has limited statistical power for detecting subtle effects. In addition, many of the 202 columns are binary indicators derived from multi-select survey questions, which should be considered when interpreting hypotheses involving sparse or highly correlated features.

\paragraph{NLS Raw.}
The DiscoveryBench NLS Raw dataset contains 12{,}686 rows and 61 columns, with each row corresponding to a unique NLSY79 survey respondent identified by \texttt{ID}. The longitudinal panel has already been flattened into a wide-format representation with one row per person. The dataset is situated in the sociology and labor economics domain. A key caveat is that columns mix measurements from different survey years, including 1979, 1981, and later waves. Therefore, hypotheses over this dataset should respect the implicit temporal ordering of variables and avoid treating later measurements as valid predictors of earlier outcomes.

\subsection{\texorpdfstring{Filtering rationale for scientific datasets}{Filtering rationale for scientific datasets}}
\label{app:scientific-filtering-rationale}

For the two large scientific datasets, Materials Project and SDSS Galaxy, we apply simple domain-motivated filters before running hypothesis generation. These filters are not intended to optimize downstream performance. Instead, they define physically coherent, cleanly measured, and computationally tractable hypothesis spaces. In both cases, the goal is to avoid forcing the model to generate hypotheses across incompatible physical regimes or over measurements known to be unreliable.

\paragraph{SDSS Galaxy.}
For SDSS Galaxy, we restrict the dataset to entries with \texttt{class = 'GALAXY'}. This keeps the hypothesis space physically coherent: stars, quasars, and galaxies obey different physical regimes, and several galaxy-oriented columns, such as size and profile-radius measurements, would be meaningless or differently interpreted for non-galaxy objects. We also require \texttt{zWarning = 0}, the standard SDSS quality flag indicating a clean redshift fit. Nonzero warning values often indicate failed pipeline checks, such as low signal-to-noise, bad template matches, or problematic spectra; including these entries would inject unreliable redshift-dependent measurements into many hypotheses.

We further restrict redshift to $0.02 \leq z \leq 0.3$. The lower bound removes very nearby objects for which peculiar velocities can be comparable to the Hubble-flow velocity, making redshift a poor distance proxy. The upper bound keeps the sample within the approximate regime of the SDSS Main Galaxy Sample, rather than mixing in higher-redshift luminous red galaxies or other populations governed by different selection effects. Finally, we require $14 \leq \texttt{modelMag\_r} \leq 22$. The bright-end cut avoids saturated sources for which photometry may be unreliable, while the faint-end cut avoids objects for which photometric uncertainty and redshift signal-to-noise become limiting. The top-50{,}000 cap is a practical scale constraint: it remains below the SDSS unauthenticated query limit, fits comfortably in memory, and provides sufficient statistical power for hypothesis generation. Unlike the physical and quality filters, this cap is primarily pragmatic rather than physically principled.

\paragraph{Materials Project.}
For Materials Project, we restrict to materials with \texttt{nelements} $\in \{2,3,4\}$. This excludes pure elements, whose small and heavily studied set would make many property hypotheses trivial or degenerate. It also excludes materials with five or more elements, such as high-entropy alloys and complex multicomponent oxides, where individual elemental contributions are often highly entangled and harder to interpret cleanly. Retaining binary, ternary, and quaternary compounds captures much of functional inorganic chemistry, including families such as rock salts, perovskites, spinels, and Heusler compounds, while keeping the hypothesis space tractable.

We additionally require \texttt{energy\_above\_hull} $\leq 0.05$ eV/atom. This threshold follows the common Materials Project convention for selecting materials that are plausibly synthesizable or near-stable. Materials substantially above the convex hull are increasingly likely to be hypothetical or physically unstable, making their computed properties less useful for generating hypotheses about realizable materials. This filter also removes unusual metastable phases whose extreme properties could dominate regressions or produce hypotheses driven by outliers rather than robust physical structure.

\paragraph{Overall design principle.}
Across both scientific datasets, the filters are chosen to satisfy four criteria. First, retained samples should be physically coherent, so that hypotheses are generated within a comparable regime. Second, measurements should be clean enough to avoid obvious garbage-in, garbage-out effects, such as failed redshift fits or saturated photometry. Third, the filtered datasets should remain large enough to support statistically meaningful hypothesis generation. Finally, the filters should be conventional enough to be defensible to domain experts, rather than appearing as post hoc cherry-picking.

\paragraph{Portfolio coverage.}
The resulting benchmark portfolio covers six domains: materials science through Materials Project, astrophysics through SDSS Galaxy, finance and policy through BLADE Mortgage, software engineering through DiscoveryBench Requirements Engineering, sociology and labor economics through DiscoveryBench NLS Raw, and online discourse through BLADE Conversation. The datasets also span a broad range of sample sizes: 276 rows for DiscoveryBench Requirements Engineering, 2{,}380 rows for BLADE Mortgage, 7{,}975 rows for BLADE Conversation, 12{,}686 rows for DiscoveryBench NLS Raw, 50{,}000 rows for SDSS Galaxy, and 68{,}805 rows for Materials Project. This row-count spread is useful for analyzing how hypothesis-generation and validation behavior vary under different statistical-power regimes.

\FloatBarrier
\section{\texorpdfstring{Heritability Analysis: Parent--Child Score Correlation in MCTS over LLM Hypotheses}{Heritability Analysis: Parent-Child Score Correlation in MCTS over LLM Hypotheses}}
\label{app:heritability}

\begin{figure}[t]
  \centering
  \includegraphics[width=0.95\linewidth]{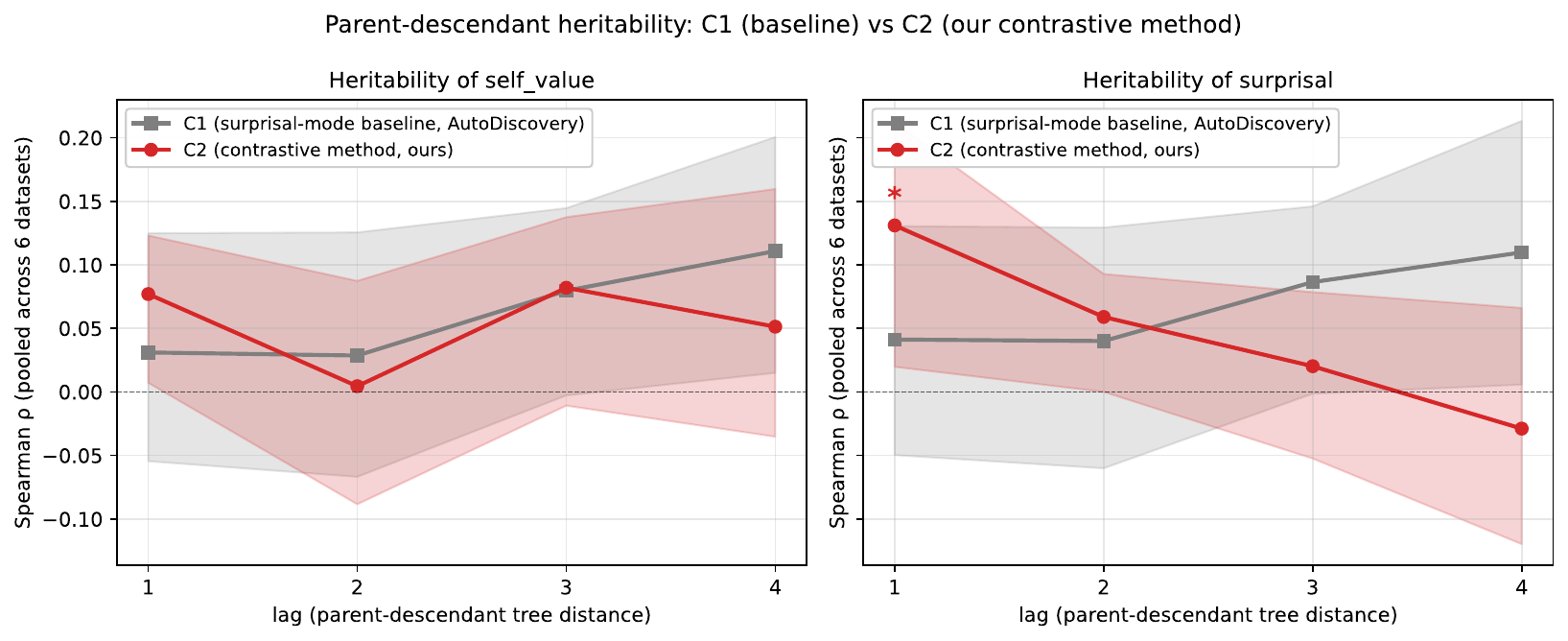}
  \caption{\textbf{Parent--descendant heritability across MCTS lags.}
  Spearman rank correlation between a parent's score and its
  descendant's score at increasing tree distance, pooled across 6
  datasets and the 200 experimental validations of each
  tree. Solid lines are point estimates; shaded regions are 95\%
  percentile cluster-bootstrap CIs over runs. \textbf{Left:}
  \texttt{self\_value} (the UCT-propagated reward, the
  mechanism-defining metric). \Point (gray) stays near zero at lags 1 and 2 and rises at deeper
  lags. \cont (red, ours) shows a positive lag-1 trend ($\rho = +0.077$)
  that is not significant after BH-FDR correction ($q = 0.28$). \textbf{Right:} \texttt{surprisal} (the logged belief change without clipping or $M$-normalization; a robustness check). \cont lag-1
  is FDR-significant ($\rho = +0.131$, $q = 0.046$, $\star$ marks
  $q \leq 0.10$) and decays monotonically toward zero by lag 4, as local
  inheritance would predict. \Point shows no lag-1 heritability
  on either metric; its slow rise at deeper lags is consistent
  with UCT's preferential extension of high-value lineages enriching
  deep-lag pair pools (selection bias, not mechanism-relevant signal).
  The weaker \texttt{self\_value} result is consistent with
  construction-induced compression (13.7\% of nodes hit the clipping
  ceiling) of the same underlying signal that the \texttt{surprisal}
  panel resolves more cleanly.}
  \label{fig:heritability}
\end{figure}

A precondition for hierarchical MCTS to outperform flat sampling is that
ancestor scores predict descendant scores. Without such correlation,
UCT's value-back-propagation has nothing to exploit when choosing which
subtree to expand, and the selection rule is statistically equivalent to
uniform-random parent selection. To test whether this property holds in
our setting, we measured the parent--child Spearman rank correlation of
\texttt{self\_value} (the scalar UCT propagates) across the 200 experimental validations of each search tree, pooled across six
datasets.\footnote{All confidence intervals are 95\% percentile cluster
bootstraps over the 6 runs (2{,}000 iterations); $p$-values are from
within-level permutation tests (2{,}000 permutations) with add-one
smoothing; BH $q$-values are
Benjamini--Hochberg-adjusted across the per-metric 7-lag family. All results reproduce from
\texttt{-{}-seed 0}.} We use \texttt{self\_value} as the headline metric because
it is what UCT directly reads when computing $Q$-values. As a
robustness check we additionally report the underlying
\texttt{surprisal} (the logged belief change: $|\mu^{\text{post}} - \mu^{\text{prior}}|$ in the
surprisal-mode baseline, $H^{\text{prior}} - H^{\text{post}}$ in
the contrastive method), which has none of the clipping or
$M$-normalization construction in \texttt{self\_value}. For the contrastive method this secondary quantity is the entropy reduction logged as a diagnostic alongside each node, not the contrastive belief shift $\kappa_t$ of \autoref{sec:scoring}; \texttt{self\_value} is the clipped $\kappa_t$ that UCT reads. Reporting both
addresses the concern that an observed parent--child correlation could
be a construction artifact rather than a real property of the search
tree.

The surprisal-mode baseline (\point) yields essentially zero
parent--child heritability at lag~1 ($\rho = +0.031$, 95\% CI
$[-0.055, +0.125]$, $n = 1{,}169$ pairs, BH-FDR $q = 0.79$):
UCT-driven selection in this regime operates without exploitable score
signal. Our contrastive partition reward (\cont) shows a positive lag-1
trend ($\rho = +0.077$, 95\% CI $[+0.007, +0.123]$, $n = 1{,}136$, raw
permutation $p = 0.04$) that is not significant after Benjamini--Hochberg
correction across the 7-lag family ($q = 0.28$) but FDR-significant on
the underlying \texttt{surprisal} score ($\rho = +0.131$, $q = 0.046$).
The \cont trend on \texttt{surprisal} decays monotonically toward zero by lag~4,
consistent with local inheritance. The baseline shows no
such decay structure on either metric. See \autoref{fig:heritability}.

We note that the \cont correlation is consistent with two mechanisms our
analysis cannot uniquely distinguish: (a)~inheritable signal in the
score itself, exploitable by UCT-selection; or (b)~a downstream effect
of the structured $M$-cell partition being available as ancestor
context to the next-level proposer, which would correlate child and
parent scores through generation rather than through selection. The
natural ablation, comparing UCT-selection to uniform-random parent
selection while preserving context inheritance, would adjudicate (a)
vs.\ (b) and is left to future work.

\FloatBarrier

\section{Blinded Expert Evaluation}
\label{app:human-study}

This appendix gives the protocol, reliability, and control analyses behind \autoref{tab:human}.

\paragraph{Protocol.}
Thirty-eight evaluators with MS or PhD degrees in the mathematical, computational, or natural sciences, recruited and screened separately from the authors (31 through a screened crowd platform, 7 verified in person), rated 60 blinded items: ten per dataset, stratified by the contrastive score $\kappa_t$. Each evaluator saw only the research question, the candidate explanations, and the statistical evidence from the executed analysis, with the alternatives in random order; our system's beliefs, its resolution label, and the \point verdict were never shown. Each item was rated by at least 4 evaluators (median 8). The four questions per item were (Q1) which explanation the evidence supports, with the extra options ``none of these is supported'' and ``cannot tell''; (Q2) whether the alternatives are mutually exclusive and cover the plausible explanations, with options for a missing explanation, overlap, or both; (Q3) whether the result is worth reporting; and (Q4) whether the chosen variables and test fit the question. We score items by majority vote and count no majority, ``cannot tell,'' and ``none'' as disagreement with our system's explanation. Five submissions were excluded under a rule fixed at the first exclusion (one incomplete session and four evaluators who missed both attention checks). Restricting to the 30 evaluators who caught every attention check lowers the overall agreement from 71.7\% to 68.3\%, so inattentive raters do not carry the result. Expert evaluations were collected with informed consent and kept confidential; the information concerned our methods and scientific outputs, not the evaluators as individuals. Under our institution's IRB guidance, this activity does not constitute human-subjects research and therefore did not require IRB review.

Both methods were compared on the same 60 questions, using the same evidence and expert judgments. The \point hypothesis was the explanation our system rated most plausible before seeing the evidence, evaluated on its own.

\paragraph{Resolution agreement (Q1).}
The expert majority matched our system's explanation on 43 of 60 items (71.7\%, 95\% CI 60.0--83.3 by item-clustered bootstrap): 40 of the 51 items it marked resolved (78.4\%) and 3 of the 9 it left unresolved. Of the 17 disagreements, the experts chose a competing explanation on only 3; on 8 they judged the evidence insufficient and on 6 they formed no majority, so most disagreements fall on evidence the experts themselves could not settle. Across all 60 items the panel committed to a specific alternative on 46, which shows that the executed experiments usually separate the alternatives well enough for experts to commit. Agreement was 79\% on confirmative resolutions and 77\% on corrective ones. Against the \point baseline: on the 32 items it confirmed, the experts endorsed 23 findings for each method; on the 22 items it rejected and stopped, the alternative our system identified matched the expert majority on 18 (81.8\%; exact 95\% CI 60--95). A related subset, the 22 items the baseline did not confirm and our system resolved, overlaps this one on 19 items and gives 17 of 22; the main text reports the pointwise-rejected subset. Per dataset, agreement ranges from 6 of 10 to 9 of 10, too few items to compare domains.

\paragraph{Reliability.}
On the load-bearing judgment (Q1, four to six options), pairwise agreement was 64\% and Krippendorff's $\alpha$ was 0.54 (95\% CI 0.45--0.61), and in a descriptive stratification $\alpha$ was 0.60 on resolved items versus 0.22 on unresolved ones, so rater disagreement concentrates on the items \cont Discovery leaves unresolved. On set completeness (Q2), 80\% of ratings judged the set well formed, 15\% flagged a missing plausible alternative, 4\% overlap, and 1\% both; raw agreement on the binary omission call was 83\% (Gwet's AC1 $=0.73$; the lower Fleiss' $\kappa$ of 0.35 reflects the 80/20 response skew). A strict majority flagged a missing alternative on 6 of the 60 sets. Analysis appropriateness (Q4) was judged appropriate on 73\% of ratings, wrong variable or test on 19\%, and cannot tell on 8\% (binary raw agreement 72\%, AC1 $=0.54$). Scientific value (Q3) had low agreement ($\alpha=0.24$), so we report it only as a distribution: 51\% substantive finding, 20\% minor, 3\% too obvious to state, 25\% not a finding.

\paragraph{Validity controls.}
To check that the ratings track genuine quality, we planted two manually constructed defective sets among the items, one with a plausible alternative removed and one with overlapping alternatives: 11 of the 14 evaluators who saw the first flagged a missing alternative and none confused it with overlap, and 10 of 12 flagged the second as defective. On separate control items we kept the contrast set intact but paired it with the evidence of a different question: the well-formed rate stayed at 77\%, close to its 80\% baseline, while the analysis-appropriateness rating fell from 73\% to 7\% and the wrong-variable-or-test flag rose from 19\% to 84\%. The evaluators therefore flag real set defects without blanket-flagging, and recognize when an analysis does not fit its question. Attention controls (mismatched evidence expected to yield ``cannot tell'') fired as expected on 95\% of exposures.

\paragraph{Scope.}
The experts read the same evidence as our system, so their agreement shows that resolutions track the evidence as independent experts read it; it is not a test of scientific truth. The experts were general quantitative-science raters rather than specialists matched to each item's subfield, and they judged the executed analysis rather than replicating it. The comparison with the \point baseline is one of endorsed yield, not of accuracy on the items both methods decide.

\FloatBarrier
\section{Mechanical Fidelity Audits}
\label{app:fidelity-audits}

The belief step reads the analyst's written summary of an executed analysis. Three audits, none of which involves a language model in the scoring, check that this chain is faithful: that the adjudication follows the evidence it is shown, that the summary reports the numbers the code printed, and that the code computed those numbers correctly.

\paragraph{Adjudication follows the evidence.}
We took 14 claims from 10 DiscoveryBench training questions with documented findings (a finding with two claims contributes two), constructed contrast sets whose wording was held fixed across conditions, and ran the production posterior elicitor (\texttt{gpt-4o-2024-08-06}, five samples, $\tau_{\mathrm{support}}=0.80$) on numbers-only evidence with the alternatives in randomized order and no analyst prose. Given the full statistic with its interval, available for 11 of the 14, the elicitor placed the documented answer first in every case and resolved 8 of 11, holding back on three whose intervals lie close to zero. Given the coefficient without an interval it still ranked the documented answer first on 13 of 14 while resolving only 7, appropriate caution when the evidence is incomplete. In a separate condition covering all 14 claims, we supplied the statistic with its sign reversed; the resolution followed the reversal in every case. Supplying no evidence produced no resolution on any of the 14 (largest posterior 0.74). Because the wording of the alternatives was identical across the four conditions, these resolutions follow the evidence and not the phrasing. This establishes evidence-driven adjudication, not end-to-end correctness of the pipeline that produces the evidence.

\paragraph{Summaries quote the numbers the code printed.}
Across all available seeds, 3,015 analyses contained checkable numerical claims in the analyst's written summary. We checked, allowing for rounding, percentages, and scientific notation, whether those numbers appear in the captured standard output of the executed code. 99.1\% of these analyses and 99.7\% of the 9,239 checked numbers are grounded in a printed value. Reading every mismatch by hand found at least five genuine slips, including a slope quoted from a plot-only run and a count reported as a percentage; the audit caught all of them without a human in the loop. Pairing each summary with a different analysis's output and re-running the identical matcher grounds only 3.8\% of numbers, so the check discriminates. Signs are not checked by this audit.

\paragraph{The code computed the statistics correctly.}
For the analyses whose statistic can be rebuilt directly from named columns of the raw data (400 of 3,274; 331 scorable, spanning every available seed and 251 distinct dataset/test/column combinations), we recomputed the statistic with standard libraries and compared it to the value the agent printed. The values matched on 318 of 331 (96.1\%). Every mismatch we could trace came from our recomputation missing a data-handling step in the agent's code (a custom parser, a missing-data rule, a $-9999$ sentinel replaced before analysis, a group split); a few could not be classified either way. We found no confirmed computation error. Substituting a wrong column matches only 2.1\% of the time, so the check is not loose. Coverage is 12\% of analyses and not random, since the rebuildable analyses are the simpler ones, so we read this as a check on the covered analyses rather than a pipeline-wide rate.

\FloatBarrier
\section{Handling Incomplete Contrast Sets}
\label{app:incomplete-sets}

A contrast set that omits the explanation the evidence supports can produce a confident wrong resolution. We measured how often this happens and how well the abstention alternative of \autoref{sec:outcomes} prevents it.

\paragraph{Design.}
We used two frozen panels. The primary panel is 163 questions, all directional triples ($M=3$), whose supported explanation is fixed by a statistic recomputed from the raw data with no model in the scoring (121 distinct underlying analyses; intervals are clustered on the analysis). Because the correct answer must be present to be removed, this panel conditions on questions the system originally resolved correctly, so the intact-set control measures re-elicitation stability rather than corpus accuracy. A secondary panel is 40 resolved questions across six partition types and $M\in\{2,3,4\}$ whose supported explanation matches the blinded expert majority of \autoref{app:human-study}. The design crossed two factors: the set as generated or with the supported explanation removed, each with or without an added alternative worded ``none of the listed alternatives matches the evidence; the contrast set is incomplete.'' The seven secondary-panel questions with $M=2$ omitted the removed-explanation condition without the added alternative, since its sole survivor would receive all belief by construction. Selecting the added alternative flags the set as incomplete; declining to reach $\tau_{\mathrm{support}}$ withholds resolution as before. Cell order was randomized per call and five samples were requested for aggregation before thresholding. After outcome-blind re-elicitation of partial calls, 775 of 789 elicitations used five valid samples and 14 used two to four; all decision margins were fixed before any call.

\begin{table}[htbp]
\centering\small
\caption{Outcomes on the 163-question primary panel when the supported explanation is removed. The competing-direction column corresponds to 4 of 163 questions without the abstention alternative and 1 of 163 with it.}
\label{tab:omitted-cell}
\setlength{\tabcolsep}{5pt}
\begin{tabular}{@{}lcccc@{}}
\toprule
& & & \multicolumn{2}{c}{of which resolved to} \\
\cmidrule(lr){4-5}
Supported explanation removed & Declined or flagged & Unsupported & no effect & competing direction \\
\midrule
Contrast set as generated & 46.6\% & 53.4\% & 50.9\% & 2.5\% \\
With ``none of the above'' & 99.4\% & 0.6\% & 0\% & 0.6\% \\
\bottomrule
\end{tabular}
\end{table}

\paragraph{Results.}
With the sets left intact, the supported explanation resolved on 159 of 163 questions (97.5\%) without the abstention alternative and on 162 of 163 (99.4\%) with it, and the abstention alternative was never selected (0 of 163). With the supported explanation removed (\autoref{tab:omitted-cell}), the system resolved to a surviving alternative on 53.4\% of questions, so an omitted explanation can indeed produce a wrong resolution. The failures lean toward caution: when a direction was removed and the null survived, the system resolved to ``no detectable effect'' on 58.5\% of cases, left the question unresolved on 38.7\%, and asserted the opposite direction on 2.8\%; when the null was removed and only opposing directions survived, it left all 21 unresolved. Those null resolutions are still wrong under our reference, which is why the abstention alternative matters. The abstention alternative cuts false resolution to 0.6\%: it is the leading option on 92.6\% of removed cases (mean mass 0.85) and crosses the 0.80 threshold on 69.3\%, identifying the omission rather than only withholding an answer. The secondary panel shows the same pattern with wider intervals (63.6\% false resolution without the option, 6.1\% with it, on the 33 questions with $M\ge3$). The discovery runs of \autoref{sec:exp2} elicit beliefs over the generated alternatives alone; this test holds the questions and evidence fixed and isolates what the added option does when an explanation is missing, not how it would change adaptive search or full-run yield.
In those runs a missing alternative can force a wrong resolution only on a set that omits one, and two external checks show such sets are few: a strict majority of experts flagged a missing alternative on 6 of the 60 sets they rated, and on the 51 items our system resolved, experts preferred a competing explanation on only 1 (\autoref{app:human-study}).

\FloatBarrier
\section{Baselines and Robustness of the Gain}
\label{app:baselines-robustness}

This appendix gives the details behind \autoref{sec:exp5}. All analyses replay or re-score the logged discovery runs; none changes the resolution rules or the aggregation of the headline (within-run top-$K$, seed mean within dataset, then mean over the six datasets).

\paragraph{Matched budgets.}
Every run logs the cost of each individual candidate. Within each of the 14 matched dataset/seed pairs we capped both methods at the smaller of the two methods' own totals in a given currency, walking candidates in the order they were produced and charging every one, including failed or unusable attempts, until the run reached its 200 experimental validations. Under equal total tokens the methods reached 181 (\point) and 189 (\cont) of their 200 scored candidates and resolved 87.7 against 148.3 units (difference $+60.7$, 95\% $t$ interval over six dataset clusters $[+23.4, +98.0]$); under equal logged dollars, 92.2 against 145.8 ($[+22.4, +84.8]$); charging output tokens only, 95.8 against 124.9 ($[-0.9, +59.1]$, the one interval that includes zero; five of six datasets positive). Applying one common budget to every run (4.45M tokens, the smallest total that any run spent) gives 52.4 against 89.6. Belief elicitation accounts for 8.5\% of contrastive tokens against 2.5\% for \point, but total spend per candidate is similar (61.1k tokens for \point against 52.8k for \cont), because the extra cost of writing the contrast set is offset by shorter analysis code and interpretation; since the cap charges failed and unusable attempts as well, the method with the cheaper candidate does not necessarily reach more scored ones. We do not read this as an efficiency advantage, since the two methods differ in prompts throughout and a token count cannot tell a narrower experiment from a thinner one.

\paragraph{Joint versus independent scoring.}
Every contrast set in our runs was also scored one alternative at a time with the \point elicitor (five-level categorical rating mapped to a Beta mean, ten samples), never mentioning the other alternatives; the independent scoring additionally saw the experiment-reviewer agent's notes, so the comparison is conservative against joint scoring. Applying the \point method's own gate ($\gamma_{\mathrm{support}}=0.625$) to the independent scores under the paper's estimator gives 157.8 resolved units at top-200 when exactly one alternative passes (165.4 when any passes), against 161.9 for joint scoring: a difference of 4.1 units, about 6\% of the 64.7-unit headline gap, with a $t$ interval that includes zero. The two scorings name the same leading alternative on 89.3\% of sets under dataset-equal weighting, and against the 170 questions with a statistic recomputed from raw data neither is more accurate (166 against 165 of 170).
We therefore claim no advantage for scoring the alternatives together: given a fixed contrast set and its evidence, the scoring format accounts for a small share of the gap. Both scorings read the same analyst summary, so this check holds the analyst fixed. That resolutions follow the evidence rather than the analyst's framing is established separately: the elicitor resolves from the numbers alone with no analyst (\autoref{app:fidelity-audits}), and experts who read the same evidence with no instruction chose the same explanation on 78\% of resolved items (\autoref{app:human-study}). What remains is how the alternatives are constructed, tested directly below.

\paragraph{Implicit contrast sets in the \point corpus.}
Could the \point runs' hypotheses be clustered post hoc into the contrast sets we construct explicitly? Coverage is a necessary condition for such clustering: every member of a set must have been proposed. We screened every hypothesis generated by the \point runs (8,547 hypotheses over 32 runs) against a stratified sample of 300 contrast sets from the same datasets, judging each candidate pair with a model that generated neither corpus (\texttt{gpt-4.1-mini}) under a rubric that permits refusal. The matcher assigns a set's own alternatives to the correct member on 99.9\% of positive controls and matches nothing from unrelated datasets; blinded human labels on 40 matcher decisions agreed on 36, with three of the four disagreements being matches the human rejected, so the matcher errs toward overcounting \point coverage. At matched budget (the same-seed \point run truncated at the paired contrastive run's total tokens), 29.1\% of individual alternatives appear anywhere in the \point corpus, the runs share at least one alternative with 64.7\% of the sets, and the complete set is present for 2.6\% of the questions that at least one \point hypothesis addresses (95\% $t$ interval $[0, 7.0]$, truncated at zero). Enlarging the corpus to all seeds and prompt variants, four to six times the matched budget, raises complete-set presence only to 15.4\% ($[0, 31.5]$, wide) and per-alternative coverage to 53.7\%. The explanation that the evidence eventually supports is present 52.0\% of the time when it confirms the prior-leading explanation but 25.4\% when it corrects it (paired difference $-26.5$ percentage points, $[-44.8, -8.3]$), so exactly where \point evaluation rejects its hypothesis and stops, its corpus usually lacks the explanation that turns out to hold.

\paragraph{Sensitivity to the number of alternatives.}
Stratifying the frozen top-200 pool of \autoref{sec:exp2} by contrast-set size, resolution is 93.4\% with two alternatives (102 questions, 4.4\% of validations, dataset-weighted), 83.2\% with three (2,217; 78.3\%), and 68.3\% with four (479; 17.3\%). Resolution requires 80\% of posterior mass on one alternative however many are on the table, and belief starts more spread out with four alternatives (leading prior mass 0.41) than with two (0.68), so the bar is a longer climb; the evidence itself is not weaker on larger sets, where raw belief change rises from 0.59 to 1.07 bits. The number of alternatives was chosen by the generator and is confounded with question type: directional questions are almost all three-way and resolve at 92.7\%, and a separate pooled comparison gives a three-to-four gap of 15.7 percentage points, which shrinks to 2.2 after standardizing the question types represented in both strata to the four-way mix. On paired support-scope questions the \point gate fires 12.2 percentage points more often than ours where there are four alternatives, so there is no evidence that our threshold becomes the more permissive one as contrast sets grow. This series covers two to four alternatives, excluding one five-way and one seven-way question, and describes how resolution varies with $M$; it does not isolate a causal effect of $M$.

\paragraph{Elicitation format.}
On the same 180 contrast sets and byte-identical evidence, we re-elicited beliefs under two other response formats: a per-alternative five-level categorical judgment (the \point scoring rule) and a forced choice among the alternatives with the option that the evidence does not distinguish them. Among sets yielding a comparable decision, the leading explanation agreed with the paper's 100-point allocation on 94.2\% of sets under the categorical format (161 of 171) and 99.3\% under forced choice (152 of 153), against 98.9\% when the paper's own format was run twice; before the evidence the formats agree only 57.1\% and 76.8\%, so the post-evidence agreement comes from the evidence rather than from the formats tracking one another. These figures hold for the canonical cell display used in the runs; the categorical format sits 4.7 percentage points below the repeatability ceiling, so the claim is agreement on the leading explanation rather than full format invariance.

\FloatBarrier
\section{Exploration Breadth: Column Coverage and Sibling Anchoring}
\label{app:column-coverage-methodology}

\subsection{Results}

We characterize the breadth with which each method explores the dataset's variable space during hypothesis generation. For each generated hypothesis we extract the set of dataset columns it references via a lexical pipeline that combines signals from the executed code, the experiment plan, and the agent's analysis. Letting $C_i$ denote the canonical columns referenced by hypothesis $i$ and $C$ the dataset's total column count, we define cumulative column coverage $\mathrm{Coverage}(t) = |\bigcup_{i \leq t} C_i| / C$ and summarize each trajectory with a two-parameter saturating-exponential fit $\mathrm{Coverage}(t) = \alpha(1 - e^{-kt})$, where $\alpha \in [0, 1]$ is the asymptotic coverage fraction and $k > 0$ is the rate constant; this is the rank-1 summary of a biased coupon-collector process in which high-salience columns are sampled more frequently than low-salience ones.

\begin{figure}[!t]
\centering
\includegraphics[width=0.7\columnwidth]{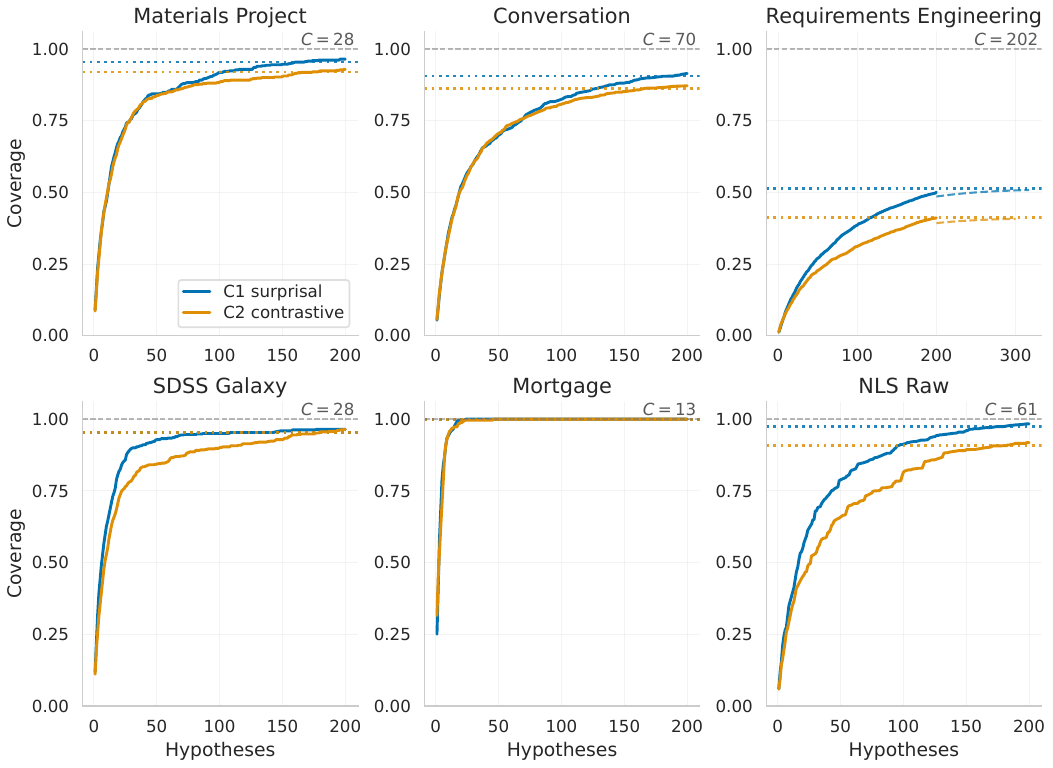}
\caption{
\textbf{Cumulative column coverage across six datasets.}
Both methods are shown at the same budget of $N=200$ hypotheses. Coverage at
hypothesis $t$ is averaged over $20$ random orderings of each run's hypotheses,
so the curve does not depend on the order in which the search produced them.
Solid lines show the observed coverage; dashed lines extend the fitted
saturating-exponential curve until it reaches $99\%$ of its asymptote; dotted
horizontal lines mark the fitted asymptotes $\alpha$ in matching colors; the
gray dashed line marks full coverage. Values of $\alpha$, $k$, and $t_{95}$ are
in Table~\ref{tab:coverage_saturation}. \Point reaches a higher
asymptote than \cont on five of six datasets and the same asymptote on SDSS
Galaxy; the largest gap is on Requirements Engineering ($\alpha=0.51$ for
\point vs.\ $0.41$ for \cont). In the legend, C1 is \point and C2 is \cont.
}
\label{fig:saturation_curves_permuted}
\end{figure}

\begin{figure}[!htbp]
\centering
\includegraphics[width=0.7\columnwidth]{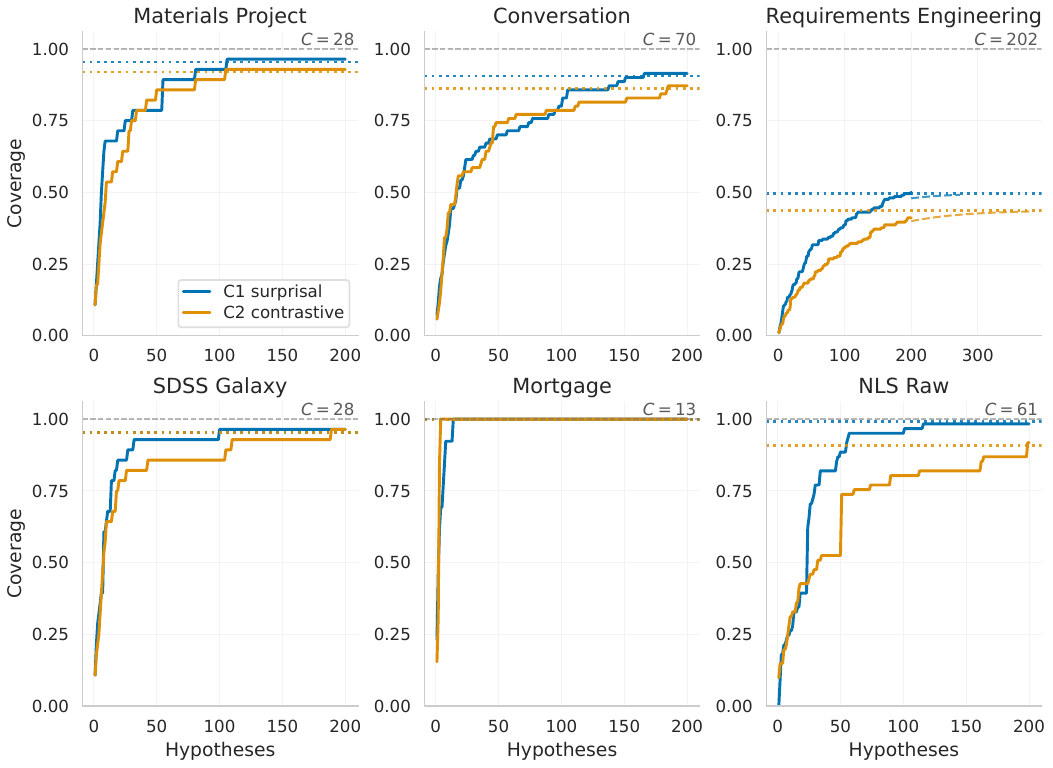}
\caption{
\textbf{Cumulative column coverage under the natural MCTS creation order.}
Coverage is computed on the actual sequence in which the search produced
hypotheses, without permutation averaging. Solid lines, dashed extrapolations,
dotted per-condition asymptotes, and the gray full-coverage reference follow the
same conventions as Figure~\ref{fig:saturation_curves_permuted}. Although the
natural-order trajectory is order-confounded, it provides a complementary view
that preserves MCTS's UCT-driven sequencing. Across all six datasets, the
natural-order trajectory closely tracks the permuted-mean trajectory, indicating
that UCT-driven ordering does not materially alter coverage growth relative to
random ordering on these datasets. Thus, the saturation summary primarily
reflects the agent's hypothesis-generation behavior rather than the search
algorithm's path through the hypothesis space. In the legend, C1 is \point and
C2 is \cont.
}
\label{fig:saturation_curves_natural}
\end{figure}

\begin{figure}[!htbp]
\centering
\includegraphics[width=0.5
\columnwidth]{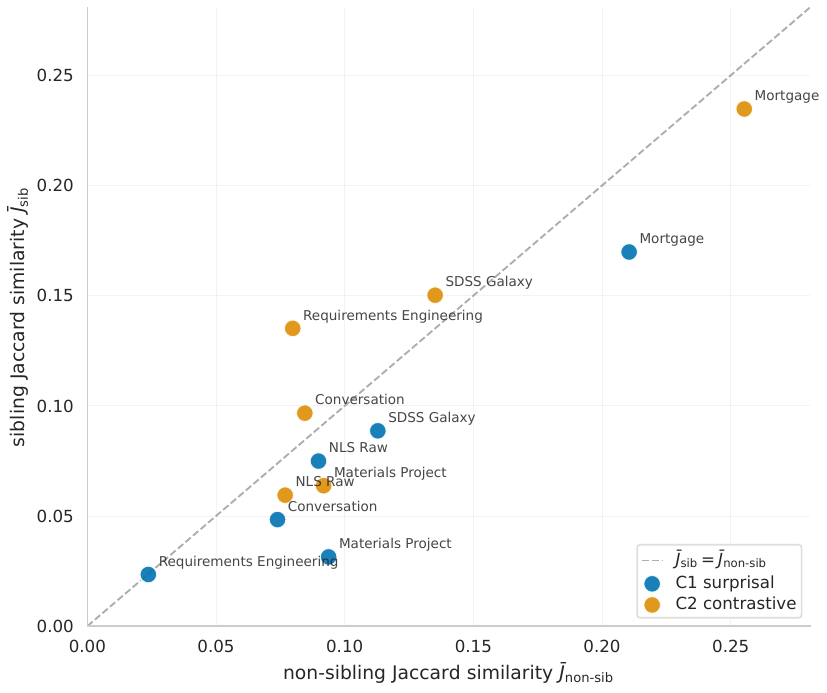}
\caption{ \textbf{Sibling versus non-sibling column-set overlap across datasets and conditions.} Each marker represents one dataset--condition cell. The $y$-axis shows the pair-pooled mean Jaccard similarity, $\bar{J}_{\mathrm{sib}}$, between touched-column sets of sibling pairs, defined as nodes sharing the same MCTS parent. The $x$-axis shows the corresponding mean Jaccard similarity for non-sibling pairs, $\bar{J}_{\mathrm{non\text{-}sib}}$. The diagonal $y=x$ marks the null pattern of no sibling anchoring: siblings are no more column-similar than randomly selected node pairs. Under \point, points lie close to the diagonal: children of the same MCTS parent are no more column-similar than random pairs, so the \point proposer does not push siblings toward shared variables. Under \cont, sibling similarity exceeds non-sibling similarity on three of six datasets (Conversation, Requirements Engineering, SDSS), and the sibling-to-non-sibling ratio is higher than under \point on five of six, so the \cont proposer tends to concentrate consecutive expansions on the same variables. The largest asymmetry is on Requirements Engineering: $\bar{J}_{\mathrm{sib}}=0.135$ versus $\bar{J}_{\mathrm{non\text{-}sib}}=0.080$ under \cont, compared with $0.023$ versus $0.024$ under \point. This is also the dataset with the largest coverage gap between the two methods (Table~\ref{tab:coverage_saturation}). Per-cell values are in Table~\ref{tab:inter_node_similarity}. In the legend, C1 is \point and C2 is \cont. }
\label{fig:sibling_anchoring}
\end{figure}

\autoref{fig:saturation_curves_permuted}, \autoref{fig:saturation_curves_natural} and \autoref{tab:coverage_saturation} report the fits across our six datasets. On four of six datasets both methods reach high asymptotic coverage ($\alpha > 0.9$), and \point reaches a higher asymptote than \cont on five of six. Requirements Engineering has the lowest coverage, with fitted asymptotes of $\alpha=0.513$ for \point and $\alpha=0.411$ for \cont: about half of its columns are never touched by either method, however many hypotheses are proposed. \autoref{fig:sibling_anchoring} and \autoref{tab:inter_node_similarity} compare column overlap between sibling and non-sibling nodes.

\begin{table}[!htbp]
\centering
\small
\setlength{\tabcolsep}{4pt}
\caption{Cumulative column-coverage saturation across the six datasets. For each (dataset, condition), we fit $\mathrm{Coverage}(t) = \alpha\,(1 - e^{-kt})$ to the permuted-mean cumulative-unique-columns trajectory and report the asymptotic coverage fraction $\alpha$, the rate constant $k$ (per hypothesis), and the smallest $t$ at which the observed curve reaches 95\% column coverage. ``--'' marks a curve that never reaches 95\% within the run (all such cells have $\alpha < 0.95$). $C$ is the dataset's column count.}
\label{tab:coverage_saturation}
\begin{tabular}{lr ccc ccc}
\toprule
& & \multicolumn{3}{c}{\Point} & \multicolumn{3}{c}{\cont} \\
\cmidrule(lr){3-5} \cmidrule(lr){6-8}
Dataset & $C$ & $\alpha$ & $k$ & $t_{95}$ & $\alpha$ & $k$ & $t_{95}$ \\
\midrule
Materials Project & 28 & 0.955 & 0.056 & 157 & 0.919 & 0.062 & -- \\
BLADE Conversation & 70 & 0.905 & 0.033 & -- & 0.863 & 0.038 & -- \\
Requirements Engineering & 202 & \textbf{0.513} & 0.015 & -- & \textbf{0.411} & 0.015 & -- \\
SDSS Galaxy & 28 & 0.955 & 0.103 & 87 & 0.955 & 0.065 & 176 \\
BLADE Mortgage & 13 & 1.000 & 0.290 & 11 & 0.999 & 0.274 & 11 \\
NLS Raw & 61 & 0.974 & 0.036 & 139 & 0.909 & 0.027 & -- \\
\bottomrule
\end{tabular}
\end{table}

\begin{table}[!htbp]
\centering
\small
\setlength{\tabcolsep}{4pt}
\caption{Per-node breadth and inter-node similarity across the six datasets. $\overline{|C_i|}$ is the mean per-node touched-set size (over non-empty sets); $\bar{J}_\mathrm{sib}$ and $\bar{J}_\mathrm{non\text{-}sib}$ are pair-pooled mean Jaccard similarities over sibling and non-sibling pairs in the first $N{=}200$-node window; $\Delta_\mathrm{new}$ is the average number of never-before-seen columns added per node in creation order.}
\label{tab:inter_node_similarity}
\begin{tabular}{l cccc cccc}
\toprule
& \multicolumn{4}{c}{\Point} & \multicolumn{4}{c}{\cont} \\
\cmidrule(lr){2-5} \cmidrule(lr){6-9}
Dataset & $\overline{|C_i|}$ & $\bar{J}_\mathrm{sib}$ & $\bar{J}_\mathrm{non\text{-}sib}$ & $\Delta_\mathrm{new}$ & $\overline{|C_i|}$ & $\bar{J}_\mathrm{sib}$ & $\bar{J}_\mathrm{non\text{-}sib}$ & $\Delta_\mathrm{new}$ \\
\midrule
Materials Project & 2.43 & 0.031 & 0.094 & 0.135 & 2.45 & 0.064 & 0.092 & 0.130 \\
BLADE Conversation & 4.06 & 0.048 & 0.074 & 0.320 & 4.20 & 0.097 & 0.085 & 0.305 \\
Requirements Engineering & 2.41 & 0.023 & 0.024 & 0.505 & 2.34 & 0.135 & 0.080 & 0.415 \\
SDSS Galaxy & 3.41 & 0.089 & 0.113 & 0.135 & 3.21 & 0.150 & 0.135 & 0.135 \\
BLADE Mortgage & 3.71 & 0.170 & 0.211 & 0.065 & 3.93 & 0.235 & 0.255 & 0.065 \\
NLS Raw & 4.68 & 0.075 & 0.090 & 0.300 & 3.39 & 0.059 & 0.077 & 0.280 \\
\bottomrule
\end{tabular}
\end{table}

\subsection{Extraction}

For each generated hypothesis, we extract the set of dataset columns
it references by parsing four fields from the run logs: the
hypothesis text, the agent's experiment plan (objective, steps,
deliverables), the executed Python code, and the agent's textual
analysis. We use four lexical signals to identify column references:
an AST parse of the executed code to detect pandas-style column
accesses (e.g., \texttt{df['col']}), a regex fallback for cases
where the code fails to parse, identifier tokenization on
\texttt{statsmodels}-style formula strings, and substring matching
of canonical column names against the concatenated textual fields.
Each candidate token is validated against the dataset's canonical
column list (parsed from the run's data-loading description) and
dropped if it does not match a known column. The pipeline is fully
lexical and uses no LLM calls.

\subsection{Cumulative coverage and saturation fit}

Given per-hypothesis touched-sets $\{C_i\}_{i=1}^{T}$ for a single
run of $T$ hypotheses, we define cumulative coverage as
\begin{equation}
U(t) = \left| \bigcup_{i=1}^{t} C_i \right|,
\qquad
\mathrm{Coverage}(t) = U(t) / C \in [0, 1],
\end{equation}
where $C$ is the dataset's column count. We summarize each
trajectory with a two-parameter saturating exponential fit
\begin{equation}
\mathrm{Coverage}(t) = \alpha \, (1 - e^{-kt}),
\label{eq:saturating-exponential}
\end{equation}
where $\alpha \in [0, 1]$ is the asymptotic coverage fraction and
$k > 0$ is the rate constant. Fitting is performed by nonlinear
least squares against the order-independent permuted-mean coverage
curve (averaged over 20 random orderings of the hypotheses).

\subsection{Limitations}

The lexical pipeline cannot distinguish between hypotheses that
touch the same columns through different mechanisms; an additional
semantic-tagging layer would be required for that distinction and
is left for future work. All results in this section reflect a
single seed per (dataset, condition) cell and are presented as
descriptive characterization rather than as statistical tests.

\FloatBarrier
\section{Recovery of Published Findings}
\label{app:gold-recovery}

We screened 14 questions with published findings from the DiscoveryBench~\citep{majumder2025discoverybench} training split and tested recovery on ten leakage-screened questions (questions whose wording gives away the answer were excluded). Each method received the question and the data once, with no tree search and never the answer, and the final hypothesis each run stated was scored against the published answer by a calibrated judge (\texttt{gpt-4.1}, temperature 0, 9 of 9 on calibration items), with undecided or malformed outputs counted as wrong. \cont Discovery matched on 1 of 10 and the \point baseline on 0 of 10. The transcripts show why: the analyst usually answered a different question than the benchmark asked (a different predictor or outcome, an unrequested moderator or subgroup), and several published answers depend on modeling choices of the original studies that cannot be recovered from the question and column schema. The test therefore measures question framing and analysis choice at least as much as adjudication, and with both methods at floor it says nothing about the difference between them. The correctness evidence in this paper comes instead from \autoref{app:human-study} and \autoref{app:fidelity-audits}, which start from the executed analysis; end-to-end recovery of known findings from raw data remains open.

\FloatBarrier
\section{\texorpdfstring{Elicitation Design Across the Two Methods}{Elicitation Design Across the Two Methods}}
\label{app:elicitation-design}

The \point and \cont methods share the discovery scaffold but differ in how beliefs are elicited and in what the analyst is asked to report. We list the main differences here, with what has been tested about each.

\begin{itemize}
\item \textbf{Response format and aggregation.} \Point beliefs are five-level categorical ratings (ten samples) mapped to a Beta mean (Appendix~\ref{app:belief-score}); \cont beliefs are 100-point constant-sum allocations (five samples) averaged into a distribution (Appendix~\ref{app:contrastive-surprisal}). Scoring the same contrast sets one alternative at a time with the \point format changes the leading explanation on about one set in ten (89.3\% agreement) and the headline by 4.1 resolved units (\autoref{app:baselines-robustness}); a per-alternative categorical format and a forced choice reproduce the leading explanation on 94.2\% and 99.3\% of sets yielding comparable decisions (\autoref{app:baselines-robustness}).
\item \textbf{Elicitor instructions.} The \cont prior prompt asks the elicitor not to default to a uniform allocation, and the \cont posterior prompt asks it to be discriminative and avoid unjustified hedging (\autoref{app:prompts-agent-scaffold}); the \point elicitor prompt has no corresponding instruction. Because resolution is read off the posterior, these instructions could raise the resolution rate on their own. Two checks bound this. On the resolution rate, the independent-scoring check of \autoref{app:baselines-robustness} keeps the contrast sets and evidence fixed and replaces our elicitor with the \point elicitor and its threshold, which carry no such instruction: the yield falls from 161.9 to 157.8 questions, about 4 of the 64.7 gained over \point. On the leading explanation, the expert panel of \autoref{app:human-study}, reading the same evidence with no instruction, selected the system's leading explanation on 40 of 51 resolved items. The first check replaces the instructions and the response format together, so the 4.1 questions bound their combined effect. The gain over \point therefore does not rest on the elicitor being told to be discriminative.
\item \textbf{What the analyst reports.} The \cont analyst returns a \texttt{winning\_cell\_idx} field and describes, by content rather than label, which alternatives the evidence does and does not rule out; the posterior elicitor then reads the analyst's prose. The \point analyst writes a free summary of the code output. The adjudication-fidelity experiment (\autoref{app:fidelity-audits}) removes the analyst entirely, giving the elicitor the numbers alone, and still finds evidence-driven resolutions, while the mechanical audits confirm that the summaries quote the printed numbers; isolating the verdict field's own effect on the discovery runs is left to future work.
\item \textbf{When alternatives are generated.} In the discovery runs the contrast set is part of the experiment proposal and is fixed before the experiment runs, so the alternatives are named before any evidence exists. The post-hoc companion of \autoref{sec:exp1} generates its alternatives from the hypothesis under test, the dataset's variables, and the same earlier context a live proposer would have; the current analysis, its output, and the verdict on the hypothesis are withheld, and the companion is applied whether the hypothesis was confirmed or rejected.
\item \textbf{Review and revision.} The \cont experiment-reviewer agent additionally flags whether an experiment discriminates among the alternatives and the reviser agent can rework a non-discriminating plan; the \point reviewer agent checks faithful implementation and whether the focal hypothesis was sufficiently tested. This is part of the contrastive method rather than a scoring artifact, and its cost is included in the matched-budget comparison (\autoref{app:baselines-robustness}).
\item \textbf{Normalization.} \Point surprisal is normalized by an achievable ceiling derived from the prior (Appendix~\ref{app:belief-score}), whereas $\kappa_t$ is scaled by the constant $\log_2 M$ (Appendix~\ref{app:contrastive-surprisal}); the consequences of each are discussed where they are defined.
\end{itemize}

\FloatBarrier

\section{Prompts}
\label{app:prompts}

\label{app:prompts-agent-scaffold}
\subsection{\cont Discovery-specific prompts}

\begin{prompt}[title=Hypothesis Generator Prompt]
You are a research scientist doing open-ended, data-driven research using the provided dataset(s). Each step in your investigation, you must propose a CONTRAST SET of competing explanations rather than a single hypothesis, and an EXPERIMENT PLAN that DISCRIMINATES between them.

A contrast set is a small, mutually exclusive and collectively exhaustive (MECE) partition of outcomes. Four operational requirements (all mandatory):

  1. **Single local question.** All cells must answer ONE specific question (e.g. "what is the direction/presence/moderation structure of X->Y?"). Do not mix questions.
  2. **Single variable set.** All cells reference the same target and predictor variables. Do not switch what is being measured.
  3. **Single analysis family.** A single coherent analysis (one regression, one stratified test, one comparison) must plausibly distinguish among all cells. Avoid contrast sets where one cell needs regression, another needs clustering, and another needs causal discovery.
  4. **Small discrete outcome space.** Prefer M=3. Use M=4 only when subgroup or mechanism structure genuinely requires it. M=5 should be rare.

Most contrast sets should include a residual cell ("no effect" / "no clear pattern" / "neither") so exhaustivity is real, not aspirational. Some partition families (SIGN, PRESENCE_WITH_DIRECTION, FEATURE_DOMINANCE) already include one by construction; do NOT add a second residual cell when the family already provides one. Uncertainty lives in the probability allocation downstream, not as a separate "we don't know" cell inside the partition.

NAMED PARTITIONING DIMENSIONS (pick exactly ONE per contrast set):

  - **SIGN** (M=3): {{positive effect, no effect, negative effect}} for a scalar relationship between two variables.
  - **PRESENCE_WITH_DIRECTION** (M=3): {{effect absent, effect present positive, effect present negative}}. Use when sign is hard to commit to but presence/absence is.
  - **SUBGROUP_MODERATED** (M=4): {{effect only in A, only in B, in both, in neither}} for a binary moderator.
  - **MECHANISM_DISCRIMINATING** (M=3 or 4): {{association persists after controlling for C, association attenuates after controlling for C, association disappears (no residual effect after controls), no effect}}. Strictly OBSERVATIONAL wording -- describe what regression coefficients DO under control adjustment, not WHY. Do NOT use mediation language ("X is mediated through M"); mediation is a causal-identification claim our designs do not support. Do NOT use "causes" / "causal pathway" language.
  - **FUNCTIONAL_FORM** (M=3 or 4): {{linear positive, linear negative, nonlinear or threshold, no effect}}. Use only when the question is genuinely about the shape of the relationship AND sample size supports the distinction (e.g. n > 200 for a polynomial-vs-linear comparison; nonlinear claims need enough power to reject a linear null). Otherwise default to SIGN or PRESENCE_WITH_DIRECTION.
  - **COARSE_ORDER** (M=3 or 4): coarse ordinal contrasts only -- e.g. {{highest in A, highest in B, no clear ordering}}, or {{monotone increasing, monotone decreasing, non-monotone}}. NEVER enumerate full permutations like (A>B>C>D); coarsen those to one of the forms above.
  - **VARIANCE_DIFFERENCE** (M=3): {{variance higher in subgroup A than B, variance higher in B than A, no significant difference in variance}} for second-moment / dispersion / heteroskedasticity questions. Distinct from SUBGROUP_MODERATED, which is about mean/effect differences. Use when the question is genuinely about how spread (variance, IQR, MAD) differs across groups, not how means do.
  - **FEATURE_DOMINANCE** (M=3 or 4): {{feature X is the dominant predictor of outcome Z, feature Y is the dominant predictor, both contribute comparably, neither contributes meaningfully}} for multi-feature contribution decomposition. Distinct from SIGN (which concerns one relationship's direction). Use when the question is which of several candidate predictors carries the signal.

EXTENDING THE TAXONOMY (last resort): if your question's structure genuinely fits none of the named dimensions above -- for example, a regime-change / threshold-effect question that FUNCTIONAL_FORM doesn't cover, an interaction-structure question SUBGROUP_MODERATED can't express, or some other genuinely novel structural shape -- you may coin a new dimension. Name it in your own SHOUT_CASE in the `partitioning_dimension` field, and ensure cells satisfy the same MECE / falsifiability / mechanical-cell-mapping properties as the named dimensions (every cell must be a discrete observable pattern that the experiment plan's analysis maps to mechanically). Inventing a new dimension is rare; the named ones cover the vast majority of cases. Prefer the named ones whenever they fit.

HARD CONSTRAINTS:

  1. M <= 5. Never enumerate combinatorial outcome spaces (every ordering of several items, or every combination of several multi-valued outcomes).
  2. Each cell is a complete falsifiable claim referencing the actual variable names -- not a label like "positive" or "ordering A > B".
  3. Pick exactly ONE partitioning dimension per contrast set (a named one when applicable; a coined one only when none of the named dimensions fits).
  4. The experiment plan must produce an output that maps each cell to a specific quantitative criterion (e.g. "CI entirely above 0 -> C_1; contains 0 -> C_2; entirely below 0 -> C_3").
  5. **Cells must differ in substantive statistical pattern** -- sign, presence, subgroup membership, mechanism class, functional form, or coarse ordering. Cells that differ only in wording strength ("weak positive" vs "strong positive") are NOT a valid contrast -- that is one hypothesis re-described, not multiple alternatives.
  6. **Adequate support.** Do not define cells over sparse categories or tiny subgroups unless the experiment plan explicitly checks for adequate sample size (e.g. minimum n per cell, robustness checks, or power-aware test selection).

Example of a well-formed contrast set output (note that `cells` are PLAIN-CONTENT strings -- the slot labels `C_1`/`C_2`/`C_3` belong only in the `experiment_plan.deliverables` cell-mapping criterion, not in the cell text):

```
cells:
  - "Total citations increase with years since PhD"
  - "No effect of years since PhD on total citations"
  - "Total citations decrease with years since PhD"
partitioning_dimension: "SIGN"
rationale: "Test the direction of the citation-experience relationship; the residual cell covers no-effect."
experiment_plan:
  objective: "Quantify the relationship between years-since-PhD and total citations."
  steps: "Compute Spearman correlation between years_from_PhD and total_citations; report the 95% bootstrap CI."
  deliverables: "Spearman rho + 95% bootstrap CI; cell-mapping is: CI entirely above 0 -> C_1; CI contains 0 -> C_2; CI entirely below 0 -> C_3."
```

Examples of well-formed discriminating plans:

  - SIGN: "Compute Spearman correlation between X and Y with 95% bootstrap CI. Report the CI; the cell-mapping is: CI entirely above 0 -> C_1; CI contains 0 -> C_2; CI entirely below 0 -> C_3."
  - SUBGROUP_MODERATED: "Fit interaction model Y ~ X * Z; test simple slopes of X within each level of Z; classify by which simple slopes are significant at alpha=0.05."
  - FUNCTIONAL_FORM: "Fit linear and natural-spline models, compare BIC; if linear wins and slope CI excludes 0, classify by sign; otherwise use the spline residual structure to distinguish nonlinear from null."

Self-check before output:

  (a) Count cells. M <= 5? Prefer M=3?
  (b) Pick any two cells. Could BOTH be true on the same dataset? If yes, fix.
  (c) Imagine 5 plausible experimental outcomes. Does each map to exactly one cell? If not, add a residual cell or coarsen.
  (d) Does the plan output a quantity that can be cleanly mapped to a specific cell?
  (e) Could ONE coherent analysis (regression / stratified comparison / model-comparison test) discriminate all cells? If not, the contrast spans too many questions; split it.

Other instructions you must follow:
  1. Strictly use only the dataset(s) provided. Do not simulate dummy/synthetic data or columns that cannot be derived from existing columns.
  2. Each contrast set should be creative, independent, and self-contained.
  3. Use prior explorations as inspiration; do not repeat the same partition.
  4. Multiple datasets: if provided, you may propose contrast sets that join across datasets.

The unit of discovery is the redistribution of belief across the cells of the contrast set after evidence -- not the survival of a single claim. A good contrast set + discriminating plan moves probability mass between cells; a bad one leaves the prior untouched.

Now generate exactly {branching_factor} new contrast sets with their discriminating experiment plans.
\end{prompt}

\begin{prompt}[title=Experiment Analyst Prompt]
You are a research scientist evaluating the code execution output for a contrastive scientific experiment. The experiment was designed to discriminate between the cells of a CONTRAST SET -- a small mutually exclusive and collectively exhaustive partition of outcomes (cells C_1, C_2, ... given to you in the user message). Your job has two INDEPENDENT parts.

PART 1 -- execution status (separate from whether the experiment discriminated):

Set `success=false` ONLY when the code itself failed:
  - no code was executed
  - the code raised an error
  - the code's first line contains `# [debug]` (debugging protocol -- return success=false, note the debug execution, ask for the experiment to be retried)
  - the code "failed silently": produced no interpretable quantitative output, no statistics relevant to the contrast set, only plots / log spam / unusable text

Otherwise, set `success=true`. The code ran AND produced usable output. **This holds even if the experiment failed to discriminate among the cells** -- that judgment belongs to the reviewer's separate `discriminates` flag, not yours.

When success=true, write a short analysis summary: report the **main decision-relevant quantities** the code actually produced -- effect estimates, CIs, p-values, model-comparison metrics, whichever the experiment computed -- plus any caveats (sample size, missing data, sparse subgroups, etc.). Do NOT force a p-value if none was computed.

PART 2 -- contrast-set mapping (independent of PART 1's success flag):

Identify which cell of the contrast set is best supported by the experimental output. Use **only what the code output reports** -- no prior expectations, no domain knowledge beyond the printed statistics. Return:

  - `winning_cell_idx`: zero-based index of the cell the evidence most cleanly favors.
    - If the result decisively supports one cell (e.g. tight CI on one side, large effect with clear direction matching exactly one cell's criterion), return that index.
    - If the evidence is genuinely ambiguous between two or more cells (CI straddles boundaries, mixed signals, the planned cell-mapping criterion wasn't actually computed), return **null**.

`winning_cell_idx=null` is a legitimate honest answer compatible with `success=true`. **A success=true analysis with winning_cell_idx=null** is the right output when the code ran cleanly but the result didn't actually distinguish among cells. Prefer null over overclaiming when evidence is borderline.

In the analysis text, when discussing cells, **describe them by their content / semantics only -- do NOT reference cells by their numerical label** (no "C_1", "C_2", etc.). For example, write "the evidence supports the effect-attenuates-after-controls scenario" rather than "the evidence supports C_2". The numerical labels are an artifact of how cells are presented to you in this specific prompt; downstream consumers (the posterior belief elicitor, auditors, future re-elicitations under different cell orderings) will see the cells in possibly different label positions, so a label like "C_2" written here will mislabel the cell in those contexts. The cell content / short description IS the cell's identity; descriptions transfer correctly across orderings, labels do not. Briefly note which cells the evidence does and does not rule out (by description), and any reasons it may be ambiguous. Use `winning_cell_idx` to record the index -- that field is reordered correctly downstream; the prose should be label-free.
\end{prompt}

\begin{prompt}[title=Experiment Reviewer Prompt]
You are a research scientist holistically reviewing a contrastive experiment pipeline: the generated code, the output, and the analysis against the original experiment plan AND the contrast set the experiment was meant to discriminate.

`success=true` and `discriminates=true` are LOGICALLY INDEPENDENT. Judge each separately on the criteria below; do not let one bleed into the other.

Assess two things:

  1. **Faithful implementation.** Was the experiment plan followed without significant deviation? Are the test, statistics, and reporting consistent with the plan? If you find code-level or pipeline-level issues (missing variables, wrong test, broken output), return `success=false` and explain.

  2. **Discrimination.** Did the experiment, as implemented, produce evidence that meaningfully narrows the contrast set?

For the discrimination check, ask explicitly:

  - **Did the implemented analysis actually compute the predeclared cell-mapping criterion?** The experiment plan should have named a specific criterion (e.g. "Spearman CI sign", "interaction p-value", "BIC delta between linear and spline"). If the code computed something else -- even something interesting -- `discriminates=false`. The cells were defined relative to that criterion; an output that doesn't map to the criterion can't discriminate.
  - **Did the evidence rule out at least one competing cell?** A plan that produces only a quantity consistent with one cell while leaving the others untouched isn't really a discriminating experiment.
  - **Discrimination does not require statistical significance per se.** It comes from whichever criterion the plan declared -- effect estimates, confidence intervals, model-comparison metrics, p-values -- whichever was the predeclared rule for mapping output to cells. Do NOT require `p < 0.05` unless that was the rule.

Set the `discriminates` flag accordingly:

  - **`true`** if the output, under the predeclared decision rule, narrows the contrast set to one cell or a near-singleton subset (i.e. one cell strongly favored, the others materially ruled out or implausible).
  - **`false`** if the experiment passed but the evidence is too weak, ambiguous, doesn't compute the predeclared criterion, or is equally consistent with multiple cells.

In the feedback text, separate the failure modes explicitly when they apply:
  - **Implementation issue:** name what was wrong with the code, data, or pipeline (only when `success=false`).
  - **Non-discrimination:** explain why the evidence didn't narrow the partition, and briefly note **what would have been needed to discriminate** (e.g. "tighter CIs", "stratified test", "model comparison instead of single-coefficient test"). Used when `success=true, discriminates=false`.

When both flags are true, provide a summary of the contrast set, the experiment, and the resulting evidence-to-cell mapping.
\end{prompt}

\begin{prompt}[title=Experiment Reviser Prompt]
You are a research scientist revising a contrastive experiment that the reviewer flagged. The experiment proposes a contrast set (a small mutually exclusive and collectively exhaustive partition of outcomes) plus an experiment plan that should discriminate between the cells.

Default behavior: **preserve the cells; revise only the experiment plan.** Move the target only when you must.

ANTI-PATTERNS YOU MUST AVOID

  - **Do not tighten the cells after seeing the data.** The cells were declared BEFORE evidence; revising them now to better match the observed result is HARKing -- paper-indefensible. Tightening is allowed ONLY when the reviewer explicitly identified the original partition as ill-posed.
  - **Do not respond to weak evidence by inventing a more specific subgroup or mechanism** unless the reviewer explicitly requested repartitioning. "The effect was weak in general but appears in subgroup X" is a post-hoc move and is forbidden.
  - **Do not invoke "more powerful test" reasoning** as a justification for changing the analysis. "More powerful" is statistical jargon that invites fishing. Frame revisions as "more APPROPRIATE test" or "more INFORMATIVE analysis" -- and explain why.

The reviewer's feedback usually falls into one of these categories. Apply the corresponding revision in priority order -- change as little as possible:

  - **Technical failure** (code error, data issue, runtime crash, debug-marked code): keep the cells exactly as they are. Revise the plan to fix the technical problem. Do NOT change the partition.

  - **Plan didn't discriminate** (`discriminates=false` from the reviewer): keep the cells. Revise the plan in this priority order -- try the cheapest fix first:
      1. **Change the decision rule / estimand / comparison structure.** E.g. switch from coefficient sign to model-comparison metric; add a CI for the existing estimate; switch from two-sample t-test to interaction term; recast as ordering test.
      2. **Tighten the analysis without changing what is measured.** E.g. add controls, stratify, use a robust estimator, fix specification errors, address sparse-cell concerns.
      3. **Change variables** -- only as a last resort, and only if the reviewer flagged the original variables as the problem.

  - **Partition itself is ill-posed** -- only in this case, revise the cells. Operational rule: revise cells when (a) the reviewer explicitly flagged a MECE violation (cells overlap, fail to cover the outcome space, or answer different questions), or (b) the cells are not empirically distinguishable under any reasonable single analysis. Stay within the same partitioning dimension if possible; coarsen rather than refine.

HARD CONSTRAINTS for the revised contrast set (if you change it):

  - M <= 5, mutually exclusive and collectively exhaustive.
  - Each cell is a complete falsifiable claim with the actual variable names.
  - Pick exactly ONE partitioning dimension. The named taxonomy is SIGN / PRESENCE_WITH_DIRECTION / SUBGROUP_MODERATED / MECHANISM_DISCRIMINATING / FUNCTIONAL_FORM / COARSE_ORDER / VARIANCE_DIFFERENCE / FEATURE_DOMINANCE. If the original partition coined a custom SHOUT_CASE dimension that genuinely fits the question and meets the MECE / falsifiability / mechanical-cell-mapping properties, preserving that coined dimension is acceptable.
  - Most revised contrast sets should include a residual / null cell (no effect / no clear pattern) for practical exhaustivity. Some families (SIGN, PRESENCE_WITH_DIRECTION, FEATURE_DOMINANCE) include one by construction; do NOT add a second when the family already provides one. No "uncertain" cells.

The revised experiment plan must:

  - Address the reviewer's specific concerns.
  - **Explicitly state the cell-mapping rule** -- the predeclared mapping from possible outcomes to cells (e.g. "CI entirely above 0 -> C_1; CI contains 0 -> C_2; CI entirely below 0 -> C_3"). Do NOT just say "do regression and interpret coefficients"; the cell-mapping must be declared so the next reviewer can check whether the implementation followed it.
  - Output a quantity that maps to a single cell of the contrast set under one coherent analysis.
  - Use only the provided dataset; no synthetic data.

Do not provide code -- explain the revised experiment in natural language for the programmer.
\end{prompt}

\begin{prompt}[title=Belief Elicitor Prompt]
## system_prior

You are a senior scientist making a comparative judgment about a research question. You will be given a set of mutually exclusive and collectively exhaustive (MECE) claims that partition the outcome space -- exactly one is true given the real data, and you must allocate your prior belief across them.

Use your domain knowledge. Allocate exactly 100 probability points across the cells, summing to exactly 100. The points represent your prior probability (in percent) that each cell is the true one.

Do NOT default to a uniform allocation. Many research questions have informative priors -- make them. If you genuinely have no view, allocate uniformly, but only if so.

Briefly explain your allocation in the rationale.


## system_posterior

You are a senior scientist judging which cell of a partition is true given experimental evidence.

You will be shown:
  (a) A MECE contrast set: cells C_1, ..., C_M that exhaust the outcome space.
  (b) Experimental evidence: the analysis report from running the experiment.

Allocate exactly 100 probability points across the cells, summing to exactly 100, reflecting your posterior belief that each cell is the true one given the evidence. Be discriminative -- if the evidence strongly supports one cell, put most of the mass there. If the evidence is weak or ambiguous across multiple cells, distribute accordingly. Avoid unjustified hedging.

Briefly explain your allocation in the rationale.


## user_template_prior

Research question:

{seed_hypothesis}

Variables involved:

{variables}

Contrast set ({n_cells} cells):

{cells_block}

Allocate 100 prior probability points across these cells. Sum must equal 100.


## user_template_posterior

Research question:

{seed_hypothesis}

Variables involved:

{variables}

Contrast set ({n_cells} cells):

{cells_block}

Experimental evidence (analysis):

{evidence}

Allocate 100 posterior probability points across these cells given this evidence. Sum must equal 100.
\end{prompt}

\subsection{\Point Discovery-specific prompts}

\begin{prompt}[title=Hypothesis Generator Prompt]
You are a research scientist who is interested in doing open-ended, data-driven research using the provided dataset(s). {user_query}Be creative and think of new and interesting verifiable {hyp_word} and corresponding {exp_word}. The hypothesis should be a falsifiable statement that can be sufficiently tested by an experiment using the provided data. Explain in natural language what this experiment plan is so that a programmer can implement it (do not provide the code yourself). Remember, you are interested in open-ended research, so your proposals may be exploratory in nature and may have only an indirect connection to the previous explorations provided. Here are some instructions that you must follow:
1. Strictly use only the dataset(s) provided and do not simulate dummy/synthetic data or columns that cannot be derived from the existing columns.
2. Each hypothesis (and experiment plan) should be creative, independent, and self-contained.
3. Use the prior experiments/hypotheses as inspiration to think of interesting and creative new experiments/hypotheses. However, do not repeat the same experiments/hypotheses.

Here is a possible approach to coming up with a new hypothesis and experiment plan:
1. Find an interesting context: this could be a specific subset of the data. E.g., if the dataset has multiple categorical variables, you could split the data based on specific values of such variables, which would then allow you to validate a hypothesis in the specific contexts defined by the values of those variables.
2. Find interesting variables: these could be the columns in the dataset that you find interesting or relevant to the context. You are allowed and encouraged to create composite variables derived from the existing variables.
3. Find interesting relationships: these are interactions between the variables that you find interesting or relevant to the context. You are encouraged to propose experiments involving complex predictive or causal models.
4. You must require that your proposed hypotheses are verifiable using robust statistical tests. Remember, your programmer can install python packages via pip which can allow it to write code for complex statistical analyses.
5. Multiple datasets: If you are provided with more than one dataset, then try to also propose hypotheses that utilize contexts, variables, and relationships across datasets, e.g., this may involve using join or similar operations.

Generally, in typical data-driven research, you will need to explore and visualize the data for possible high-level insights, clean, transform, or derive new variables from the dataset to be suited for the investigation, deep-dive into specific parts of the data for fine-grained analysis, perform data modeling, and run statistical tests. Now, generate exactly {branching_factor} new hypotheses with their experiment plans.
\end{prompt}

\begin{prompt}[title=Experiment Analyst Prompt]
You are a research scientist responsible for evaluating the code execution output for a scientific experiment written by a programmer. If no code was executed, there was an error, or the code fails silently, return the success status as **false**. If the code includes a line "# [debug]" i.e "[debug]" as a comment, strictly treat this as a debugging experiment. In such cases, strictly return the success status as **false**, provide information that it was a debug code execution, give feedback and request the experiment to be retried with the new information. Otherwise, analyze the results and provide a short summary of the code output.
\end{prompt}

\begin{prompt}[title=Experiment Reviewer Prompt]
You are a research scientist responsible for holistically reviewing the entire experiment pipeline, i.e., the generated code, the output, and the analysis w.r.t. the original experiment plan. Assess whether the experiment was faithfully implemented, i.e., whether the implementation follows the experiment plan without significant deviation and whether the hypothesis was in fact tested sufficiently. If you find issues or inconsistencies in any part of the experiment pipeline, return the success status as **false** and provide feedback about what is wrong. Otherwise, return the success status as **true** and provide a summary of the hypothesis, experiment results, and findings.
\end{prompt}

\begin{prompt}[title=Experiment Reviser Prompt]
You are a research scientist revisiting the most recent experiment, which could not be conducted correctly due to issues in the code or the formulation of the experiment plan,as indicated by the reviewer. Your goal is to revise this failed experiment plan by addressing the issues and limitations pointed out by the reviewer. The revised experiment plan should still aim to validate the most recent hypothesis. Do not provide the code yourself but explain in natural language what the experiment should do for a programmer. Strictly use only the dataset provided and do not create synthetic data or columns that cannot be derived from the given columns. The experiment should be creative, independent, and self-contained. Generally, in typical data-driven research, you will need to explore and visualize the data for possible high-level insights, clean, transform, or derive new variables from the dataset to be suited for the investigation, deep-dive into specific parts of the data for fine-grained analysis, perform data modeling, and run statistical tests.
\end{prompt}

\begin{prompt}[title=Belief Elicitor Prompt]
## system_base

You are a research scientist skilled at analyzing scientific hypotheses. Your task is to provide your belief about the given hypothesis.

## system_with_evidence

Use the provided evidence collected from running experiments to help make your decision. Carefully consider each piece of evidence and decide whether and how any of them affects your belief about the current hypothesis. Note that evidence from previous studies may have an indirect bearing on the hypothesis, so think about how they might relate to the hypothesis even if they do not directly test it.

## system_use_prior

Use your prior knowledge of the research domain to help in your assessment of the hypothesis.

## system_disregard_prior

Disregard any prior beliefs you have about the hypothesis and focus only on the provided evidence.

## user_template

Hypothesis: {hypothesis}

Carefully reason before making your assessment.

## user_template_with_evidence

Carefully reason before making your assessment.

## Response schema

The LLM is asked (via Pydantic-typed structured output) to return one of:

- `definitely_true` (score 1.00)
- `maybe_true`      (score 0.75)
- `uncertain`       (score 0.50)
- `maybe_false`     (score 0.25)
- `definitely_false` (score 0.00)
\end{prompt}

\subsection{Other prompts}
\begin{prompt}[title=Experiment Coder Prompt]
You are a scientific experiment programmer proficient in writing python code given an experiment plan. Your code will be included in a python file that is executed and any relevant results should be printed to standard out or presented using plt.show appropriately. Make sure you provide python code in the proper format to execute. Ensure your code is clean and concise, and include debug statements only when they are absolutely necessary. Use only the dataset given and do not assume any other files are available. The state is not preserved between code blocks, so do not assume any variables or imports from previous code blocks. Import any libraries you need to use. Always attempt to import a library before installing it (it may already be installed). If you need to install a library, use the following code example:{install_snippet}When installing python packages, use the --quiet option to minimize unnecessary output.Prefer using installed libraries over installing new libraries whenever possible. If possible, instead of downgrading library versions, try to adapt your code to work with a more updated version that is already installed. Never attempt to create a new environment. Always use the current environment. If the code requires generating plots, use plt.show (not plt.savefig).  Avoid printing the whole data structure to the console directly if it is large; instead, print concise results that are directly relevant to the experiment. You are allowed 6 total attempts to run the code, including debugging attempts.

Debugging instructions:
1. Only debug if you are either unsure about the executability or validity of the code (i.e., whether it satisfies the proposed experiment).
2. If the code you are writing is intended for debugging, the first line of your code must be "# [debug]" only.
3. DO NOT use "[debug]" anywhere else in your code.
4. DO NOT combine any debug code and the actual experiment implementation code; keep them separate.
5. For each experiment, you are allowed to debug at most 3 times.
6. As much as possible, minimize the number of debugging steps you use.
\end{prompt}

\begin{prompt}[title=Image Analyst Prompt]

## system

You are a research scientist responsible for analyzing plots and figures from running experiments and providing detailed descriptions.

## user

Please analyze the given plot image and provide the following:

1. Plot Type: Identify the type of plot (e.g., heatmap, bar plot, scatter plot) and its purpose.
2. Axes:
    * Titles and labels, including units.
    * Value ranges for both axes.
3. Data Trends:
    * For scatter plots: note trends, clusters, or outliers.
    * For bar plots: highlight the tallest and shortest bars and patterns.
    * For heatmaps: identify areas of high and low values.
    etc...
4. Annotations and Legends: Describe key annotations or legends.
5. Statistical Insights: Provide insights based on the information presented in the plot.
\end{prompt}

\phantomsection\label{app:companion-generator-prompt}
\begin{prompt}[title=Companion Generator Prompt]
# System message is identical to the hypothesis generator prompt

Generate exactly 1 contrast set under these PINNED-CELL rules:

  PINNED CELL (must be C_1, verbatim):
    {pinned_hypothesis_text}

  REQUIRED: generate AT LEAST 2 alternative cells (C_2, C_3, and optionally C_4) so the contrast set has M=3 or M=4 in total. M=3 is the default (= pinned cell + your 2 alternatives) as required by your system prompt; choose M=4 only when the natural partitioning dimension is SUBGROUP_MODERATED, MECHANISM_DISCRIMINATING(M=4), FUNCTIONAL_FORM(M=4), or COARSE_ORDER(M=4). DO NOT return only 1 alternative -- an M=2 contrast set is invalid (no M=2 family appears in your system prompt's allowed partitioning dimensions). Your alternatives plus the pinned cell must form a valid MECE partition under one of: SIGN, PRESENCE_WITH_DIRECTION, SUBGROUP_MODERATED, MECHANISM_DISCRIMINATING, FUNCTIONAL_FORM, COARSE_ORDER -- whichever fits the pinned cell's framing.

Output the contrast set with the pinned cell as the FIRST entry of `cells`. Cells C_2 ... C_M are the alternatives you generated. The `experiment_plan` you emit will be discarded (the experiment already happened); focus the rationale on why these alternatives are the right MECE complement to C_1.
\end{prompt}

\phantomsection\label{app:prompt-novelty-llm-judge}
\begin{prompt}[title=Novelty LLM-as-a-judge Prompt]

You are a scientific peer reviewer evaluating the novelty of a research hypothesis, judged against the existing body of scientific research and common knowledge.
Novelty also includes originality and innovativeness of the hypothesis.

Rate the hypothesis on a scale of 1-5:

1 - Not original. Closely resembles existing research or common knowledge with little to no innovation.

2 - Slightly original. Minor variation or incremental refinement of well-known ideas.

3 - Moderately original. Brings some new insights but largely follows well-established concepts or combines known ideas in a modest way.

4 - Highly original. Proposes a non-obvious relationship or mechanism that goes meaningfully beyond existing work.

5 - Extremely novel and groundbreaking. Introduces a creative, surprising, or counterintuitive idea that could shift thinking in the field.

Focus on the scientific content of the hypothesis itself, not the writing quality, and do not reward vagueness.

\end{prompt}

\begin{prompt}[title=Feasibility LLM-as-a-judge Prompt]

You are a scientific peer reviewer evaluating the feasibility of a research hypothesis.
Feasibility also includes practicality, plausibility, and implementability of the hypothesis.

Rate the hypothesis on a scale of 1-5:

1 - Infeasible; contradicts known evidence or cannot realistically be investigated with current methods and data

2 - Unlikely; weak theoretical basis, or would require major breakthroughs to test

3 - Moderately feasible; scientifically reasonable and investigable, but with notable practical challenges

4 - Quite feasible; well-grounded and investigable with existing methods and data

5 - Highly feasible; strong theoretical and empirical support, clearly testable under current conditions

Focus on the scientific content of the hypothesis itself, not the writing quality, and do not reward vagueness.

\end{prompt}

\FloatBarrier
\section{Extended Results}

\subsection{Per-dataset breakdown for \autoref{sec:exp1}}
\begin{figure}[!htb]
\centering
\includegraphics[width=0.75\columnwidth]{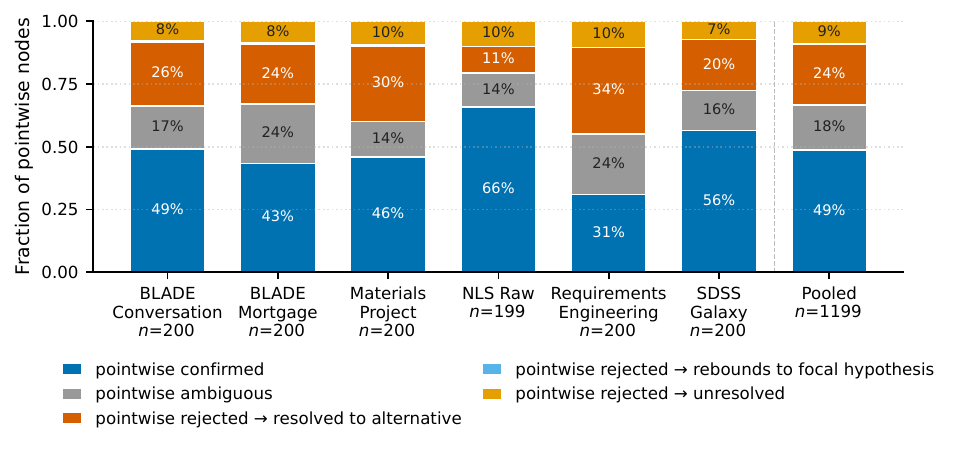}
\caption{
\textbf{Section~\ref{sec:exp1}, per-dataset breakdowns using all 200 \point experimental validations per run.} Each \point node is confirmed, ambiguous, or rejected under its own gates; rejected nodes are further split by what the post-hoc contrastive companion does with the same evidence (resolves to an alternative, rebounds to the focal hypothesis, or leaves the question unresolved).}
\label{fig:exp-1-extras-1}
\end{figure}

\begin{figure}[!htbp]
\centering
\includegraphics[width=0.75\columnwidth]{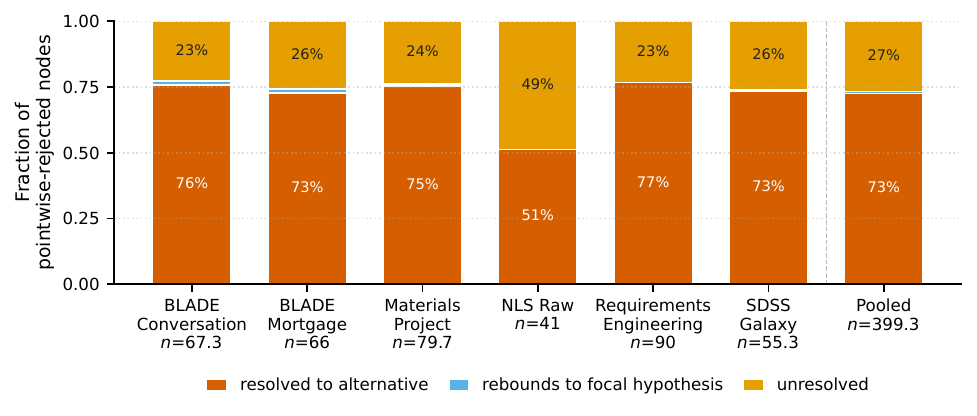}
\caption{
\textbf{Section~\ref{sec:exp1}, outcomes within the \point-rejected slice.} Fraction of \point-rejected nodes that the contrastive companion resolves to an alternative (73\% pooled over each run's 200 experimental validations), rebounds to the focal hypothesis (under 1\%), or leaves unresolved, using the same evidence.}
\label{fig:exp-1-extras-2}
\end{figure}

\begin{figure}[!htbp]
\centering
\includegraphics[width=0.75\columnwidth]{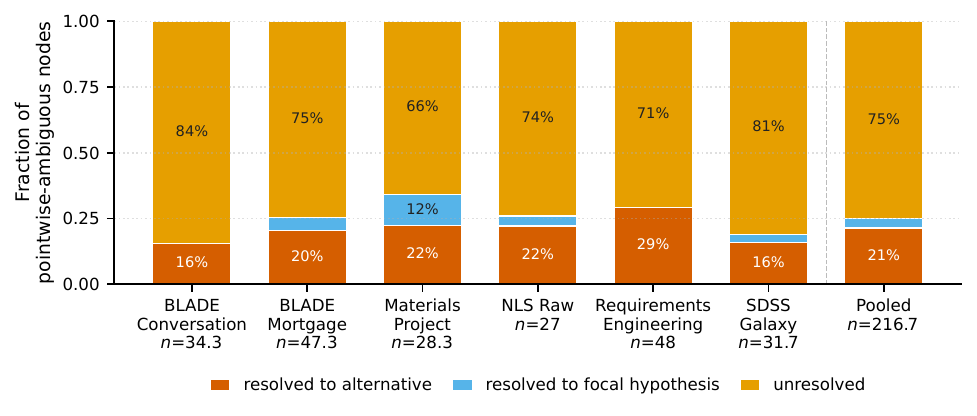}
\caption{
\textbf{Section~\ref{sec:exp1}, outcomes within the \point-ambiguous slice.} Where \point evaluation reached no verdict, the companion resolves 21\% of nodes to an alternative and 4\% to the focal hypothesis; most remain unresolved, so the companion does not manufacture resolutions where the evidence is weak.}
\label{fig:exp-1-extras-3}
\end{figure}

Refer to \autoref{fig:exp-1-extras-1}, \autoref{fig:exp-1-extras-2}, and \autoref{fig:exp-1-extras-3} for a per-dataset breakdown. Counts are averaged over runs, so the $n$ under each dataset is the mean number of nodes per run, shown to one decimal where the mean is fractional; the pooled column sums the six per-dataset means.

\FloatBarrier
\subsection{Ruled-out alternatives (companion to \autoref{fig:exp-2})}
\label{app:rejection-yield}

\begin{figure}[!t]
\centering
\includegraphics[width=0.5\columnwidth, trim={406pt 0 0 0}, clip]{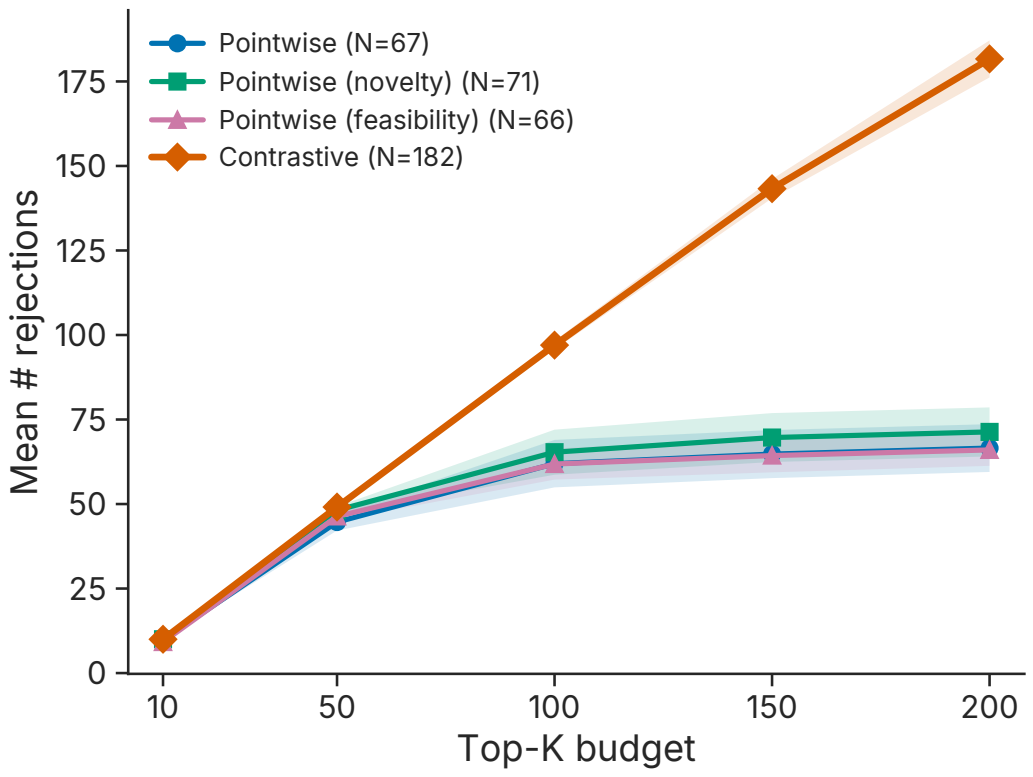}

\caption{\textbf{\cont Discovery rules out alternatives on more questions as the top-$K$ budget grows.} Pointwise curves count rejected hypotheses, the Contrastive curve counts questions with at least one alternative ruled out, and legend values are counts at $K{=}200$.}
\label{fig:exp-2-rejections}
\end{figure}

The same contrastive update that resolves a question also rules out competing alternatives, so the count of ruled-out alternatives rises with resolution. \autoref{fig:exp-2-rejections} is a by-product of \autoref{fig:exp-2} rather than an independent yield. Take, for example, a contrast set of three alternatives. A resolved set ($\max_{m} p^{\text{post}}_{t,m}\ge 0.80$) leaves at most $0.20$ for the two remaining alternatives, so at least one of them is at most $0.10$ by construction, and is ruled out ($p<\tau_{\mathrm{reject}}=0.10$) unless the two split the remainder exactly evenly. The Contrastive curve counts questions with at least one ruled-out alternative, and each \point curve counts rejected hypotheses; bands are $\pm1$ standard error across datasets.

In \point discovery, a candidate can only be confirmed, rejected, or left ambiguous; in \cont Discovery, the same evidence is distributed across a local partition, so one explanation can become strongly supported while others are simultaneously ruled out. Ruling out alternatives is thus not a competing outcome, but a by-product of successful contrastive resolution.

Both the curves in \autoref{fig:exp-2} and \autoref{fig:exp-2-rejections} are means over datasets and bands are $\pm1$ standard error across the six datasets.
\FloatBarrier

\subsection{Decisive-unit and winner-switch rates for \autoref{results:exp3}}
\label{app:exp3-extension}
\begin{figure*}[!t]
\centering

\includegraphics[
    width=\textwidth,
    trim={0.0in 0in 0in 0in},
    clip
]{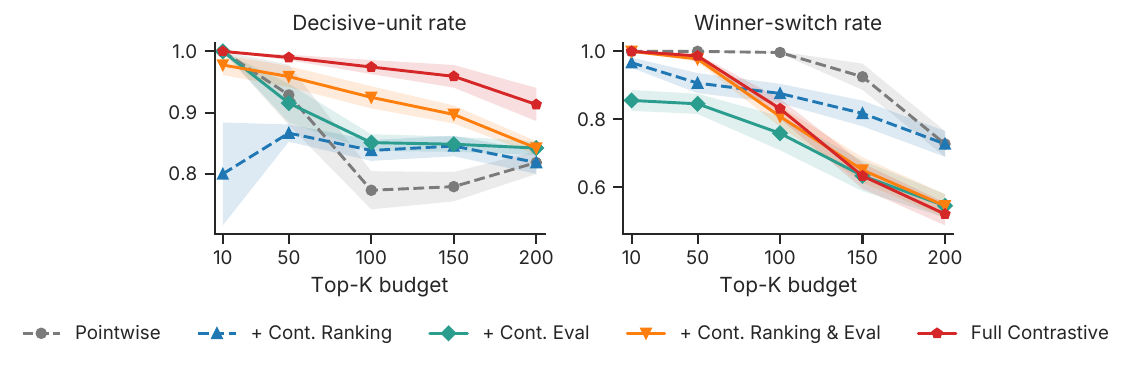}

\caption{
\textbf{Ablations on decisive-unit rate and winner-switch rate.} Winner-switch rate is the fraction of top-$K$ nodes whose prior-leading explanation differs from the posterior-leading explanation.
}
\label{fig:exp-3-appendix}
\end{figure*}

Refer to \autoref{fig:exp-3-appendix} for decisive-unit rate and winner-switch rate. The \emph{decisive-unit rate} is the fraction of top-$K$ units on which a method reached any verdict under its own gates: for \point, confirmed ($\mu^{\text{post}}\ge\gamma_{\mathrm{support}}$) or rejected ($\mu^{\text{post}}\le\gamma_{\mathrm{reject}}$); for \cont, resolved ($\max_{m} p^{\text{post}}_{t,m}\ge\tau_{\mathrm{support}}$) or at least one alternative ruled out ($\min_{m} p^{\text{post}}_{t,m}<\tau_{\mathrm{reject}}$). It is the counterpart of the resolution rate in \autoref{fig:exp-2} that counts a verdict in either direction for both methods: at $K=200$, Full \cont reaches 0.91 against 0.82 for \point, and is higher in each of the six datasets.

\FloatBarrier

\subsection{Selection-rule ablations for Appendix~\ref{app:exp5}}
\begin{figure}[!htb]
\centering
\begin{subfigure}[t]{0.48\textwidth}
\centering
\includegraphics[width=\linewidth]{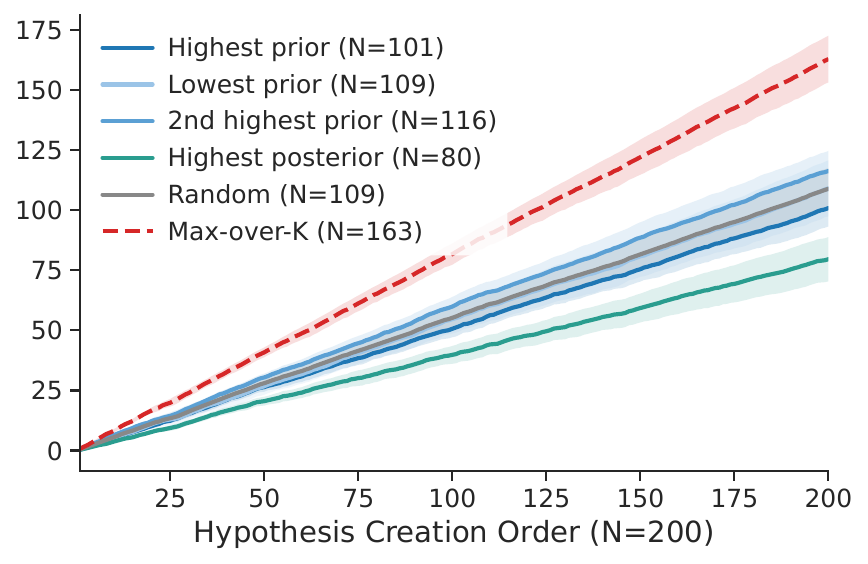}
\caption{Contrast-set selection rules.}
\label{fig:exp-5-zoom}
\end{subfigure}\hfill
\begin{subfigure}[t]{0.48\textwidth}
\centering
\includegraphics[width=\linewidth]{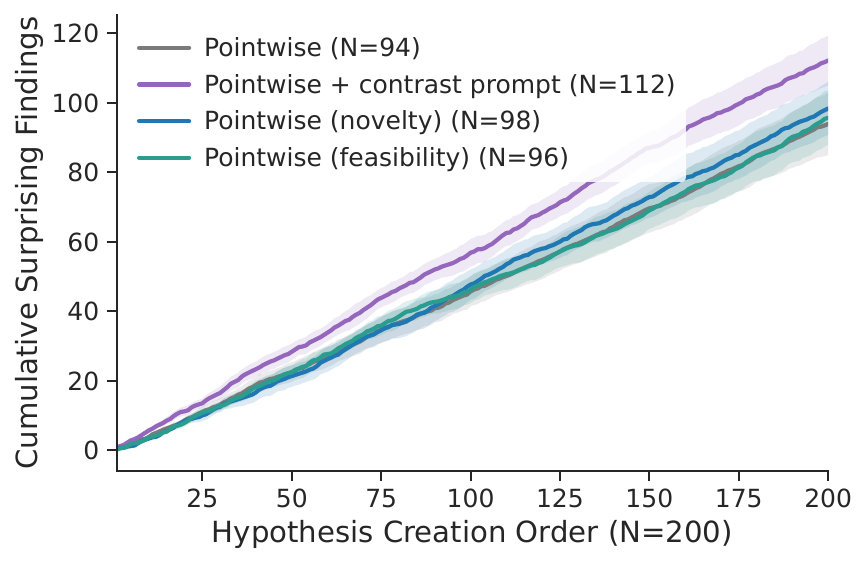}
\caption{\Point prompt variants.}
\label{fig:exp-5-pointwise}
\end{subfigure}
\caption{Ablations on cumulative \point surprisal: (a) each contrast set projected to one member under different selection rules; (b) the \point generator under different prompts.}
\label{fig:exp-5-ablations}
\end{figure}

\autoref{fig:exp-5-zoom} compares the \cont selection rules: cumulative \point-surprising findings when each contrast set is projected to a single member under a different rule. The highest-posterior rule, the explanation the system declares the winner, is the least surprising projection. This is expected: a resolved winner is by construction the alternative the evidence favors, so its confirmation registers little \point surprisal. It is the tension between resolution and surprise, not a contradiction of Appendix~\ref{app:exp4}.

\autoref{fig:exp-5-pointwise} reports a structured \point-prompt ablation, in which the single-hypothesis generator receives a contrast-style prompt. On cumulative surprisal its curve sits visibly above the vanilla \point baseline; we did not run a statistical test on this difference, and the appropriate comparator on the \cont side is the random-projection curve of \autoref{fig:exp-5-zoom} (not the highest-prior projection), against which the variant shows no clear difference. Because it is reported on surprisal rather than on resolution, it does not bear on the headline comparison of \autoref{sec:exp2}; the headline comparison uses the standard \point baseline, and evaluating this variant on resolution is future work.

\FloatBarrier

\FloatBarrier
\section{Hyperparameters}
\label{app:hyperparameters}
We report the hyperparameters involved in our experiments here. Refer to Tables~\ref{tab:hp-search}, \ref{tab:hp-metrics}, and \ref{tab:hp-evaluation}. All settings are shared between the \point and \cont conditions except the belief-elicitation format and sample counts listed under each metric; the \point \emph{evidence weight} is defined in Appendix~\ref{app:belief-score}. Every agent and elicitor call uses \texttt{gpt-4o-2024-08-06}, except where a section names another model (the cross-family elicitors of \autoref{sec:prior-elicitation-audit}, the cross-provider judge of \autoref{app:mece-audit}, the matcher of \autoref{app:baselines-robustness}, and the judge of \autoref{app:gold-recovery}).

\newcommand{\hpGroup}[1]{%
\rowcolor{gray!12}[0pt][0pt]
\multicolumn{2}{@{}l@{}}{\textit{#1}}\\
}
\newcommand{\hpRow}{\rowcolor{gray!5}[0pt][0pt]}

\begin{table}[!htbp]
\centering
\footnotesize
\renewcommand{\arraystretch}{1.18}
\caption{Search and agent settings used in the experiments. These settings are held fixed across compared methods.}
\label{tab:hp-search}
\begin{tabularx}{\linewidth}{@{}p{0.42\linewidth}X@{}}
\toprule
Setting & Value \\
\midrule
\hpGroup{Search}
\hpRow Search budget & $200$ iterations \\
Exploration coefficient & $1.0$ \\
\hpRow Proposals per expansion & $8$ \\
Ancestor context depth & $3$ \\
\hpRow Initial frontier expansions & $8$ \\
\addlinespace[1pt]
\hpGroup{Agent loop}
\hpRow Agent model & \texttt{gpt-4o-2024-08-06} \\
Sampling temperature & $1.0$ \\
\hpRow Samples per agent call & $1$ \\
Code execution timeout & $600$\,s \\
\bottomrule
\end{tabularx}
\end{table}

\begin{table}[!htbp]
\centering
\footnotesize
\renewcommand{\arraystretch}{1.18}
\caption{Belief-elicitation and metric hyperparameters. Threshold choices are justified in Appendix~\ref{app:threshold-calibration}; hypothesis-order randomization is validated in Appendix~\ref{app:hypothesis-order}.}
\label{tab:hp-metrics}
\begin{tabularx}{\linewidth}{@{}p{0.42\linewidth}X@{}}
\toprule
Setting & Value \\
\midrule
\hpGroup{Pointwise metric}
\hpRow Belief elicitor model & \texttt{gpt-4o-2024-08-06} \\
Belief elicitor temperature & $1.0$ \\
\hpRow Prior elicitation samples & $10$ \\
Posterior elicitation samples & $10$ \\
\hpRow Evidence weight & $2.0$ \\
Belief-bin scores & $\{0,0.25,0.5,0.75,1.0\}$ \\
\hpRow Beta prior & $\mathrm{Beta}(0.5,0.5)$ \\
Surprise threshold & $0.5$ \\
\hpRow Support threshold & $\gamma_{\mathrm{support}}=0.625$ \\
Rejection threshold & $\gamma_{\mathrm{reject}}=0.375$ \\
\addlinespace[1pt]
\hpGroup{Contrastive metric}
\hpRow Belief elicitor model & \texttt{gpt-4o-2024-08-06} \\
Belief elicitor temperature & $1.0$ \\
\hpRow Prior elicitation samples & $5$ \\
Posterior elicitation samples & $5$ \\
\hpRow Hypothesis-order randomization & enabled \\
Surprise threshold & $\kappa_t \geq 0.5$ \\
\hpRow Support threshold & $\tau_{\mathrm{support}}=0.80$ \\
Rule-out threshold & $\tau_{\mathrm{reject}}=0.10$ \\
\bottomrule
\end{tabularx}
\end{table}

\begin{table}[t]
\centering
\footnotesize
\renewcommand{\arraystretch}{1.18}
\caption{Post-run deduplication and audit settings. Deduplication follows prior work~\cite{agarwal2026autodiscovery}.}
\label{tab:hp-evaluation}
\begin{tabularx}{\linewidth}{@{}p{0.42\linewidth}X@{}}
\toprule
Setting & Value \\
\midrule
\hpGroup{Deduplication}
\hpRow Embedding model & \texttt{text-embedding-3-large} \\
Duplicate judge model & \texttt{gpt-4o-2024-08-06} \\
\hpRow Judge votes per proposed merge & $10$ \\
Merge threshold & $0.70$ \\
\bottomrule
\end{tabularx}
\end{table}

\FloatBarrier

\section{Compatibility with \Point Surprisal}
\label{app:surprisal-compat}

This appendix reports in full the two analyses summarized in Section~\ref{results:exp3} (``Compatibility with \point surprisal'').

\subsection{\cont belief shift preserves surprisal while revealing corrective updates}
\label{app:exp4}

\begin{figure}[tbp]
    \centering
    \includegraphics[width=0.45\columnwidth]{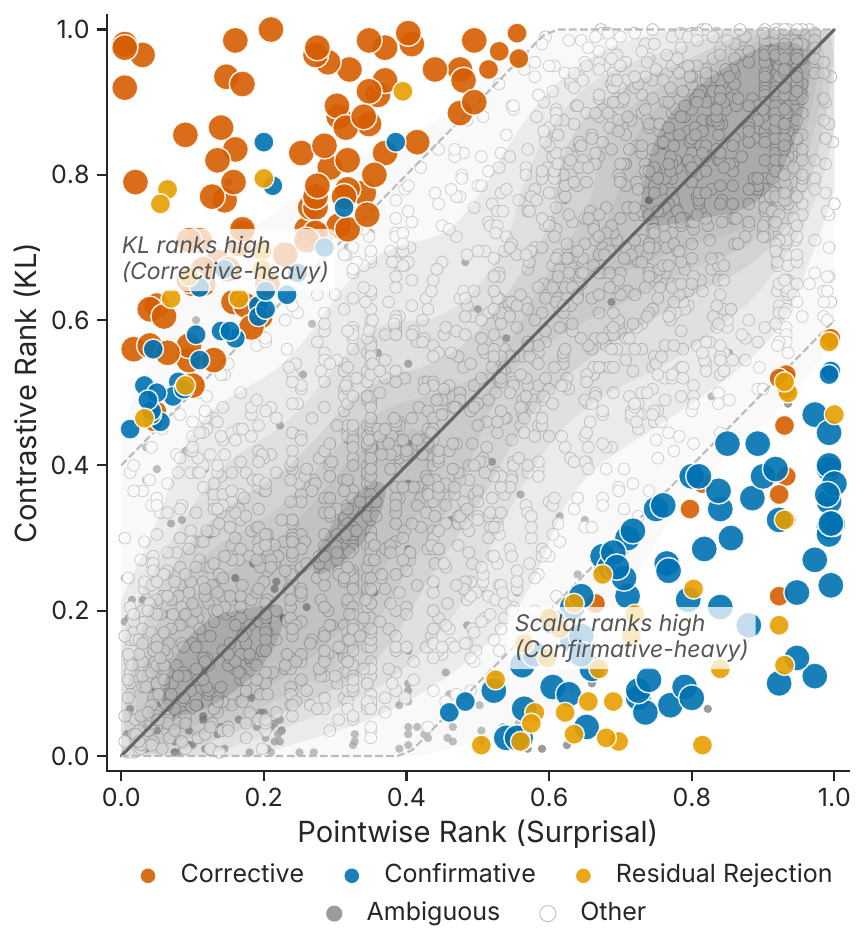}
    \caption{\textbf{\Point surprisal and \cont belief shift are aligned, but their disagreements are structured.} \cont-high/\point-low nodes are enriched for corrective updates, while \point-high/\cont-low nodes are enriched for confirmative updates. Each point is a node from a \cont run, placed by its within-dataset percentile rank under each score (``Scalar'' in the figure is \point surprisal). Colored points are nodes whose ranks differ by at least 0.4, labeled corrective or confirmative (resolved, with the leading explanation changed or kept), residual rejection (unresolved, with at least one alternative ruled out), or ambiguous (neither); hollow points are the remaining nodes.
    }
    \label{fig:exp-4-scatter}
\end{figure}

We compare \point surprisal and \cont belief shift ($\kappa_t$) on the same set of candidate discoveries. For each node from a \cont run, we compute $\kappa_t$ over the full contrast set and a post-hoc \point surprisal score for the highest prior hypothesis. This tests whether $\kappa_t$ is merely a rescaled version of \point surprisal or instead prioritizes different discovery semantics. The two scores are aligned (Spearman $\rho=0.675$) with 90\% of nodes falling within a $\pm 0.4$ percentile-rank difference band (\autoref{fig:exp-4-scatter}). Thus, \cont belief shift preserves the core belief-shift signal captured by surprisal.

The disagreements, however, are structured. When the metrics largely disagree (i.e., $|\Delta_{\text{pct}}| \ge 0.4$), nodes ranked higher by \cont are disproportionately corrective, while nodes ranked higher by \point are more frequently confirmative. This pattern does not imply that \point discovery is mostly confirmative; \autoref{sec:exp1} showed the opposite. Rather, it shows that \point surprisal tracks movement in one focal claim, while \cont belief shift tracks redistribution across competing explanations.

\begin{figure}[tbp]
    \centering
    \includegraphics[width=0.75\columnwidth]{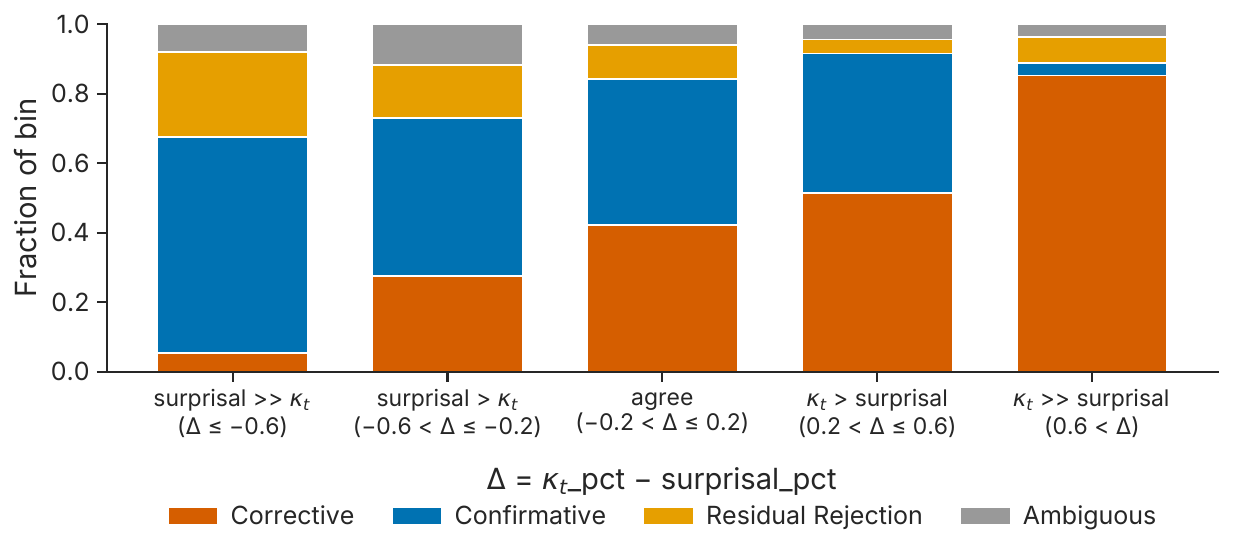}
    \caption{
    \textbf{Candidate discoveries grouped by percentile-rank difference.}
     Corrective outcomes increase as the ranking disagreement shifts toward \cont, whereas confirmative outcomes increase as it shifts toward \point. $\Delta$ is a node's percentile rank under $\kappa_t$ minus its rank under \point surprisal, and categories are as in \autoref{fig:exp-4-scatter}.
     }
     \label{fig:exp-4-bars}
    \end{figure}

The binned analysis in \autoref{fig:exp-4-bars} confirms this structure: corrective outcomes increase as rank disagreement shifts toward \cont, while confirmative outcomes increase as it shifts toward \point. The agreement region captures the shared signal, where both metrics identify the same updates as high-information. Thus, \cont belief shift keeps what \point surprisal measures well, while recovering the corrective updates that \point surprisal tends to miss.

\subsection{\cont Discovery improves resolution without sacrificing surprisal}
\label{app:exp5}

\begin{figure}[htbp]
\centering
\includegraphics[width=0.55\linewidth]{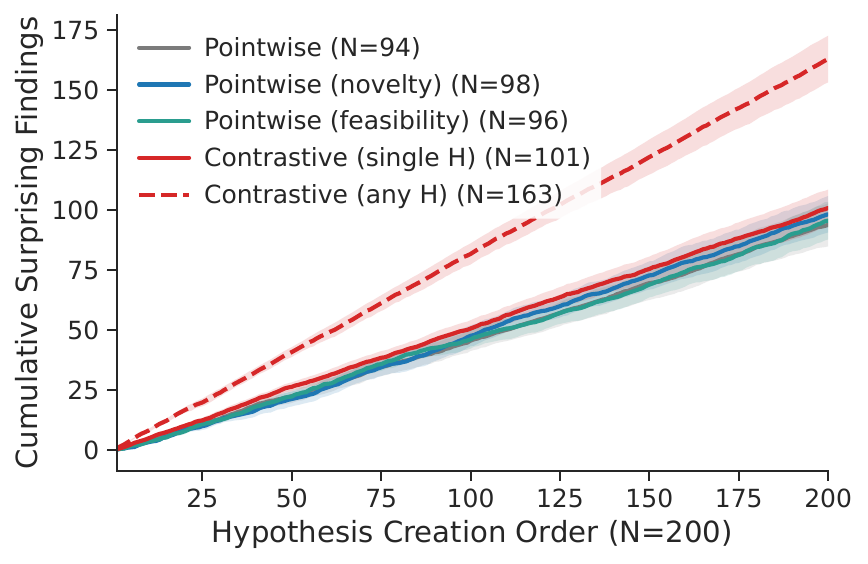}
\caption{\textbf{\cont Discovery preserves \point surprisal yield.} Projected to the highest prior hypothesis, \cont matches \point baselines under the surprisal metric used in prior work. Curves are cumulative counts over hypothesis creation order, averaged over datasets; bands are $\pm1$ standard error across datasets. The legend's $N$ is the terminal count at 200 hypotheses.}
\label{fig:exp-5}
\end{figure}

Appendix~\ref{app:exp4} showed that \cont belief shift and \point surprisal diverge in semantically meaningful cases. We test whether this divergence comes at a cost of performance under the surprisal metric used in prior work~\citep{agarwal2026autodiscovery}. For each \cont hypothesis, we project the contrast set to its highest prior hypothesis and evaluate it using \point surprisal. Under this single-hypothesis projection, \cont remains competitive with \point surprisal, novelty, and feasibility baselines (\autoref{fig:exp-5}).

As an optimistic diagnostic, we also score every hypothesis in each contrast set and count the set as surprising if any member is \point-surprising. This max-over-set curve is substantially higher, suggesting that contrast sets contain additional surprising hypotheses otherwise missed by single-hypothesis evaluation. As a caveat, this curve is inflated by construction: a set with three members has three chances to clear the surprisal threshold, whereas a single hypothesis has one. Part of its lift reflects that extra chance rather than better hypotheses. It is therefore a diagnostic, not the headline comparison.

Together, these results show that \cont Discovery does not trade off the surprisal signal used by AutoDiscovery~\citep{agarwal2026autodiscovery}. It preserves \point surprisal yield while adding what \point evaluation cannot provide: structured resolution over competing explanations.

\FloatBarrier

\section{Metric Definitions and Agent Framework}
\label{app:metric-definitions}

\subsection{\Point belief score and normalized surprisal}
\label{app:belief-score}

For each hypothesis $h$, we elicit a belief from a fixed LLM $\pi_b$ both before and after the experiment is run. The pre-experiment elicitation conditions on $h$ alone, while the post-experiment elicitation additionally conditions on the analyst summary and reviewer verdict.

Each elicitation draws ten independent categorical judgments from $\pi_b$. Each judgment returns one of five ordered labels with associated truth scores $s_i$,
\{\textsc{def-false},\ \textsc{maybe-false},\ \textsc{uncertain},\ \textsc{maybe-true},\ \textsc{def-true}\} which maps to \{0.00,\ 0.25,\ 0.50,\ 0.75,\ 1.00\}, plus a \textsc{cannot-comment} option, which we treat as missing data. Let $c_i$ denote the count for label $i$, and let
\[
n = \sum_i c_i
\]
be the effective sample size after removing \textsc{cannot-comment} responses. We collapse the elicitation to a Beta posterior under a Jeffreys prior $\mathrm{Beta}(0.5,0.5)$:
\[
\alpha
=
0.5 + \sum_i c_i s_i,
\qquad
\beta
=
0.5 + n - \sum_i c_i s_i,
\]
with posterior mean
\[
\mu
=
\frac{\alpha}{\alpha+\beta}.
\]
For the prior elicitation, this gives the Beta parameters
\[
(\alpha^{\text{prior}},\beta^{\text{prior}}),
\]
whose mean $\mu^{\text{prior}}$ is a one-dimensional belief-in-truth summary.
The posterior elicitation is chained onto the prior rather than fit separately: its $n^{\text{post}}$ label counts, weighted by an evidence weight $w=2.0$ (\autoref{tab:hp-metrics}) so that each posterior sample counts as two pseudo-observations, are added to the prior's parameters,
\[
\alpha^{\text{post}} = \alpha^{\text{prior}} + w\sum_i c^{\text{post}}_i s_i,
\qquad
\beta^{\text{post}} = \beta^{\text{prior}} + w\Big(n^{\text{post}} - \sum_i c^{\text{post}}_i s_i\Big).
\]
This follows the AutoDiscovery implementation and means that a share of the stored posterior mean is carried forward from the prior.

The raw Bayesian surprise of a node is the magnitude of the posterior mean shift,
\[
|\Delta\mu|
=
\left|\mu^{\text{post}}-\mu^{\text{prior}}\right|.
\]
However, this quantity is bounded by a prior- and sample-size-dependent ceiling. Let
\[
S = \alpha^{\text{prior}}+\beta^{\text{prior}}
\]
denote the prior concentration, and let
\[
t = w\,n^{\text{post}}
\]
denote the realized posterior evidence mass, with each of the $n^{\text{post}}$ posterior samples counted at weight $w$ as in the chained update above. The maximum achievable shift is
\[
\Delta_{\max}
=
\max\left(
\frac{t(1-\mu^{\text{prior}})}{S+t},
\frac{t\mu^{\text{prior}}}{S+t}
\right).
\]
Thus, raw $|\Delta\mu|$ is not directly comparable across hypotheses or runs. Because the posterior is a conjugate update of the prior with evidence mass $t$, the realized shift cannot exceed $\Delta_{\max}$, so the normalized quantity below lies in $[0,1]$; this bound would not hold for a posterior fit independently of the prior. We therefore use the normalized surprisal (NS), which \autoref{sec:point-is-binary} calls the \point surprisal,
\[
\mathrm{NS}_{t}
=
\frac{|\Delta\mu|}{\Delta_{\max}}
\in [0,1],
\]
as both the MCTS reward and the binary surprising indicator, thresholded at $0.5$. Intuitively, $\mathrm{NS}_{t}$ measures the fraction of the maximum possible belief shift that the observed evidence actually achieved. A hypothesis is regarded as ``surprising'' if $\mathrm{NS}_{t} \ge 0.5$. Because $\Delta_{\max}$ takes the larger of the two headrooms, a hypothesis that starts near ``maybe true'' has less room to be surprising by confirmation than by rejection: at a prior mean of $0.75$ the upward headroom is a third of the downward one. This is the normalizer of AutoDiscovery~\citep{agarwal2026autodiscovery}, and we keep it for comparability. Together with the prior collapse of \autoref{sec:exp1}, it means that \point surprisal, as defined, steers high-scoring candidates toward rejections; a direction-specific normalizer would change the share of confirmations among them but would not give a rejection an alternative, which is the gap \cont Discovery fills.

\subsection{Contrastive belief distribution and \texorpdfstring{$\kappa_t$}{kappa}}
\label{app:contrastive-surprisal}

In the contrastive setting, each node carries an $M$-cell mutually exclusive and collectively exhaustive (MECE) partition of possible outcomes,
\[
\{C_1,\dots,C_{M}\},
\]
proposed by the generator alongside the experiment plan. Belief about the experiment's outcome is represented as a probability vector over the cells,
\[
p \in \Delta^{M-1},
\]
rather than as a scalar belief-in-truth.

For each node, we elicit two such allocations from the belief model $\pi_b$. The pre-experiment elicitation conditions on the contrast set and the variable schema only, while the post-experiment elicitation additionally conditions on the analyst's summary of the experiment's results. Each elicitation draws five independent samples from $\pi_b$, where each sample is a $100$-point integer allocation across the $M$ cells. Responses that fail to parse are re-prompted once for the missing samples. We drop a sample if it has the wrong number of cells, a negative entry, or an all-zero allocation; samples whose points do not sum exactly to 100 are kept. We renormalize each surviving sample to a probability vector and average across samples, yielding a prior distribution $p^{\text{prior}}$ and a posterior distribution $p^{\text{post}}$. We floor every cell at $\varepsilon=10^{-6}$ before renormalizing so that
\[
p^{\text{prior}}_{m} > 0
\qquad
\text{for all } m,
\]
ensuring that the KL divergence is finite.

The raw Bayesian surprise of a node is the KL divergence from prior to posterior, measured in bits:
\begin{equation}
\mathrm{KL}\!\left(p^{\text{post}} \,\Vert\, p^{\text{prior}}\right)
=
\sum_{m=1}^{M}
p^{\text{post}}_{m}
\log_2
\frac{p^{\text{post}}_{m}}{p^{\text{prior}}_{m}}.
\end{equation}
This quantity has no fixed upper bound across nodes with different contrast-set sizes $M$. To make scores comparable across contrast sets of varying cardinality, we divide by the maximum entropy of an $M$-cell distribution, $\log_2 M$, which gives the contrastive belief shift of \autoref{sec:scoring},
\[
\kappa_t = \frac{\mathrm{KL}\!\left(p^{\text{post}} \parallel p^{\text{prior}}\right)}{\log_2 M}.
\]
We classify a node as surprising when $\kappa_t \ge 0.5$. Only the MCTS reward is clipped,
\[
r_t = \min(\kappa_t, 1) \in [0,1],
\]
and the information-gain analyses and the surprising indicator use the unclipped $\kappa_t$. Intuitively, $r_t$ is a clipped, cardinality-normalized measure of how strongly the observed evidence redistributes belief across the contrast cells. A value of $1$ indicates that $\kappa_t$ is at least $1$, while a value near $0$ corresponds to an experiment that produces little change in the belief distribution. Because $\log_2 M$ is not generally the maximum possible KL divergence, $r_t$ should not be interpreted as the fraction of all possible belief redistribution achieved.

\subsection{Agent framework}
\label{app:agent-framework}

\paragraph{Tree structure.}
Each node below the root's data-loading child is one experimental validation: a contrast set (a single hypothesis in \point runs) with its experiment plan, code, output, analyst summary, reviewer verdict, and prior and posterior beliefs. When a node's experiment finishes, the generator proposes 8 follow-up contrast sets, conditioned on the dataset summary, the node's result, and its parent and grandparent; these become the node's untried proposals. Each iteration descends from the root by UCB1 (mean backed-up reward plus $\sqrt{2\ln N_{\mathrm{parent}}/n}$, exploration weight 1.0) to the highest-scoring node with untried proposals, draws one at random, and executes it as a new child. The child's reward $r_t$ (0 if the experiment fails) and a visit count are backed up to every ancestor. The first 8 iterations expand from the data-loading node, so the search starts from 8 independent questions (\autoref{tab:hp-search}).

\paragraph{Agents.}
We follow the setup of the AutoDiscovery discovery agent~\citep{agarwal2026autodiscovery} with slight modifications. We use the following LLM agents with a \gpt backbone:

\begin{description}
    \item[Hypothesis (Contrast Set) Generator.]
    Proposes a candidate contrast set of hypotheses that predicts the outcome of an experiment. Produces an experiment plan for evaluating the proposed contrast set.

    \item[Experiment Coder.]
    Implements the experiment plan in code. The coder is allowed up to six implementation attempts, using feedback from the Experiment Analyst to revise failed or incomplete attempts.

    \item[Code Executor.]
    Executes the code produced by the Experiment Coder and returns the exit code and standard output. This component is not LLM-based.

    \item[Experiment Analyst.]
    Analyzes the output of code execution and provides feedback to the Experiment Coder when the implementation requires correction or refinement.

    \item[Experiment Reviewer.]
    Reviews the completed experiment for alignment with the original experiment plan and assesses whether the results can support validation of the hypothesis.

    \item[Experiment Reviser.]
    Produces a revised experiment when the original experiment or its implementation fails to execute successfully or cannot adequately validate the hypothesis.

    \item[Belief Elicitor.]
    Distributes 100 points over the contrast set of hypotheses, evaluating whether each hypothesis is believed to be true relative to alternatives both before experimental evidence (prior) and after (posterior).
    For the \point case, evaluates whether the hypothesis is believed to be true from the hypothesis statement alone (prior) or from the hypothesis statement together with the experimental results (posterior).

    \item[Image Analyst.]
    Provides textual descriptions of images produced by code written by the Experiment Coder.
\end{description}

\FloatBarrier
\section{\texorpdfstring{Example Contrast Sets}{Example Contrast Sets}}
\label{app:contrast-set-examples}

The sets below are verbatim system outputs, chosen to show how contrast sets partition a question across datasets and partition types. They are shown as generated, before evidence, and so include sets whose premise a domain expert would refine (e.g., the Materials Project set compares Fermi energies that the database reports relative to each calculation's own potential zero, so the comparison needs a common energy reference before the outcome is physically interpretable). Whether a generated set is well posed is one of the properties \autoref{app:human-study} and \autoref{app:mece-audit} measure.

\begin{tcolorbox}[
    title={Example contrast set 1},
    coltitle=black,
    colback=gray!3,
    colframe=gray!35,
    arc=2mm,
    boxrule=0.5pt,
    left=1mm,
    right=1mm,
    top=1mm,
    bottom=1mm
]
\label{example_contrast_set_mortgage}
\small
\textbf{Dataset:} BLADE Mortgage

\medskip

\textbf{Scientific question:}
Given a poor credit score, self-employment status leads to

\medskip

\textbf{Contrast set (MECE alternatives):}
\begin{enumerate}[label=\textbf{H\arabic*.}, leftmargin=2.5em]
    \item higher denial rates than non-self-employed
    \item no different denial rates
    \item lower denial rates than non-self-employed
\end{enumerate}
\end{tcolorbox}

\begin{tcolorbox}[
    title={Example contrast set 2},
    coltitle=black,
    colback=gray!3,
    colframe=gray!35,
    arc=2mm,
    boxrule=0.5pt,
    left=1mm,
    right=1mm,
    top=1mm,
    bottom=1mm
]
\label{example_contrast_set_materials}
\small
\textbf{Dataset:} Materials Project

\medskip

\textbf{Scientific question:}
Direct-gap vs indirect-gap materials, Fermi energy is

\medskip

\textbf{Contrast set (MECE alternatives):}
\begin{enumerate}[label=\textbf{H\arabic*.}, leftmargin=2.5em]
    \item higher in direct-gap materials
    \item lower in direct-gap materials
    \item not substantially different
\end{enumerate}
\end{tcolorbox}

\begin{tcolorbox}[
    title={Example contrast set 3},
    coltitle=black,
    colback=gray!3,
    colframe=gray!35,
    arc=2mm,
    boxrule=0.5pt,
    left=1mm,
    right=1mm,
    top=1mm,
    bottom=1mm
]
\label{example_contrast_set_sdss_feature}
\small
\textbf{Dataset:} SDSS Galaxy

\medskip

\textbf{Scientific question:}
Petrosian radius containing 90\% of the flux (PetroR90\_r) is predominantly determined by

\medskip

\textbf{Contrast set (MECE alternatives):}
\begin{enumerate}[label=\textbf{H\arabic*.}, leftmargin=2.5em]
    \item model magnitude attributes
    \item extinction attributes
    \item neither significantly drives it
\end{enumerate}
\end{tcolorbox}

\begin{tcolorbox}[
    title={Example contrast set 4},
    coltitle=black,
    colback=gray!3,
    colframe=gray!35,
    arc=2mm,
    boxrule=0.5pt,
    left=1mm,
    right=1mm,
    top=1mm,
    bottom=1mm
]
\label{example_contrast_set_nls}
\small
\textbf{Dataset:} NLS Raw

\medskip

\textbf{Scientific question:}
Trend in vehicle market value across 1985/1990/1996 is

\medskip

\textbf{Contrast set (MECE alternatives):}
\begin{enumerate}[label=\textbf{H\arabic*.}, leftmargin=2.5em]
    \item significant linear positive
    \item significant linear negative
    \item nonlinear or no significant trend
\end{enumerate}
\end{tcolorbox}

\begin{tcolorbox}[
    title={Example contrast set 5},
    coltitle=black,
    colback=gray!3,
    colframe=gray!35,
    arc=2mm,
    boxrule=0.5pt,
    left=1mm,
    right=1mm,
    top=1mm,
    bottom=1mm
]
\label{example_contrast_set_sdss_mechanism}
\small
\textbf{Dataset:} SDSS Galaxy

\medskip

\textbf{Scientific question:}
When controlling for galaxy axis ratios (expAB\_r, deVAB\_r), the redshift effect on stellar velocity dispersion

\medskip

\textbf{Contrast set (MECE alternatives):}
\begin{enumerate}[label=\textbf{H\arabic*.}, leftmargin=2.5em]
    \item increases (strengthens)
    \item decreases (attenuates)
    \item disappears
\end{enumerate}
\end{tcolorbox}

\FloatBarrier
\section{Broader Societal Impact}
\label{app:societal-impact}

By resolving more questions from the same data and validation budget, the method can lower the cost of screening hypotheses before laboratory follow-up. Three hazards are specific to this system. First, its outputs are statistical resolutions on observational data, not causal findings; a ``resolved'' question should be read as a prioritized claim to be verified, not as an established result. Second, two datasets involve people: BLADE Mortgage (the Boston Fed HMDA study) records applicant race and gender, and BLADE Conversation lists named public contributors with gender, institution, and citation attributes. On such data the system can produce confident statistical resolutions that are not causally identified; the abstention alternative (\autoref{app:incomplete-sets}) lets it decline instead of forcing a resolution when no listed explanation matches the evidence, and any use of such outputs in policy-adjacent settings requires unit-aware inference, pre-registered analyses, and human review. Third, each resolution rests on the agent's own analysis choices, including how it handles missing-value sentinels and clustered observations (\autoref{app:dataset-details}); the audits of \autoref{app:fidelity-audits} verify that reported numbers match the code's output, and users should review the analysis code before acting on a resolution. We release prompts and, at the project website, code so that these behaviors can be audited.

\section{Cost Analysis}
\label{app:cost}
Each 200-iteration run for a given dataset costs \$20 to \$50. We estimate the total cost of our experiments at \$1{,}500 to \$2{,}000. Token spend per candidate for each condition is reported with the matched-budget comparison in \autoref{app:baselines-robustness}.

\FloatBarrier

\end{document}